%% file: main.tex
\documentclass[11pt, a4paper, copyright, gdm]{google}

\usepackage[authoryear, sort&compress, round]{natbib}

\usepackage{microtype}
\usepackage{graphicx}
\usepackage{subcaption}
\usepackage{booktabs} 
\usepackage{multirow}
\usepackage{tabularx} 
\usepackage{csquotes}
\usepackage{bbm}
\usepackage{hyperref}
\usepackage{xcolor}
\usepackage{enumitem}
\usepackage{hyperref}
\usepackage{stfloats}
\usepackage{listings}
\usepackage{enumitem}
\usepackage{xspace}
\usepackage{longtable}
\newcommand{\assistant}{Assistant\xspace}
\definecolor{darkred}{RGB}{139,0,0}

\uselogo{} 

\title{AI-Assisted Scientific Assessment: A Case Study on Climate Change}

\correspondingauthor{massi@google.com}

\reportnumber{0001} 

\renewcommand{\today}{2026-05-21}

\author[1]{Christian Buck}
\author[2]{Levke Caesar}
\author[1]{Michelle Chen Huebscher}
\author[1]{Massimiliano Ciaramita}
\author[3]{Erich M. Fischer}
\author[4]{Zeke Hausfather}
\author[2]{\"{O}zge Kart Tokmak}
\author[3]{Reto Knutti}
\author[1,5]{Markus Leippold}
\author[2]{Joseph Ludescher}
\author[6]{Katharine J. Mach}
\author[7]{Sofia Palazzo Corner}
\author[2]{Kasra Rafiezadeh Shahi}
\author[2]{Johan Rockstr\"{o}m}
\author[7,8,9]{Joeri Rogelj}
\author[2]{Boris Sakschewski}

\affil[1]{\thepa{}{}}
\affil[2]{Potsdam Institute for Climate Impact Research (PIK),  Member of the Leibniz Association}
\affil[3]{Institute for Atmospheric and Climate Science, ETH Zurich}
\affil[4]{Stripe}
\affil[5]{University of Zurich}
\affil[6]{University of Miami}
\affil[7]{Centre for Environmental Policy, Imperial College London}
\affil[8]{Grantham Institute for Climate Change and Environment, Imperial College London}
\affil[9]{Energy, Climate and Environment Program, International Institute for Applied Systems Analysis}

\begin{abstract}
\input{sections-v2/0a-abstract}
\end{abstract}

\begin{document}

\maketitle

\input{sections-v2/1-Introduction}

\input{sections-v2/1b-relatedliterature}
\input{sections-v2/2-ai-assistance}

\input{sections-v2/3-results}

\input{sections-v2/4-discussion}
\input{sections-v2/5-conclusion}

\bibliography{main}

\newpage
\appendix
\onecolumn
\input{sections-v2/appendix}
\input{sections-v2/amoc_final}
\end{document}

%% file: sections-v2/0a-abstract.tex
The emerging paradigm of AI co-scientists focuses on \emph{verification-rich} tasks, where agents explore search spaces in `guess and check' loops with immediate feedback. This paradigm does not extend to \emph{verification-poor} problems, where repeated evaluation is impossible and ground truth is an assessment of the state of knowledge established through expert consensus. We present an empirical evaluation of a Gemini-based AI environment designed to support collaborative scientific assessment, integrated into a standard scientific workflow. In collaboration with a diverse group of 13 climate scientists, we tested the system on a deliberately challenging topic: the stability of the Atlantic Meridional Overturning Circulation (AMOC).
Our results, corroborated by a post-study survey, show that AI can substantially accelerate the scientific workflow. The group produced a synthesis of 79 papers through 104 revision cycles in just over 46 person-hours---a process that participants estimated would typically require 100--200 person-hours. AI contribution was significant: most AI-generated content was retained in the report. However, expert oversight was crucial: scientists contributed the majority of the final content, expanding and elevating the assessment to rigorous scientific standards. We propose simple metrics, adapted from established NLP evaluation frameworks, to quantify human-AI co-authoring dynamics at the sentence level. The findings from our case study suggest that AI scientific assistance can extend to verification-poor tasks, but that hybrid intelligence architectures---where human expertise remains central---are essential for ensuring scientific rigor.

%% file: sections-v2/1-Introduction.tex
\section{Introduction}
\label{sect:intro}

\begin{figure*}
    \centering
    \includegraphics[width=0.99\linewidth]{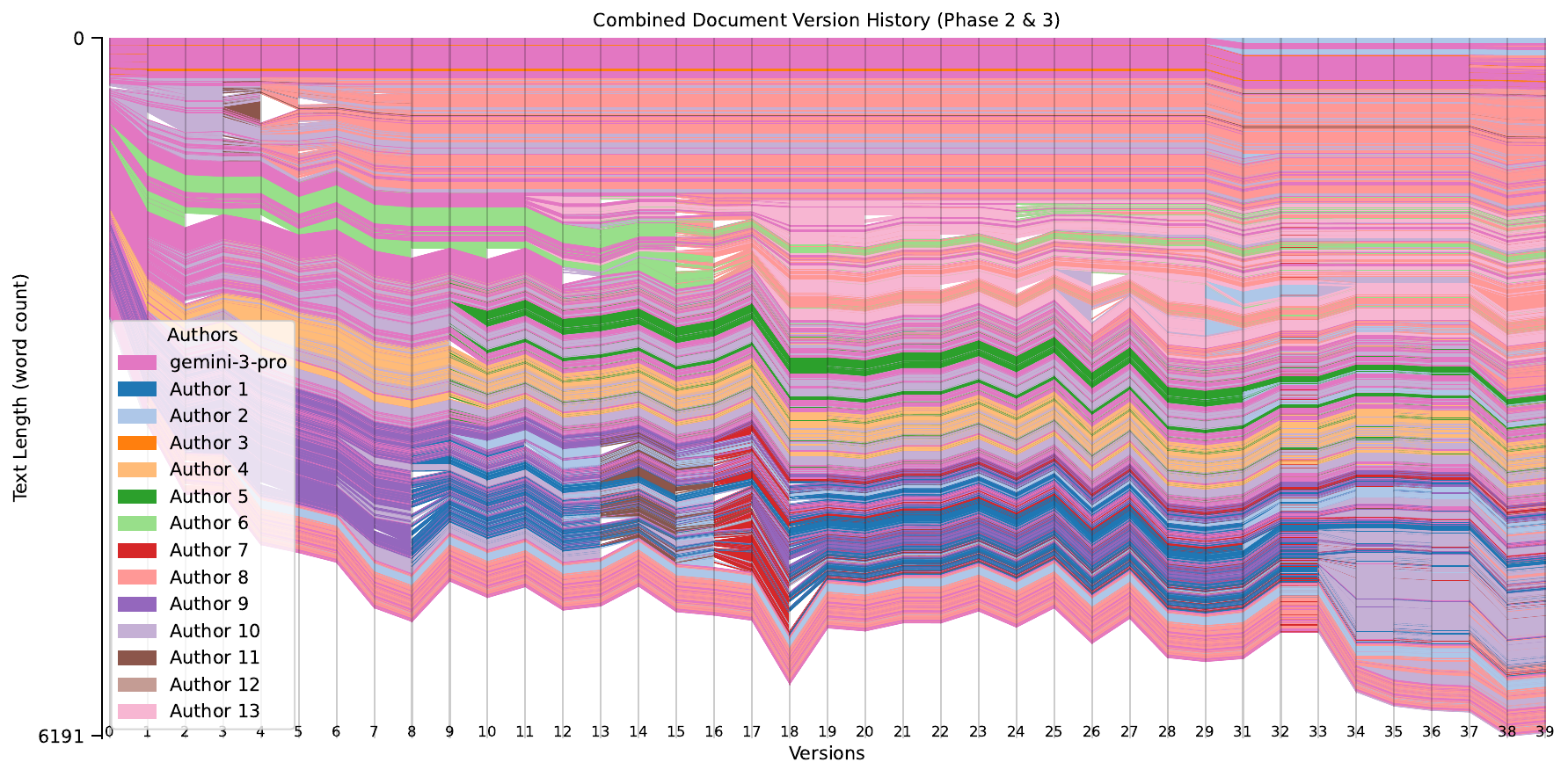}
    \caption{\small The evolution of the AMOC report, visualized as a \emph{history flow}~\citep{Viegas-et-al-2004}, a method introduced to analyze the edit history of Wikipedia articles. We visualize only the main body, excluding references which are section-specific in Phase 2 then unified in Phase 3. Transitions between versions are represented as vertical bands. Edits are attributed via color-coded horizontal bands. The \assistant provided the first version (0). The process was engaging, as shown by the rich weaving of AI and human contributions.}
    \label{fig:history-flow-phases12-anonym}
\end{figure*}
The emergence of AI \emph{co-scientists}, agentic systems capable of autonomous hypothesis generation and experimental design, marks a paradigm shift in scientific discovery~\citep{boiko2023autonomous,gottweis2025}. Current co-scientists are self-evaluating, parallelizable pipelines of Large Language Model (LLM) calls  
that search hypothesis spaces at a massive scale to brainstorm and propose candidate ideas for validation in real-world experiments. 
This approach shines in \emph{verification-rich} tasks such as drug discovery~\citep{gottweis2025} or code optimization~\citep{lu2024aiscientist}, where ground truth is accessible via immediate feedback loops in well-structured repeatable experiments. \citet{Knusel2019} characterize these as ``small problems''. The cost of AI hallucination is low because reality provides immediate error correction. In contrast, some high-stakes research questions in fields such as macroeconomics, epidemiology and climate science are \emph{verification-poor}: controlled experiments are often impossible, systems are non-stationary \citep{lucas1976econometric}, and truth is an authoritative state of knowledge that is arrived at through scientific consensus construction rather than measurement \citep{cleland2002methodological}.\footnote{Most fields span both verification-rich and verification-poor aspects. Climate science combines verification-rich foundations ($\mathrm{CO}_2$ levels, radiative forcing) and verification-poor complexities (e.g., Earth's tipping points) where model dependence and deep uncertainty require integrative assessment rather than verification.}
Consequently, establishing the state of knowledge requires not just calculation, but \emph{assessment}: weighing reinforcing and diverging evidence against background theory to form a consensus \citep{baumberger2017, oreskes2004scientific}. 

Realistically, the deployment of AI in these high-stakes applications is still in its infancy~\citep{Human_AI_DRR}. We argue that trusted full-stack AI co-scientists require their \emph{integration} within the collaborative assessment process, supporting researchers in formulating, refining, and addressing complex research questions under direct expert oversight~\citep{wang2019human}. 
We present an empirical evaluation of such an AI \assistant, deeply integrated into a realistic workflow. The \assistant was stress-tested by renowned climate scientists\footnote{Among the authors of this work.} on a complex topic: the stability of the Atlantic Meridional Overturning Circulation (AMOC), a system of ocean currents that transports heat and nutrients across the Atlantic and contributes to regulating the global climate~\citep{Rahmstorf-Oceanography-2024}. 
Ongoing scientific debate, driven by conflicting evidence from observations, proxy records, and climate models, makes AMOC an especially demanding and informative case study.

We hypothesize that an integrated assistant could enhance the scientific synthesis process by improving: efficiency (accelerating drafting and review), consistency (maintaining a unified voice and evidence assessment), traceability (the ability to easily audit and link conclusions directly to literature, comments, messages and previous versions of the text, adhering to an ``OpenReview principle''), and collaboration (facilitating communication and consensus finding). 

We present this work as a use-inspired empirical contribution, noting upfront that this is a single case study with a proprietary prototype tool. We discuss the implications and limitations of this design in detail in \S\ref{sect:limitations}.
The contributions are threefold: (1)~a new \emph{task formulation}---AI-assisted scientific assessment in verification-poor domains; (2)~simple \emph{metrics} for quantifying human-AI co-authoring dynamics; 
and (3)~a detailed \emph{empirical case study} with 13 domain experts, including a post-study expert survey.
The main findings of our experiment include:
\begin{itemize}[leftmargin=*, topsep=0pt, itemsep=2pt, partopsep=0pt, parsep=0pt]
    \item {\bf A productive hybrid framework}: The \assistant provided a productive environment for the assessment. According to the participating scientists, the final output, an 8,000 words report synthesizing findings from 79 papers, forms a solid basis for a manuscript and assessment outcome that would be suited for scrutiny by scientific peer-review, be it as a journal submission or as part of the preparation process of a scientific assessment report.\footnote{The main report versions are included in the Appendix.}
    \item {\bf High efficiency}: The process demonstrated efficiency and convergence (cf.\ Figure~\ref{fig:history-flow-phases12-anonym}, Figure~\ref{fig:convergence}). 
    Writing the assessment, through 104 versions, took the 13 scientists 46 hours, 33 minutes total. In a post-study survey, all respondents self-reported at least a 2--3x speedup, with 38\% estimating 7x or greater. Participants estimated that comparable tasks require typically 100--200 person-hours.
    \item {\bf Significant AI contributions}: 
    The vast majority of the AI-generated content was retained, after being checked and accepted by the scientists. At times the content was, however, significantly improved in the final version. AI-generated revisions were utilized 3 times more frequently than manual ones, with AI credited for approximately $25\%$ of the revisions content. 
    \item {\bf Expert human oversight is critical}: While AI content is retained, human intervention is critical for scientific rigor. The scientists contributed $58.3\%$ of the final report's content. Human edits were essential to elevate the content from a high-quality general summary to a rigorous scientific synthesis and assessment. 
    \item {\bf Challenges}: We occasionally observed well-known limitations of LLMs involving difficulties synthesizing quantitative data, obsequiousness (especially when pushed adversarially), a lack of holistic reasoning (e.g., identifying caveats), and hallucinations (e.g., reconciling reference numbering). Overall, the negative impact of such behaviors was mitigated by extensive human oversight. Notably, this framework may enhance scrutiny by allowing experts to focus on review rather than drafting.
\end{itemize}

%% file: sections-v2/1b-relatedliterature.tex
\section{Related Work}

The integration of agentic AI into scientific workflows marks a transition from passive data analysis to autonomous knowledge generation \citep{gridach2025agentic, wang2023scientific}. 
The most robust successes of agentic AI have occurred in verification-rich environments, where hypotheses can be rigorously tested against programmatic or physical ground truth. \citet{hartung2025ai} describes this as the shift to ``scAInce'', exemplified by \citet{romeraparedes2024mathematical} using code execution to verify mathematical proofs, and \citet{boiko2023autonomous, bran2024chemcrow} automating chemical synthesis. \citet{lu2024aiscientist} pushed this paradigm to automate the entire lifecycle with \textit{The AI Scientist}. 
Despite advancements, technical challenges persist~\citep{beel2025evaluating_arxiv}. 
It is not yet clear how to extend this architecture to domains where truth is a theoretical consensus rather than an experimental outcome.

In climate science, 'ground truth' is complicated by data friction, the difficulty of constructing global truths from heterogeneous observations \citep{edwards2013vast}. While AI has revolutionized \textit{simulation} (e.g., GraphCast \citep{lam2023learning}, NeuralGCM \citep{kochkov2024neural}), these tools function as dynamical emulators, not epistemic assessors, often biasing collective attention toward data-rich operational questions rather than foundational ones \citep{Hao2026}. These models can predict state evolution using low-dimensional embeddings \citep{moore2025automated}, but lack the explanatory capacity to assess complex phenomena. \citet{baumberger2017} warn that empirical fit does not imply adequacy for long-term projection, mirroring \citeauthor{lucas1976econometric}'s critique that statistical models fail when structural conditions change \citep{lucas1976econometric}. Thus, climate AI requires reasoning architectures \citep{202504.2088} beyond token prediction. Ultimately, the scientific assessment problem is not a peculiarity of climate science, but a defining aspect of the scientific method itself.

In the absence of physical verification, AI relies mostly on textual evidence \citep{zheng2023large, asai2024openscholar}. While systems like DeepScholar \citep{patel2025deepscholar} and automated reviews \citep{cao2025automation} show promise, they introduce specific challenges. \citet{messeri2024artificial} warn of an "illusion of understanding", where fluent explanations mask shallow reasoning. \citet{perez-etal-2023-discovering, sharmatowards} identify "sycophancy", the tendency to align with user beliefs, as a pervasive failure mode. \citet{doshi2024generative} show that generative AI can reduce collective content diversity, potentially creating "echo chambers" \citep{jin2024agentreview, Hao2026} or navigating the "jagged technological frontier" unevenly \citep{dell2023navigating}.

Our collaborative architecture is also motivated by theories of argumentation and Human-AI synergy \citep{Zhao_Liu_Wan_Tang_Li_2025, shao2025sciscigpt}. Evaluating scientific claims requires moving beyond semantic similarity \citep{reimers-2019-sentence-bert} to structural analysis \citep{Faigley-Witte-1981, yang-etal-2017-identifying-semantic}. To this end, our \assistant featured a revision-review module that analyzes a report’s argumentative structure using the schema proposed by \citet{toulmin1958}, alongside other linguistic frameworks \citep{teufel-moens-2002-articles, Mann1988RhetoricalST}.

Analyses of Wikipedia \citep{Viegas-et-al-2004, Kittur-Kraut-2008} suggest that high-quality synthesis can emerge from complex dynamics \citep{yasseri2012dynamics} and reputation mechanisms \citep{Adler-Alfaro-2007}. We position AI not as an oracle, but as a facilitator of collective deliberation \citep{tessler2024ai, summerfield2025impact, duenez2025perceptual}. This aligns with the vision of co-scientists evaluated via rigorous methodologies \citep{cappello2025eaira}.

%% file: sections-v2/2-ai-assistance.tex
\section{The AI Environment}
\label{sect:ai-assistance}

To isolate the impact of the collaborative workflow, we utilize a standard architecture consisting of three main components (see Figure~\ref{fig:editor-screenshot} in the Appendix for a screenshot). We use Gemini, as the LLM, accessed through \href{https://cloud.google.com/vertex-ai}{Vertex AI}.

\paragraph{Editor}
An \emph{editor panel} provides a text editor which accepts Markdown inputs for document formatting, and URL links for references. Edit actions can be manual or executed by AI. Edit instructions for AI are entered as document-level, or text span-level, feedback. The latter is triggered by text highlighting (cf. Figure~\ref{fig:editor-screenshot}). The same feedback mechanism (document or text span based) can be used to message human co-authors or inquire about the edit history of the relevant content.

\paragraph{Evidence}
An \emph{evidence panel} provides a list of documents that are used in the LLM's context.
Entries include a generated summary snippet, bibliographic information (title, authors, venue) and metadata (e.g., citations) retrieved from \href{https://openalex.org}{OpenAlex}. 
For a document to be used as a reference, authors need to add it to the evidence panel first, either by direct upload or via a built-in search function. 
In addition to results from Google Search, we also implement lookup from a collection of around 1,660 scientific papers on the topic of AMOC, that were collected and retrieved in advance. For this \emph{AMOC corpus} we convert papers to Markdown using the LLM and use both BM25~\citep{bm25} and dense retrieval~\citep{karpukhin-etal-2020-dense} on document chunks. See Appendix \ref{sec:rag-settings} for implementation details.

\paragraph{Assistant Chatbox}
A common \emph{chatbox} is available to ask questions to the \assistant. The LLM is rendered 'context-aware' by including in the input a representation of the \assistant \emph{state}: document, references, chat history etc. AI provides assistance in various other forms. When a user logs in, the \assistant provides a welcome message with a summary of the changes and discussion since the previous login (cf. Figure~\ref{fig:editor-screenshot}). The \assistant generates on demand a review of a revision, suggesting ways of improving the text based on rhetorical~\citep{Mann1988RhetoricalST,teufel-moens-2002-articles} and argumentative logic analysis~\citep{toulmin1958}. This is operationalized with simple zero-shot prompts which include the sources as references (\citep{brockriede1960toulmin} for Toulmin's framework).  The \assistant also helps resolving conflicts, proposing merges of concurrent revision candidates, as in code review workflows and version control systems.

\section{Assessment Workflow} 
\begin{table*}[t]
    \centering
    \caption{\small Summary of: lengths of input/output (Length), number of references (References), unique contributing authors (Authors), number of versions excluding the AI draft (Versions) and elapsed time (Time, hours:minutes format). Length is measured in number of words. Length and number of references are given for the draft proposed by Gemini (In), and for the final version curated by the scientists (Out).}
    \label{tab:top-metrics}
    \small
    \setlength{\tabcolsep}{3pt}
    \begin{tabular}{clccccccc}\toprule
        \multirow{2}{*}{\textbf{Phase}} & \multicolumn{1}{c}{\multirow{2}{*}{\textbf{Document}}} & \multicolumn{2}{c}{\textbf{Length}} & \multicolumn{2}{c}{\textbf{References}} & \multirow{2}{*}{\textbf{Authors}} & \multirow{2}{*}{\textbf{Versions}} & {\textbf{Time}}\\
         \cmidrule(lr){3-4} \cmidrule(lr){5-6}
          & & {\bf In} & {\bf Out} & {\bf In} & {\bf Out} & & & \textbf{(h:m)} \\\midrule
         {\bf1} & Outline   & 798 & 2107 & 0 & 0 & 8 & 16 & 05:17\\ \cmidrule{1-9}
         \multirow{6}{*}{{\bf2}} & 1. Introduction & 501 & 526 & 5 & 6 & 1 & 2 & 00:30 \\
          & 2. The Physical Science Basis of AMOC & 410 & 1400 & 3 & 14 & 3 & 9 & 03:30 \\
          & 3. Evidence for Variability and Slowdown & 675 & 2376 & 7 & 31 & 4 & 30 & 14:48 \\
          & 4. The Tipping Points Debate & 606 & 1662 & 3 & 13 & 4 & 12 & 04:28 \\
          & 5. Impacts, Risks and Vulnerabilities & 785 & 1345 & 7 & 13 & 6 & 20 & 09:47 \\
          & 6. Conclusion and Priorities & 498 & 960 & 7 & 18 & 3 & 5 & 00:45\\\cmidrule{2-9}
          & Total Phase 2      & 3475 & 8269 & 26 & 79 & 13 & 78 & 33:48\\\cmidrule{1-9}
         {\bf3} & Full Report & 7787 & 7918 & 79 & 67 & 4 & 10 & 07:28 \\\cmidrule{1-9}
         {\bf All} & Outline+Sections+Full Report & 798 & 7918 & 0 & 67 & 13 & 104 & 46:33\\\bottomrule
    \end{tabular}
\end{table*}
The experiment was designed to approximate key features of IPCC-like assessments, including phased scoping, drafting, and iterative revision under explicit time constraints. The experiment took place over 5 weeks between November and December 2025. It involved a deliberately heterogeneous group of 13 climate scientists, spanning diverse sub-fields, with extensive experience as leads on IPCC and other institutional reports, and with varying levels of prior expertise on the Atlantic Meridional Overturning Circulation (AMOC), only a minority of whom had previously published directly on the topic. All 13 experts participated in writing --- rather than a smaller core team with separate reviewers --- to approximate the real IPCC conditions where diverse interdisciplinary skills contribute directly to the assessment text, and where a sufficiently diverse group mitigates individual bias.
The process was executed in three \emph{phases}. Table~\ref{tab:top-metrics} reports high-level statistics about the process.

\paragraph{Phase 1 (Outline):} The \assistant generated an outline for a 10-page scientific assessment on the stability of AMOC. Afterwards, the scientists edited the draft to produce the final outline and assigned themselves to specific sections. In Phase 1, we used Gemini 2.5 Pro as the LLM. Figure~\ref{fig:history-flow-outline} plots the history flow for the development of the outline.

\paragraph{Phase 2 (Sections):} Before Phase 2, we transitioned to the newly released Gemini 3 Pro Preview, due to its superior performance across multiple benchmarks. We note this model switch as a potential confounding factor when estimating efficiency,\footnote{For analysis of AI contributions we report results for Phase 1 and 2+3 separately and focus on only Phase 2+3 for convergence assessment, avoiding the model change as a confounding factor.} though we believe its impact is limited: Phase~1 consumed only $\sim$11\% of total time (5h17m of 46h36m) and was dominated by human deliberation on report structure rather than model capability; furthermore, in a post-study survey, 69\% of participants perceived no meaningful difference in AI capabilities after the transition. 
Beyond the efficiency gains, we report separate results for outline vs. full report phases. The expert-curated outline from Phase 1 was expanded by the \assistant into a full-length report draft, then segmented into sections. Each section was grounded in the full AMOC corpus with a pipeline that evaluated all documents from the corpus, iteratively in batches. In each iteration, the \assistant identified the most relevant sources, replacing previous selections if more relevant ones were found. Finally, the \assistant revised the section text to incorporate in-text citations and references. The scientists then took over the drafts to curate, in groups, each individual section of the report.

\paragraph{Phase 3 (Final Report):} In the last stage, the \assistant produced a new draft for the complete report by collating the different sections and producing a single coherent list of references and citations. A subset of the scientists then performed a final pass to improve the presentation, resolve any outstanding issues, check and sign off the content. As the very last step the \assistant performed an automatic reference check.

%% file: sections-v2/3-results.tex
\section{Results}
\label{sect:results}

\begin{figure}
    \centering
    \begin{subfigure}[b]{0.45\textwidth}
        \centering
        \includegraphics[width=\textwidth]{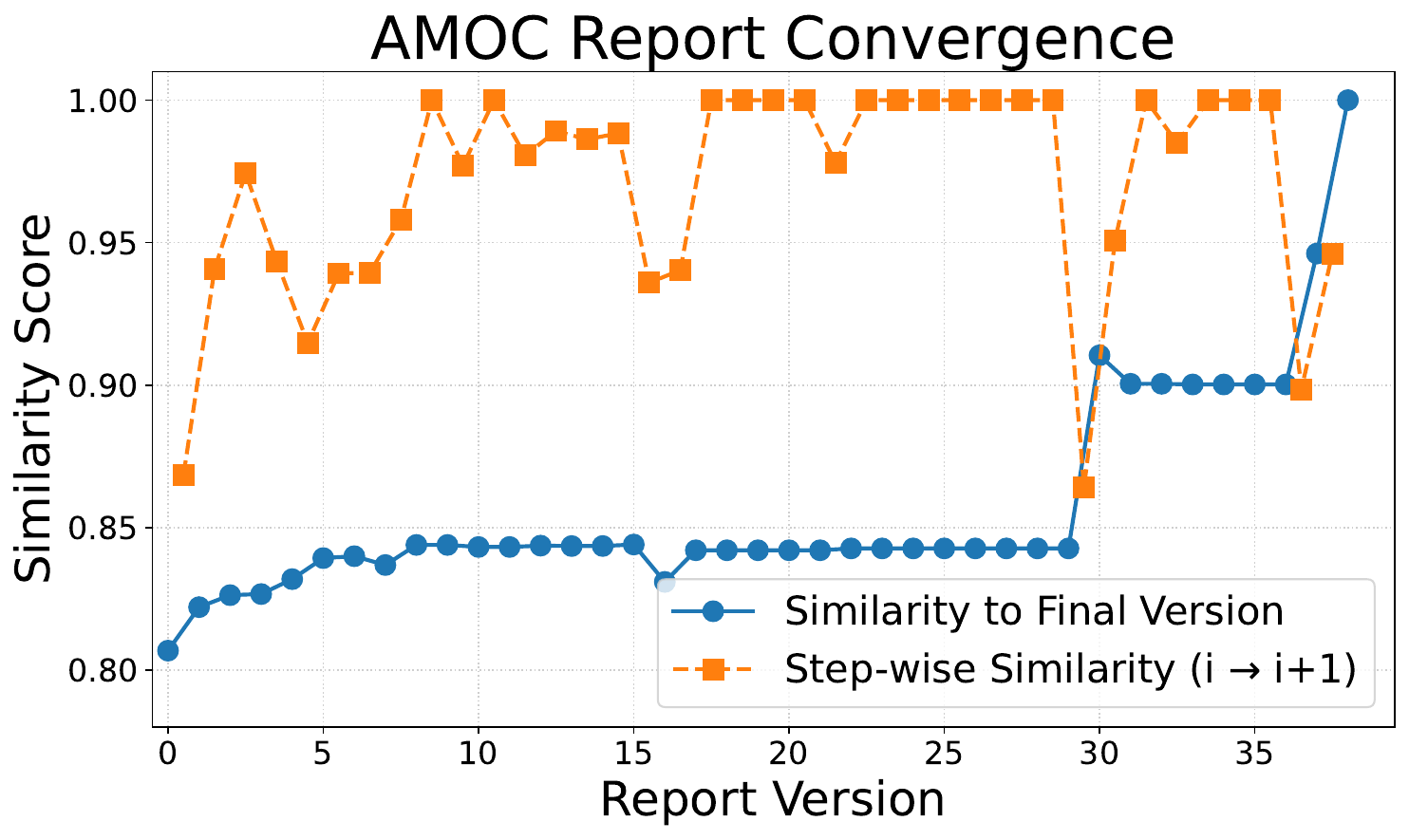}
         \caption{\small Document similarity scores between successive versions of the AMOC report (Step-wise), and document similarity between each version and the final version of the report (Similarity to final).}
    \label{fig:convergence}
    \end{subfigure}
    \hfill 
    \begin{subfigure}[b]{0.50\textwidth}
        \centering
        \includegraphics[width=\textwidth]{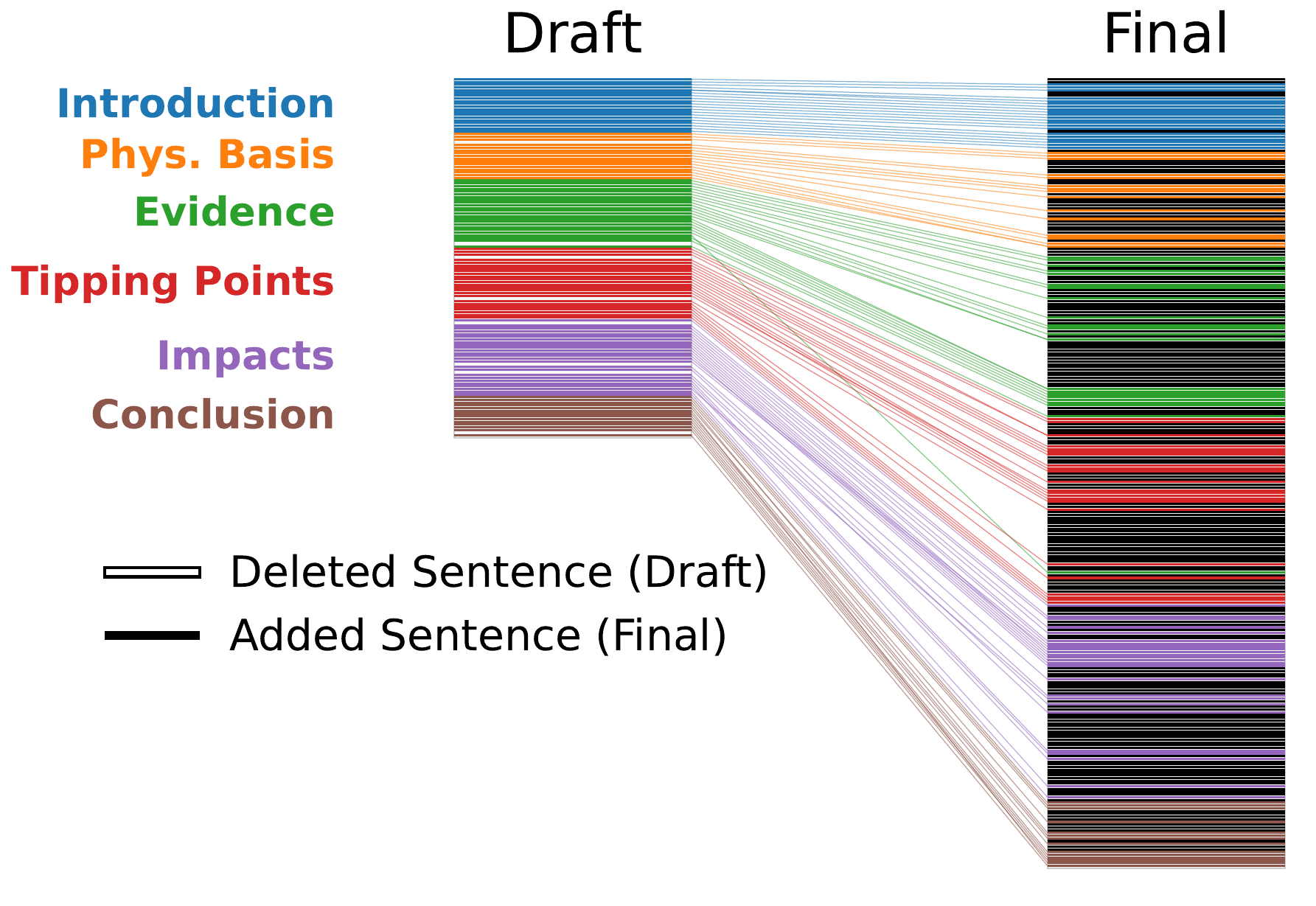}
            \caption{Visualization of the sentence alignment.}
    \label{fig:alignment}
    \end{subfigure}
    
   \caption{Report convergence measurements and sentence alignment visualization.}
\end{figure}

\subsection{Convergence}
We expect that an effective consensus-based process will lead to an asymptotic convergence toward the final version of the document. We evaluate convergence by computing text similarity metrics for each version. Specifically, we generate document embeddings\footnote{We use \texttt{\href{https://ai.google.dev/gemini-api/docs/embeddings}{gemini-embedding-001}} with task type \texttt{SEMANTIC\_SIMILARITY}} and calculate the bounded inverse Euclidean distance ($1/(1+d)$) between each intermediate version and the final report.\footnote{Cosine similarity produces the same pattern but values are contained within a smaller interval near 1.0.}
We also compute the similarity between contiguous (step-wise) versions.
Figure~\ref{fig:convergence} illustrates the results. Similarity to the final report converges almost monotonically. Three distinct stages emerge: 1) During the first 9 versions, the report undergoes rapid changes. 2) The report enters a refinement stage characterized by smaller local changes. 3) A final stage begins with the transition to the pre-final version. A period of fine-tuning then follows until completion. Three roughly similar stages are also visible in the history flow plot of Figure~\ref{fig:history-flow-phases12-anonym}, in that case as a function of the document length.

\subsection{Efficiency}
During the experiment, the 13 authors, in various configurations, iterated over 104 versions.\footnote{In Phase 2, 6 threads are developed in parallel, one per section. When combined in a single version history, they form 29 full versions of the report from Phase 2. Including the 10 versions from Phase 3 (and excluding the outline), the total is 39 report versions (cf. Figure~\ref{fig:history-flow-phases12-anonym} and Figure~\ref{fig:convergence}).} The outline length almost tripled in Phase 1, while the report sections length doubled in Phase 2, and marginally increased in the final pass. The number of references grew from 26, in the AI-generated draft of Phase 2, to 79 at the end of Phase 2, and finally to 67 at the end of Phase 3.
Table~\ref{tab:top-metrics} reports logged-in \emph{elapsed time}. This metric should be interpreted as an approximation of time spent on the task rather than a precise measure of active work time. While we apply a timeout parameter (900 seconds) we may sometimes underestimate idle time. Conversely, some activities outside of the tool, like offline proofreading, may not be accounted for. 
The total time spent writing and curating the report in the assistant is 46 hours, 33 minutes: $11\%$ of the time on Phase 1, $73\%$ on Phase 2 and $16\%$ on Phase 3, approximately 3.5 hours per author on average. \S\ref{sect:findings} puts these numbers in context.

\subsection{AI Contribution Metrics}
\label{sec:metrics}

To evaluate the \assistant's contribution, we analyze text at the sentence level. The sentence is a linguistic universal that captures a complete thought and is a meaningful proxy for thematic units. 

Let $D$ be the list of sentences in the initial version proposed by the \assistant, the \emph{draft}. Let $F$ be the list of sentences in the \emph{final}, expert-curated version: the reference. We define an alignment $\mathcal{M}$ as a set of matching sentences:
\begin{equation}\nonumber \mathcal{M} = \{ (i, j) \mid \text{Sentence } d_i \text{ is aligned with sentence } f_j \} \end{equation}
Semantically, we define a \emph{match} as the sentence $f_j$ expressing a theme or concept also expressed in sentence $d_i$. To account for many-to-many relationships (e.g., due to synthesis or expansion), we define alignments for both versions: 
\begin{align}\nonumber D_{\mathcal{M}} &= \{ d_i \in D \mid \exists f_j \in F : (i, j) \in \mathcal{M} \} \\ \nonumber F_{\mathcal{M}} &= \{ f_j \in F \mid \exists d_i \in D : (i, j) \in \mathcal{M} \} \end{align} 

We define two simple metrics, adapted from well-established, reference-based, evaluation metrics for language prediction tasks such as machine translation, BLEU~\citep{papineni2002}, and summarization, ROUGE~\citep{lin-2004-rouge}. This is conceptually similar to other human-AI co-authoring frameworks~\citep{coauthor2022}.

\noindent\textbf{Retention:} Similarly to BLEU, \emph{Retention} is a precision metric. It measures the fraction of content proposed by AI that is still thematically present in the reference, expert-curated, version:
\begin{equation} \nonumber\text{Retention [i.e., Precision, or Positive Predictive Value]} = \frac{|D_{\mathcal{M}}|}{|D|} \end{equation} 
\noindent\textbf{Intervention:} Similarly to ROUGE, \emph{Intervention}, is a recall-oriented metric. It measures the amount of reference content that is new human work, with no thematic correspondence to the AI draft:
\begin{equation}\nonumber \text{Intervention [i.e., False Negative Rate, or 1-Recall]} = 1 - \frac{|F_{\mathcal{M}}|}{|F|}\end{equation}
Together, Retention and Intervention characterize the two complementary dimensions of human-AI co-authoring: the \emph{utility} of the AI draft (high Retention is better) and the extent of additional expert \emph{effort} beyond it (low Intervention is better). We view these metrics as an incremental, stepping-stone proposal. Exploring richer measures for conceptual influence, different forms of logical and rhetorical elaboration are important directions for future work.

To compute the sentence alignment between versions, we used Gemini. Sentence segmentation for well-edited formal text is an easy task for most languages, including English. Sentence alignment is also a deliberately simple task for this data, for which simple methods can be expected to perform well, minimizing the risk of bias. This is true particularly for LLMs that can effectively leverage self-attention, thus considering all possible cross-version and intra-version dependencies. One of the authors manually inspected 100\% of the aligned pairs and unaligned segments and did not find mistakes. Figure~\ref{fig:alignment} plots the output of the alignment process.

As an additional validation, we implemented an independent open-source fully reproducible baseline, see Appendix \ref{sect:retention_baseline}, using sentence embeddings and cosine similarity for alignment. The directional findings of this baseline, see Figure \ref{fig:retention_intervention_baseline}, are consistent with the LLM-based method, confirming high AI content retention and substantial expert intervention. 
The following example demonstrates how aligned (retained) sentences can be further elaborated by the scientists:

\noindent\textbf{Draft:} "3. Theory: Refine the theoretical application of Early Warning Signals to stochastic climate data to reduce false positives and negatives [24], [18]."

\noindent\textbf{Final:} "3. Theory: Move beyond purely statistical Early Warning Signals, which are prone to false positives [47], by developing and monitoring physics-based indicators, such as the freshwater transport at the southern boundary of the Atlantic [49], [66]."

\begin{table}[t]
    \centering
    \setlength{\tabcolsep}{2pt}
    \caption{\small Sentence alignment statistics and Retention and Intervention scores for Outline and Report.}
    \label{tab:scores}
    \begin{tabular}{l rr rr rr} 
        \toprule
        & \multicolumn{2}{c}{\textbf{Sentences}} & \multicolumn{2}{c}{\textbf{Matches}} & \multicolumn{2}{c}{\textbf{Scores (\%)}} \\
        \cmidrule(lr){2-3} \cmidrule(lr){4-5} \cmidrule(lr){6-7}
        \textbf{Dataset} & \textbf{Draft} & \textbf{Final} & \textbf{Draft} & \textbf{Final} & \textbf{Retention} & \textbf{Intervention} \\ 
        \midrule
        \textbf{Outline} & 82  & 121 & 81  & 76  & 98.8 & 37.2 \\
        \textbf{Report}  & 131 & 288 & 123 & 120 & 93.9 & 58.3 \\
        \bottomrule
    \end{tabular}
\end{table}

\subsection{Evaluation of AI Contribution}
\label{sect:findings}
We compute the AI Contribution metrics for the outline (Phase 1) and for the full report (Phase 2 + Phase 3). For the latter, we compare the \assistant version prepared at the beginning of Phase 2, against the very final version of the report produced in Phase 3. Table~\ref{tab:scores} reports the results. The four versions are copied in the Appendix: Outline Draft (\ref{sect:outline:draft}), Outline Final (\ref{sect:outline:final}); Report Draft (\ref{sect:report-draft}), Report Final (\ref{sect:report-final}).
Retention is near-perfect for both outline ($98.8\%$) and report ($93.9\%$), indicating that the AI draft content is of high quality and largely preserved by the experts.
At the same time, Intervention is substantial for the outline ($37.2\%$) and high for the report ($58.3\%$) where the experts contributed the majority of the final report's content.

While the draft is a solid high-level summary, the final version of the paper is a more rigorous, complete, evidence-based scientific assessment. Retained sentences in the final version sometimes replace generic, colloquial or emotive phrasing with neutral and technical descriptions.  
The edited sentences use a more nuanced language and articulate uncertainty more transparently and consistently. 
In the two most consequential corrections to the draft, the scientists contradicted the AI statement about 'full scientific consensus' on the weakening of AMOC and downgraded the confidence level on the anthropogenic attribution of the weakening of AMOC from 'Medium' to 'Low'. These edits demonstrate that the final report is not merely an expanded version, but a substantively revised scientific assessment.
In terms of intervention, the scientists incorporated findings from recent research that depict a more nuanced and sometimes contradictory picture of AMOC's stability. For example, discussing evidence missed by the \assistant draft, from air-sea heat flux and hydrographic data that suggest the circulation has remained more stable than previously thought. In these cases, human experts embraced complexity while the \assistant seemed to smooth over conflicts to present a unified narrative. Appendix~\ref{sect:example-sentences}, lists several examples. 

We recorded 1,151 actions in the \assistant. Experts revised the text, exchanged feedback with AI and co-authors and worked with sources (cf. Table~\ref{tab:action-by-section}). AI-generated revisions were preferred over manual edits. They account for $\sim$80\% of all revisions and $\sim$75\% of submitted versions. Text span-level feedback is the most frequent interaction type ($33\%$).
To provide additional exploratory context on the workflow dynamics, we used Gemini to categorize the logged interactions and to estimate revision attribution across the 104 versions. These results are based on AI self-evaluation and cannot be easily  validated; so they should be interpreted as providing exploratory context. The full analysis and detailed tables are reported in Appendix~\ref{sec:appendix-interactions}. In summary, an estimate of revision attribution credits AI for $25\%$ of the revision content (cf.\ \S\ref{sec:attribution-example} for an example). Chat interactions covered a broad range of topics, from scientific review and critique to source management and tool help. 

The increase in number of references was accompanied by a qualitative improvement in source selection and relevance. The final list of references includes foundational sources, and introduced a broader, more balanced, set of evidence as well as increasing the technical depth of the literature.

%% file: sections-v2/4-discussion.tex
\subsection{Post-Study Expert Survey}
\label{sect:survey}
To better understand the results, we ran a survey with all the  scientists. The full survey and responses are provided in the Appendix~(\ref{sec:survey-instrument},~\ref{sec:survey-responses}, Table~\ref{tab:survey-full}). Table~\ref{tab:survey-results} summarizes the key findings. We highlight the diversity and breadth of opinions (cf. \ref{sec:survey-qualitative}) which is typical of real assessments and adds credibility to the study.

The majority of experts (8) considered the report a solid foundation requiring minor polish, while 2 thought it needs more work.\footnote{A full closure would require the specific requirements of an actual venue, which is beyond the current scope of this work.}
On efficiency, at the lower end, 54\% of participants estimate a 2--3x efficiency gain, while at the high end 31\% estimate more than 10x. Free text retrospective estimates of the time required for a comparable task without AI are consistent, ranging from 100 to over 200 person-hours. \citet{Allen1999} report a median of 1,139 hours for systematic meta-analyses, and \citet{Borah2017} report a mean of 67.3 weeks for systematic reviews. While not directly comparable, they provide consistent order-of-magnitude references. 
Importantly, despite usability limitations of the prototype, the majority reported lower cognitive burden by using the \assistant. It seems plausible that more polished tools would further improve efficiency and reduce cognitive load. 

\begin{table}[tbp]
\centering
\small
\caption{Summary of the post-study survey results on key dimensions.}
\label{tab:survey-results}
\begin{tabular}{@{} l l l @{}}
\toprule
\textbf{Dimension} & \textbf{Distribution} & \textbf{Key Finding} \\
\midrule
\textbf{Report Quality} & Satisfied: 62\% $\cdot$ Neutral: 23\% $\cdot$ Needs More Work: 15\% & Solid basis, not final \\
\textbf{AI Speedup} & 2--3x: 54\% $\cdot$ 4--6x: 8\% $\cdot$ 7--10x: 8\% $\cdot$ ${>}$10: 31\% & All reported $\geq$2x \\
\textbf{Cognitive Burden} & Lower: 62\% $\cdot$ Higher: 31\% $\cdot$ Same: 8\% & Majority report relief \\
\textbf{Assistant Usability}& Satisfied: 54\% $\cdot$ Neutral: 23\% $\cdot$ Dissatisfied: 23\% & Mixed perceived usability\\ 
\textbf{AI Anchoring} & Neutral: 46\% $\cdot$ Agree: 31\% $\cdot$ Disagree: 23\% & No majority concern \\
\textbf{Trust in AI} & Partial: 69\% $\cdot$ None: 31\% & Zero ``full trust'' \\
\textbf{Sycophancy} & Observed: 62\% $\cdot$ Unsure: 15\% $\cdot$ No Sycophancy: 23\% & Known LLM limitation \\
\bottomrule
\end{tabular}
\end{table}

A concern in human-AI co-authoring is that the AI's initial draft may \emph{anchor} the experts' reasoning, limiting the diversity of perspectives considered~\citep{shaw2026thinking}, narrowing focus and hurting exploration~\citep{barrett2024,Hao2026}. Our survey found no majority concern (31\% agree vs.\ 23\% disagree).
Hybrid environments that facilitate the collaboration in large and diverse groups can help mitigate the risks of anchoring and intellectual homogenization. In our experiment it is clear that scientists did not merely edit the \assistant's work; they expanded and improved the \assistant's proposals, adding most of the final report's content and the number of references nearly tripled. This indicates that the \assistant draft served as a starting point, but did not limit the assessment. At the same time, known limitations of LLMs, such as sycophancy~\citep{perez-etal-2023-discovering, sharmatowards}, limit the degree of trust in AI.
Participants also reported AI's difficulties synthesizing quantitative ranges from multiple sources and occasional lack of comprehensiveness.

Participants expressed occasional frustrations with the prototype, citing "clunkiness", slow response times, and the tendency to merely quote, rather than genuinely synthesize. Despite the limitations, the overall sentiment was positive. Experts praised the \assistant's ability to locate objective evidence, noting that it allowed rapid cross-checking and made the review process "quite fun" by surfacing unexpected papers. Several expressed a desire to continue using the tool, noting that this initial experiment provided a compelling "glimpse of what might be possible" (cf. \ref{sec:survey-qualitative}).

%% file: sections-v2/5-conclusion.tex
\section{Limitations and Broader Impact}
\label{sect:limitations}
Our findings are based on a single domain (climate science), a single topic (AMOC), and a single group of 13 (top) domain experts. Our deliberate choice of maximizing domain expertise reduced our flexibility in experiment design. 
Running a single experiment, although complex, means that no formal control conditions could be assessed. These should include, for example, a human-only baseline and evaluate the role of variables such as team composition, domain knowledge and workflow structure. Scaling experiment capacity when expertise is necessary but scarce remains an open challenge.
The transition from Gemini 2.5 Pro to Gemini 3 Pro Preview between Phases~1 and~2 introduces also a potential confound. However, Phase~1 consumed only $\sim$11\% of total time and 69\% of survey respondents perceived no difference. That said, this confound cannot be analytically resolved, and this is a challenge of working with frontier models.
The qualitative findings and interaction patterns are expected to generalize, but exact replication is not possible. We provide all major artifacts of the study for independent inspection in the Appendix.

More broadly, the deployment of AI in high-stakes scientific assessment raises societal concerns that go beyond standard LLM risks. For example, errors in AI-generated assessment text, if insufficiently scrutinized, could propagate into policy-relevant documents with material consequences for climate adaptation and mitigation planning. AI may also lend an appearance of comprehensive rigor to insufficiently vetted claims, particularly when deployed at scale. More research on safeguards beyond human review, such as automated traceability, grounding checks, and transparent provenance tracking, is needed. The participating scientists identified evidence traceability, numeracy verification, and unbiased literature selection as the most critical safeguards.

\section{Conclusions}
The distinction between scientific \emph{review} and \emph{assessment} is central to understanding our findings. The \assistant excelled at synthesizing and summarizing the literature (the ``What''~\citep{Borton1970}). But human expertise remained essential for evaluating evidence quality, articulating uncertainty, and forming scientific judgment (the "So What"). This pattern is consistent with a productive \emph{cognitive offloading} workflow~\citep{shaw2026thinking}, where experts strategically delegate mechanistic sub-tasks (drafting, literature retrieval, cross-referencing, consistency checks) to AI while retaining deliberative control over scientific synthesis and revision.
Our findings suggest that the scope of AI assistance can extend to verification-poor tasks, where an authoritative state of knowledge is constructed through expert consensus when needed to inform high-stakes decisions (the "Now What").

Hybrid intelligence architectures are central to this paradigm: while AI is highly effective at information processing, human expertise remains essential for ensuring scientific rigor and nuance. This augmentative approach enables expert groups to accelerate research on long-horizon problems. Furthermore, embedding AI within expert workflows creates a feedback loop that may facilitate the acquisition of tacit process knowledge characteristic of advanced research, and the ability to integrate conflicting viewpoints by AI. Moving forward, we envision these environments playing a central role in the development of trusted full-stack co-scientists. 

Our verification-poor framework opens several concrete research directions for AI-for-Science: (1)~LLM-as-Judge systems trained on expert interaction traces from collaborative assessments; (2)~simulation baselining using AI agent groups as experimental controls, enabling counterfactual comparisons without human participants; (3)~longitudinal objective metrics such as policy integration speed and citation impact of AI-assisted assessments; and (4)~community protocols and shared metrics for cross-study comparison in human-AI co-authoring research. We view this work as a first step toward a systematic empirical agenda for AI in verification-poor domains.

\section*{Acknowledgments}
We thank Taylan Cemgil, Fernanda Viegas, Martin Wattenberg, Anna Koivuniemi, Ashley Thomas, Peter Battaglia, Ben Gaiarin and Elisa Rawat for feedback and support.

%% file: sections-v2/appendix.tex
\section{Document retrieval}
\label{sec:rag-settings}

We build a simple RAG system for retrieval from our pre-compiled corpus of AMOC papers. 
Starting with a collection of \texttt{.pdf} files we use an LLM\footnote{All corpus processing is done using \texttt{gemini-3-pro-preview}} to extract the DOI\footnote{\href{https://www.doi.org/}{https://www.doi.org/}} and title and use that information to retrieve metadata from OpenAlex.\footnote{\href{https://openalex.org/}{https://openalex.org}, License: \texttt{CC0}} We also use the LLM to generate a markdown representation of the document. This text-based version is segmented into 200 word chunks with 20 words of overlap between chunks. The chunks are embedded using \texttt{gemini-embedding-001}~\citep{geminiembedding2025} with task-type \texttt{RETRIEVAL\_DOCUMENT}\footnote{\href{https://docs.cloud.google.com/vertex-ai/generative-ai/docs/embeddings/task-types}{https://docs.cloud.google.com/vertex-ai/generative-ai/docs/embeddings/task-types}} yielding $3072$ dimensional embeddings. For retrieval, we embed queries using task-type \texttt{RETRIEVAL\_QUERY} and cosine similarity as a distance metric.

In addition to dense retrieval, we also use a classic retrieval engine
to retrieve markdown versions of the documents with BM25~\citep{bm25} similarity.

\section{Assistant Interface}
\begin{figure*}
    \centering
    \includegraphics[width=0.99\linewidth]{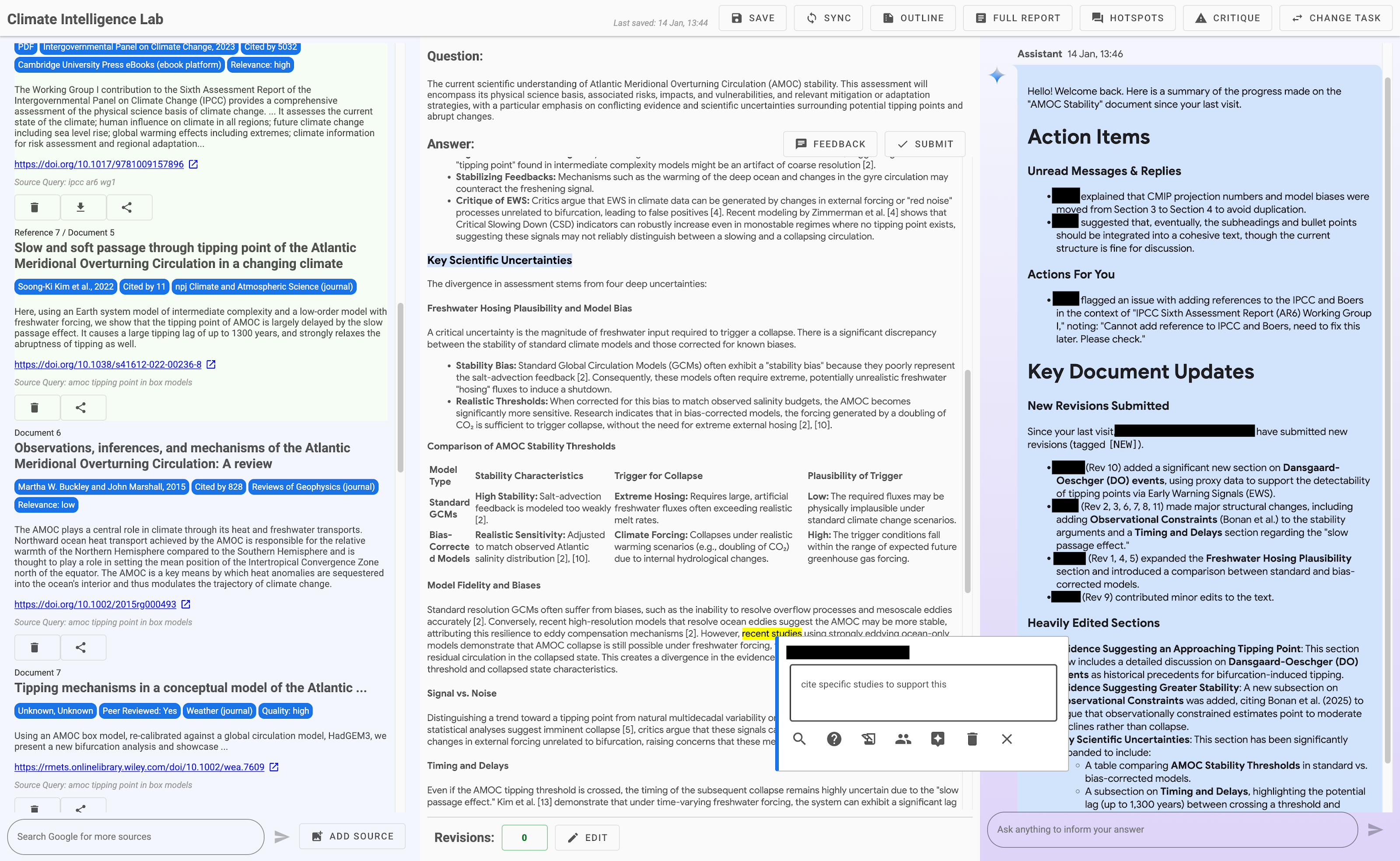}
    \caption{\small A screenshot of the \assistant user interface. The highlighted text triggers a pop-up message box that can be used to send feedback to co-authors, or to the Gemini assistant, to generate a trace of the content, to inspect how it originated and evolved over time, or to ask the \assistant about the content in general. The screenshot also highlights the 'Welcome Back!' message, which updates the users about what happened since their last session. User names have been redacted for anonymity.}
    \label{fig:editor-screenshot}
\end{figure*}
Figure~\ref{fig:editor-screenshot} shows a screenshot of the Assistant interface.

\section{Interaction Examples}
See Table \ref{tab:feedback-examples} for examples of feedback on the content directed to AI, Table \ref{tab:messages-examples} for examples of messages on the content directed to human(s), and Table \ref{tab:chat-examples} for examples of messages to the \assistant in the chat panel.
\begin{table}[ht]
    \centering
    \caption{\small Examples of feedback on the Content directed to AI.}
    \label{tab:feedback-examples}
    
    \renewcommand{\arraystretch}{1.5}
    
    \begin{tabular}{l p{10cm}}
        \hline
        \textbf{Feedback Type} & \textbf{Concrete Example} \\
        \hline
        
        Epistemological & ``for each of the following sub-points we need to consider the lines of evidence on which the insights are based (models, paleo proxies, ...) and their fit for purpose'' \\
        
         & ``Be more explicit on how a shifting ITCZ would threaten food security. The current formulation is too vague. ITCZ may refer to more precipitation in one region and less in another one. Without context it is unclear how this would affect food security.'' \\
        \hline
        
        Presentation & ``Eliminate the numbering here and make a single paragraph on the structure and pathways.'' \\
        \hline
        
        Sources & ``include document 6 as a reference for the term ``conveyor''\,'' \\
        
         & ``When referring to IPCC AR6 Wg1, cite the report itself (document 11)'' \\
        \hline
        
        Tone & ``\,`alarms' is not really appropriate for such a paper. Suggest expressing in a neutral way, e.g. `have argued'\,'' \\
        \hline
        
    \end{tabular}
\end{table}

\begin{table}[ht]
    \centering
    \caption{\small Examples of messages on the Content directed to Human.}
    \label{tab:messages-examples}
    
    \renewcommand{\arraystretch}{1.5}
    
    \begin{tabular}{l p{10cm}}
        \hline
        \textbf{Message Type} & \textbf{Concrete Example} \\
        \hline
        
        Epistemological & ``A risk framing should be adopted here. Risk is a function of hazard, exposure, vulnerability, and response.

This whole section feels ad hoc and bizarre. Why are vulnerabilities separated out from key risks. It would be better to structure this around key risks. The categories considered for vulnerabilities and risks are only partial. Social cost of carbon as the only economic connection is weak. Etc. etc.'' \\
        
         & ``contrast this more with the Northern Amazon/Sahel drying to emphasize the dipole shift'' \\
        \hline
        
        Presentation & ``It is confusing to me that key findings are being asserted at this stage. They should arise from our careful assessment and be the last component that we develop. We should not be prejudging the findings we develop.'' \\
        \hline
        
        Sources & ``I have included two new sources on the role of mountains to spark some discussion. Do we think this is worth including? If not, what's the reason to remove it?'' \\
        \hline
        
        Tone & ``This section still remains extremely weak. As I have tried to say many times: don't be prescriptive.'' \\
        \hline
        
    \end{tabular}
\end{table}

\begin{table}[ht]
    \centering
    \caption{\small Examples of messages to the Assistant in the Chat panel.}
    \label{tab:chat-examples}
    
    \renewcommand{\arraystretch}{1.5}
    
    \begin{tabular}{l p{10cm}}
        \hline
        \textbf{Message Type} & \textbf{Concrete Example} \\
        \hline
        
        Review and Critique & ``Does section 5 seem balanced and comprehensive or is anything missing?'' \\
        
         & ``Do you spot any significant differences in the assessment of the AMOC relative to the current report? Has science generated new results since AR6 that are crucial to be mentioned here?'' \\
        \hline
        
        Request Content Revision & ``please review your suggestion. Is this really less vague? Isn't is still generic? Please try to make a better job, while still sticking tightly to literature.'' \\
        \hline
        
        Manage Sources & ``I've added quite a few sources. I am asking you to a) go over the whole current version of the section to check whether this needs updates/whether there are inconsistencies. and b) describe the AMOC evolution and variability in climate models/create a draft for that section.'' \\

        & ``do you have access to "Atlantic meridional heat transports computed from balancing Earth's energy locally". Is it consistent with the cited literature and can it be added?'' \\
        \hline
        
        Get Help with the Tool & ``How to search and add more documents from the existing library to the sources?'' \\
        \hline

        Ask for Information & ``How large was the freshwater forcing for the 8.2ka event compared to what is expected in the near future?'' \\

         & ``is there an evidence in "Monitoring the Multiple Stages of Climate Tipping Systems from Space: Do the GCOS Essential Climate Variables Meet the Needs?" ref, which states the AMOC is currently in its weakest state?'' \\
        
        \hline

        Retrieve Change History & ``Getting change history for Long-term contribution to global sea-level rise from thermal expansion changes in the deep ocean...'' \\
        \hline

    \end{tabular}
\end{table}

\section{AI vs Human Attribution Example}
\label{sec:attribution-example}
\begin{figure*}[!h]
    \centering
    
    \begin{minipage}[t]{0.48\textwidth} 
        \centering
        \includegraphics[width=\textwidth]{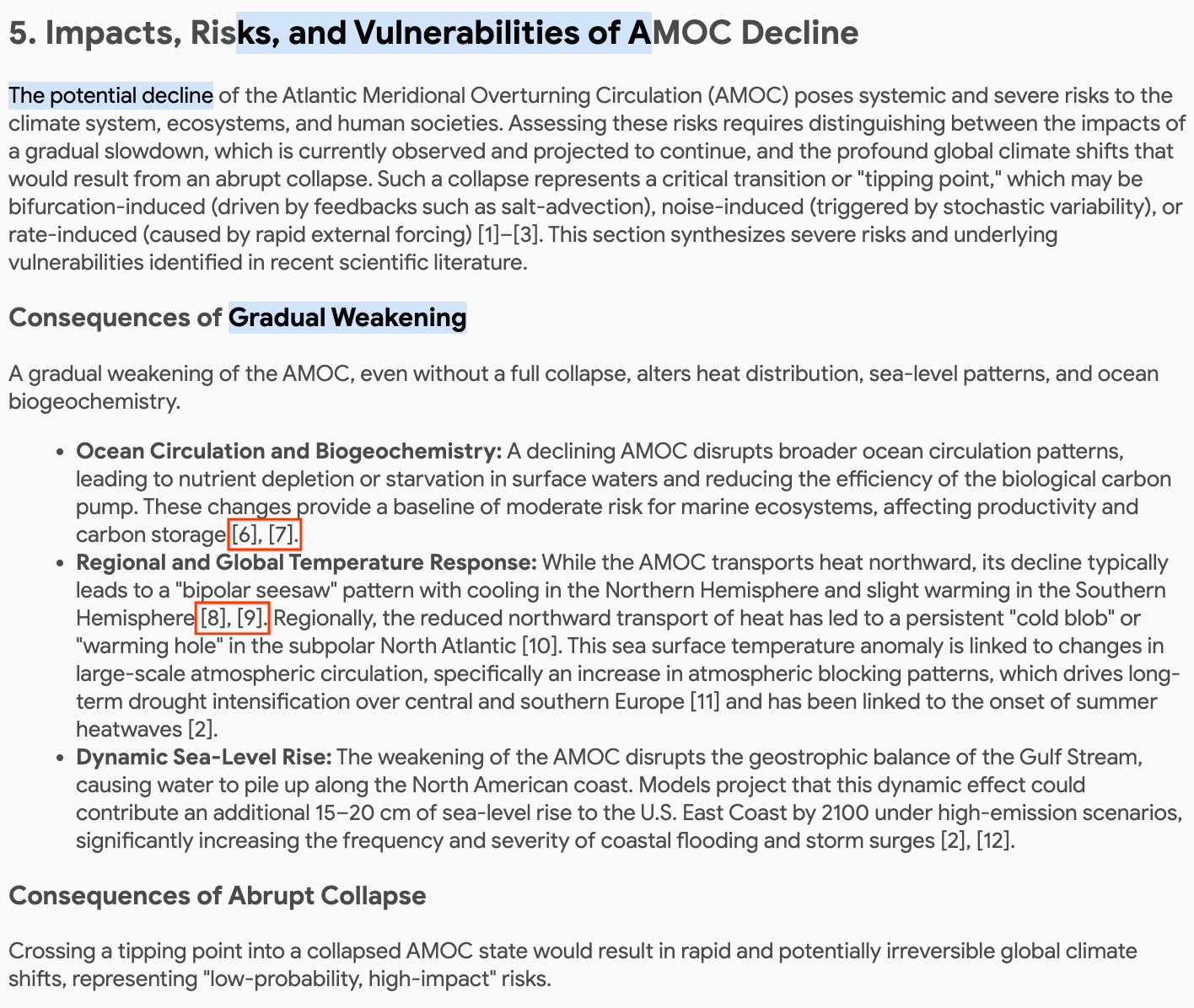}
        \par \vspace{-2pt} $\vdots$ \vspace{-2pt} \par
        \includegraphics[width=\textwidth]{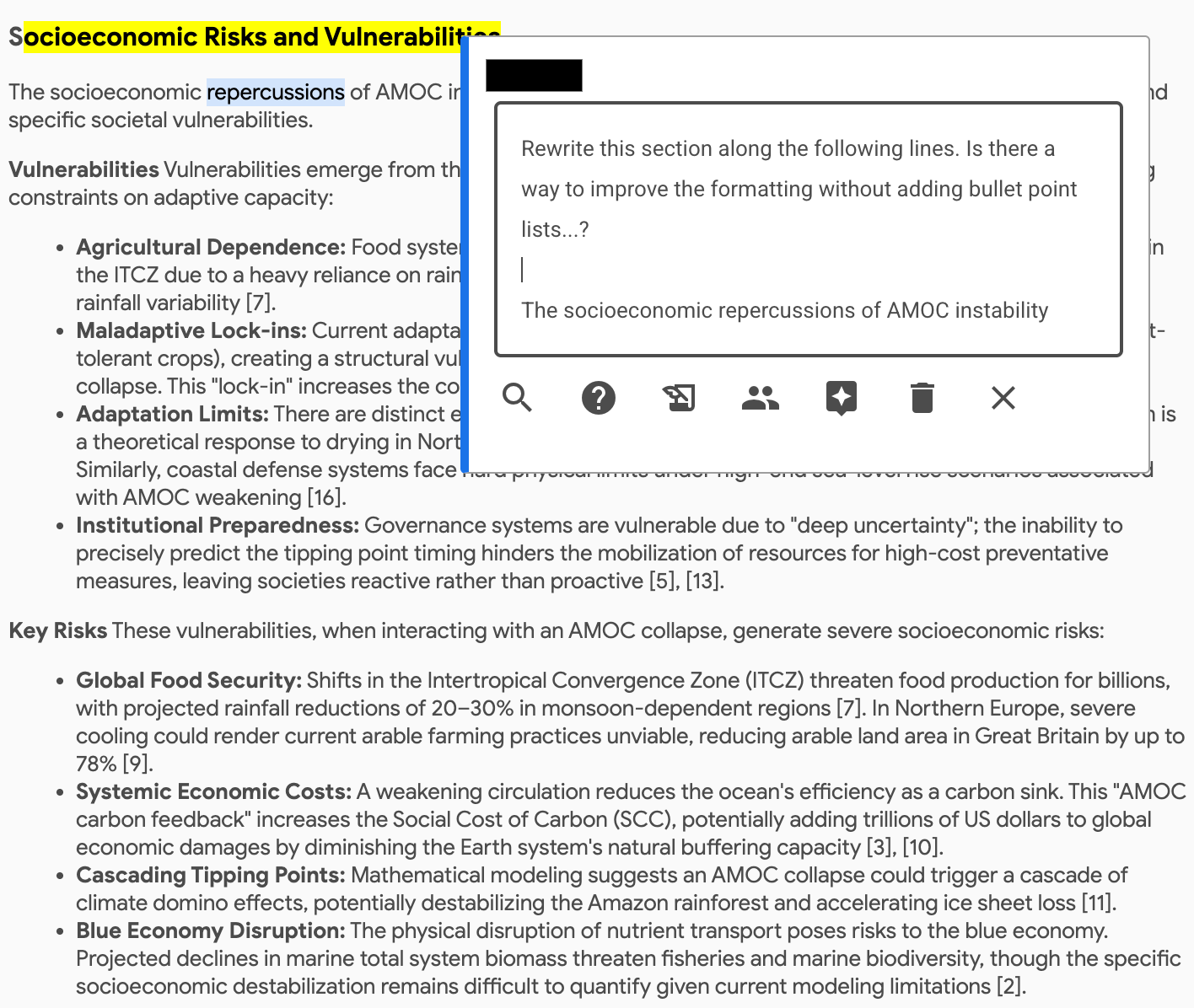} 
        \caption{\small User Instruction (Before)}
        \label{fig:before}
    \end{minipage}
    \hfill
    \begin{minipage}[t]{0.48\textwidth}
        \centering
        \includegraphics[width=\textwidth]{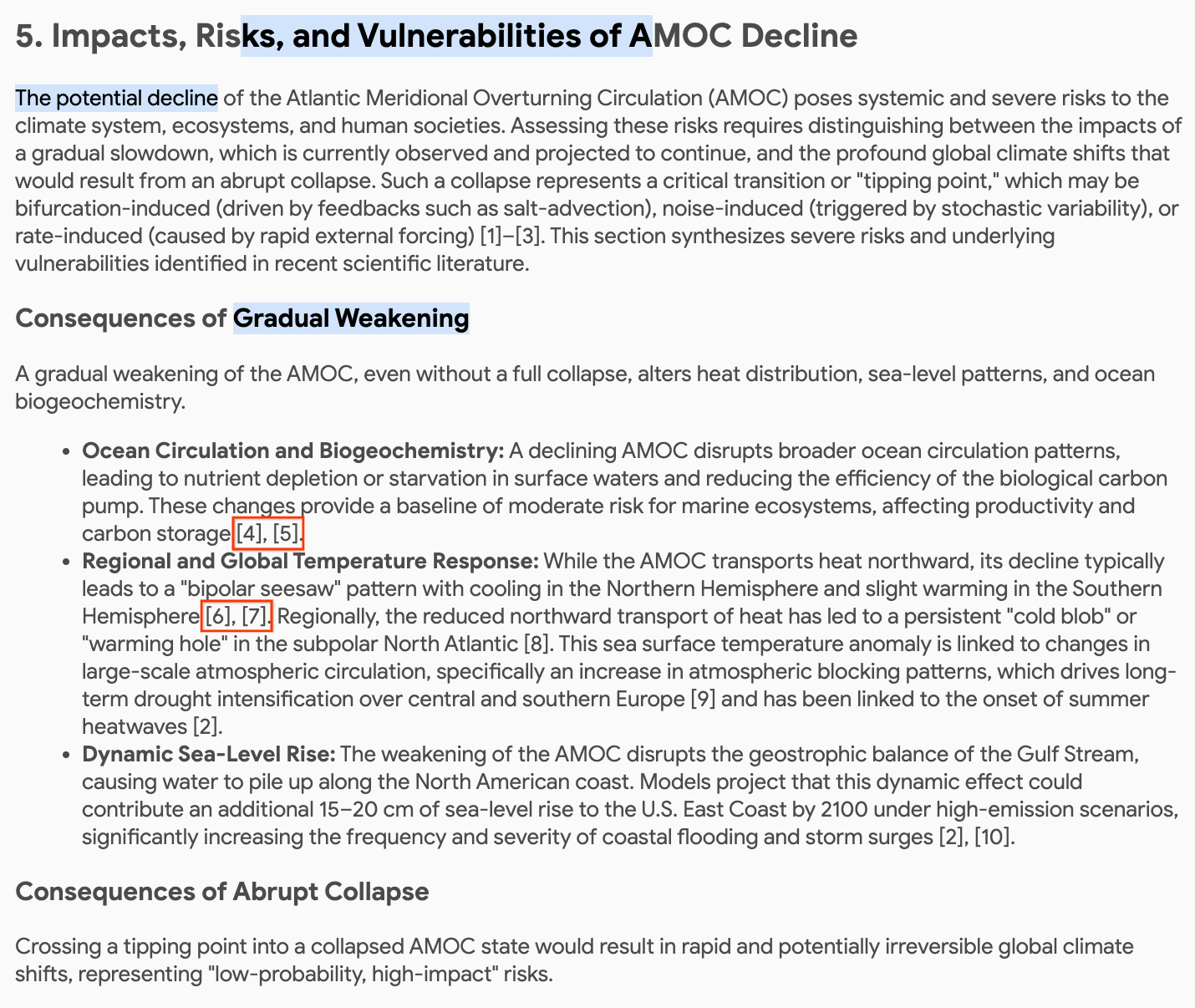}
        \par \vspace{-2pt} $\vdots$ \vspace{-2pt} \par
        \includegraphics[width=\textwidth]{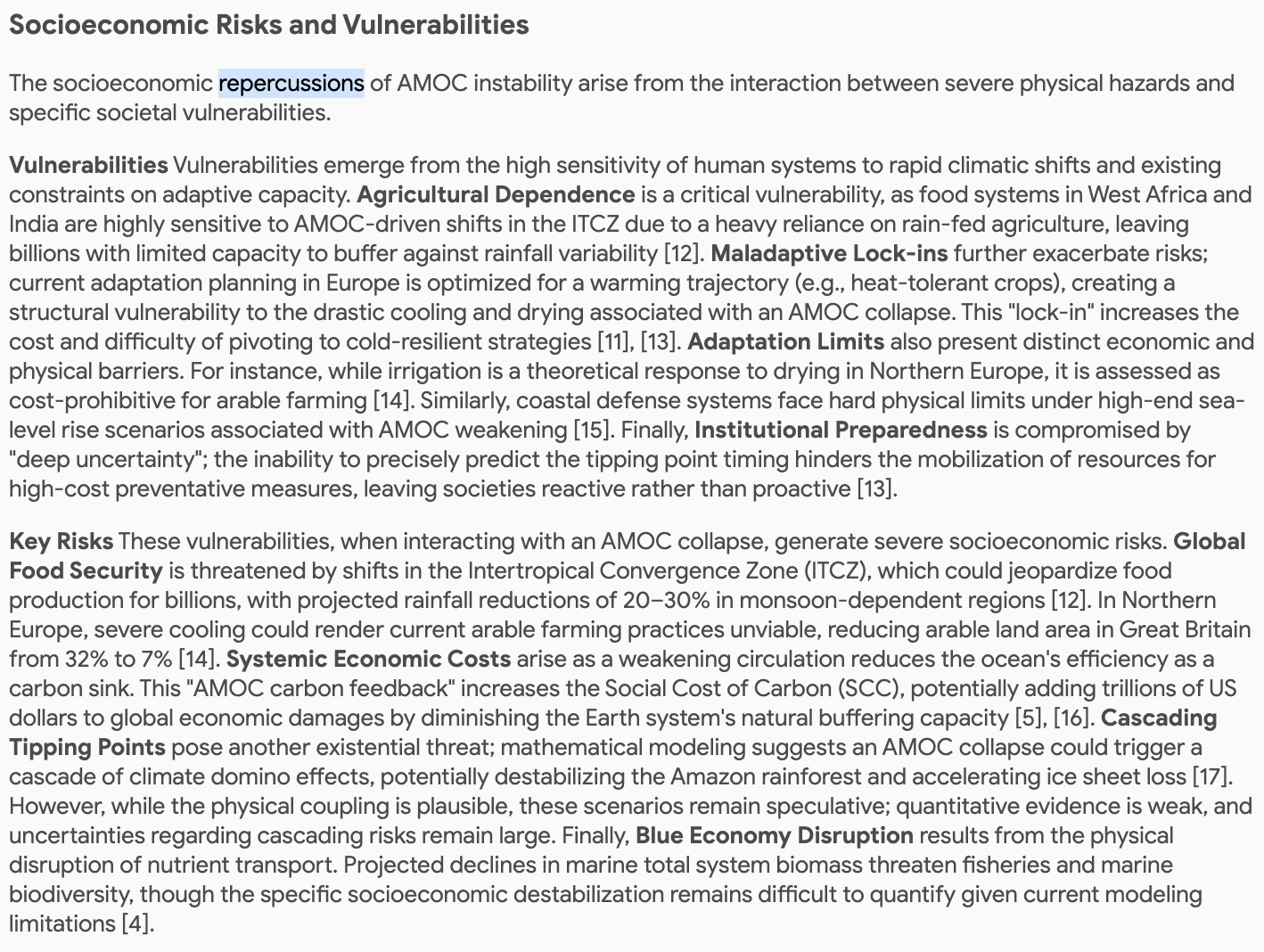} 
        \caption{\small LLM Updated Content (After)}
        \label{fig:after}
    \end{minipage}
    
    \caption{\small Comparison of the user's initial request and the LLM output.}
    \label{fig:comparison}

    \vspace{0.5em} 
    \hrule 
    \vspace{0.5em}
    
    \begin{quote} 
        \small 
        \textbf{Attribution Split:} Human: 95\% \hfill \textbf{LLM: 5\%}
        
        \vspace{0.5em} 
        
        \textbf{Justification:} Revisions 1 and 3 are explicitly marked as manual changes, attributing 100\% of the modifications in those steps to the human. In the transition from Revision 1 to Revision 2, the user provided the exact text for the `Socioeconomic Risks and Vulnerabilities' section and requested a specific formatting change (integration into paragraphs without bullet points). The LLM implemented this instruction, including the necessary connective phrases. However, the LLM also proactively renumbered and reordered the citations throughout the entire document and bibliography to ensure sequential appearance (e.g., changing reference [6] to [4]), a global `cleanup' change that was not requested. Because the vast majority of the text changes across the workflow were either manually entered or directly dictated by the user, the attribution is predominantly human-instructed.
    \end{quote}
\end{figure*}

Figure~\ref{fig:comparison} shows an example of credit assignment evaluation for a revision.

\section{AI Contribution to Revision Content}
\begin{figure}
    \centering
    \includegraphics[width=0.99\linewidth]{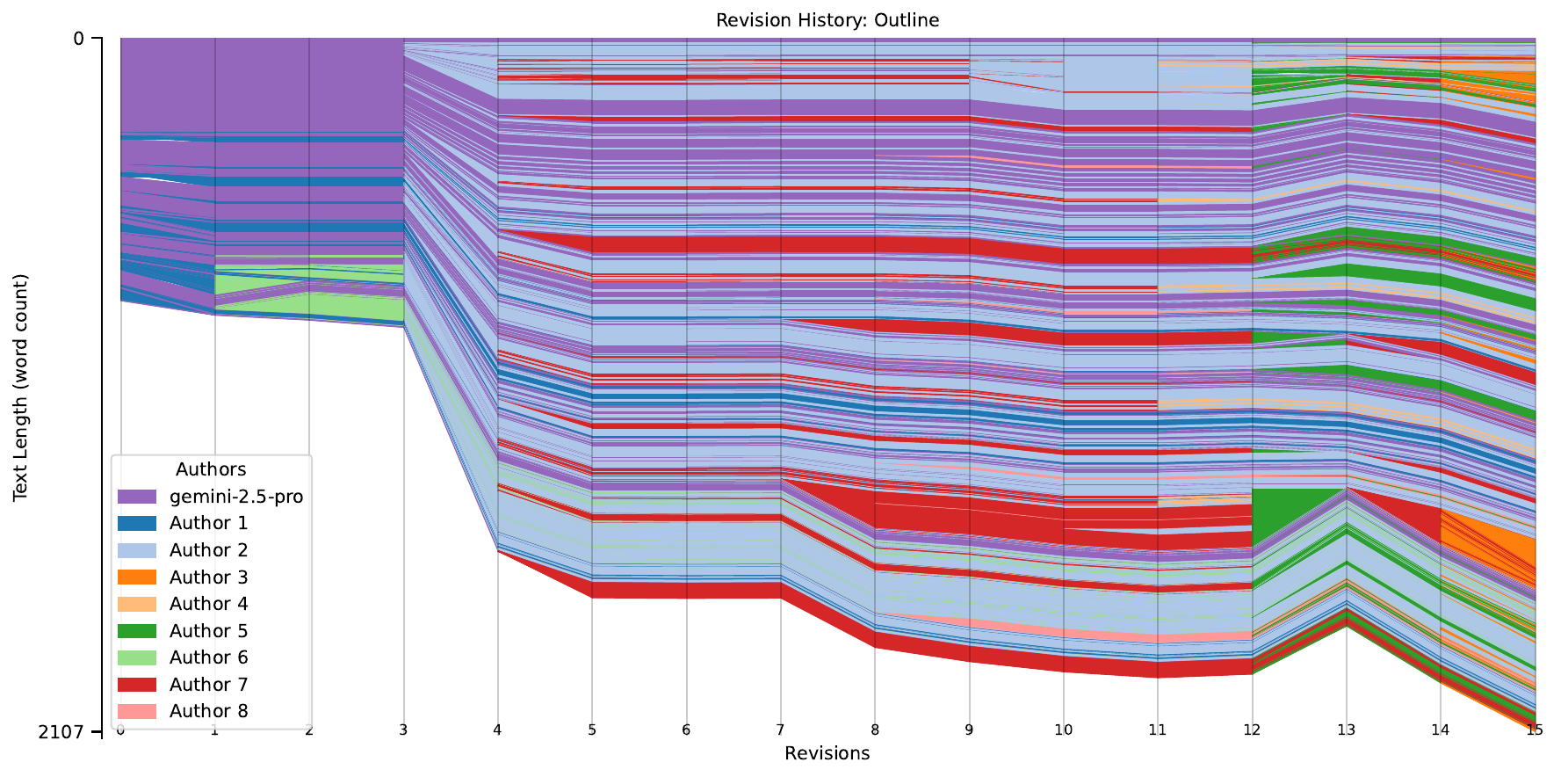}
    \caption{\small History flow for the AMOC Outline process.}
    \label{fig:history-flow-outline}
\end{figure}
Figure~\ref{fig:history-flow-outline} reproduces the history flow visualization for the outline (Phase 1).

\section{Non-LLM baseline of Retention / Intervention analysis}
\label{sect:retention_baseline}
When analyzing which sentences are retained (but possibly changed) from the initial AI written draft and conversely estimating \emph{new} sentences that were added by the experts in the editing process we use an LLM to estimate an alignment between the draft and final version. This has several benefits: (i) the alignment quality is high - we manually checked the alignment and found no errors (ii) the LLM can incorporate a fuzzy monotonicity constraint that accounts for sentence order (iii) we can extract an explanation of the changes, see Appendix \ref{sect:example-sentences}, for further analysis, and (iv) there are no hyperparameters to decide on.

For the sake of reproducibility we provide a simple baseline alignment result that uses an off-the-shelf sentence embedding and greedy matching and arrive at a similar result of both high retention and high intervention.

We preprocess both the AI draft $D$ and the expert-curated final version $F$ by stripping inline citation markers and splitting each text into sentences using the NLTK\footnote{Natural Language Toolkit \citep{nltk2009}, \url{https://www.nltk.org/}. Released under the Apache License 2.0.} \texttt{sent\_tokenize} function. 
We encode all sentences into dense vector representations using the \texttt{all-MiniLM-L6-v2} sentence-transformer model~\citep{wang2020minilm}\footnote{Released under the Apache License 2.0. Available at \url{https://huggingface.co/sentence-transformers/all-MiniLM-L6-v2}.} and compute the pairwise cosine similarity matrix and align sentence pairs based on a threshold. We compute Retention as the fraction of AI-written sentences that have a sentence with cosine similarity above the threshold in the final version. Conversely, we define Intervention as the fraction of sentences in the final report that do not have a sentence with cosine similarity above the threshold in the initial draft.

\begin{figure}[ht]
    \centering
    \includegraphics[width=0.99\linewidth]{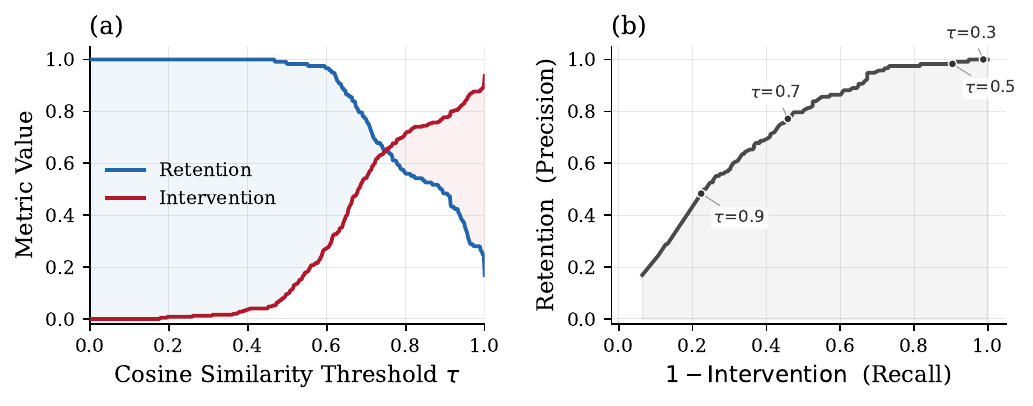}
    \caption{\small Results using baseline alignment method based on sentence embeddings.}
    \label{fig:retention_intervention_baseline}
\end{figure}

Results in Figure \ref{fig:retention_intervention_baseline} align with our findings using the LLM based alignment: using a threshold around $0.6-0.7$ the majority of the AI written sentences have a corresponding sentence in the final report (high retention)  and at the same time most of the sentences in the (longer) final report have no corresponding sentence in the AI draft (high intervention).

\section{Example Sentences}
\label{sect:example-sentences}

\noindent\textbf{Draft:} "This circulation system acts as a crucial "conveyor belt," transporting approximately 1.3 Petawatts (PW) of heat northward across the equator to the mid-latitudes [1]."

\noindent\textbf{Final:} "This circulation system contributes substantially to Atlantic meridional heat transport, with estimates reaching up to ~1.3 PW at low to mid-latitudes [1], [2]."

\noindent\textcolor{darkred}{\textbf{Explanation:}} The final version removes the subjective "crucial", and the popular science metaphor "conveyor belt".

\vspace{1em}

\noindent\textbf{Draft:} "Beyond heat transport, the AMOC plays a pivotal role in the global carbon cycle by sequestering anthropogenic carbon and heat in the deep ocean, thereby regulating the pace of global warming [2]."

\noindent\textbf{Final:} "Beyond heat transport, the AMOC plays an important role in the global carbon cycle. Observational data indicate it facilitates the sequestration of anthropogenic carbon and heat in the deep ocean, thereby influencing the vertical and regional distribution of heat within the climate system [1], [3]."

\noindent\textcolor{darkred}{\textbf{Explanation:}} The final version replaces "pivotal" with "important". This is also an example of a one-to-many sentence matching, where the content of a sentence is expanded in multiple ones in the final version.

\vspace{1em}

\noindent\textbf{Draft:} "Wind-driven upwelling in the Southern Ocean is also essential for returning deep water to the surface, closing the global overturning loop [6]."

\noindent\textbf{Final:} "This upwelling is not merely a passive suction but is mechanically driven by two main processes: strong westerly winds in the Southern Ocean induce northward Ekman transport and divergence, effectively lifting deep waters (the Drake Passage effect), while wind- and tide-generated internal waves drive diapycnal mixing in the ocean interior [8]."

\noindent\textcolor{darkred}{\textbf{Explanation:}} Replaces the qualitative "essential" with specific details of the physical mechanical processes.

\vspace{1em}

\noindent\textbf{Draft:} "This assessment aims to synthesize the state of knowledge regarding AMOC stability as of late 2023, early 2024. It specifically seeks to: [list of objectives]..."

\noindent\textbf{Final:} "This assessment aims to synthesize the state of knowledge regarding AMOC stability as of the publication date of the latest evidence included in this assessment. It specifically seeks to: [list of objectives]..."

\noindent\textcolor{darkred}{\textbf{Explanation:}} Removes the hallucinated time span, and provides a precise definition of the temporal scope.

\vspace{1em}

\noindent\textbf{Draft:} "A full collapse would result in catastrophic climate shifts: [list]..."

\noindent\textbf{Final:} "Crossing a tipping point into a collapsed AMOC state would result in rapid and potentially long-lasting global climate shifts. [list] ..."

\noindent\textcolor{darkred}{\textbf{Explanation:}} Replaces the emotive "catastrophic" with more precise descriptors.

\vspace{1em}

\noindent\textbf{Draft:} "The scientific community has achieved consensus that the AMOC is weakening and that this weakening is driven significantly by anthropogenic forcing [16]."

\noindent\textbf{Final:} "Multiple lines of observational and proxy evidence suggest that the AMOC may have weakened relative to the mid-20th century [45], [61]."

\noindent\textcolor{darkred}{\textbf{Explanation:}} The final version corrects the overconfident binary claim on consensus, aligning the statement with available lines of evidence. 

\vspace{1em}

\noindent\textbf{Draft:} "A strong AMOC pulls the ITCZ northward; a weak AMOC pushes it southward, dictating rainfall patterns over the Sahel and the Amazon [8]."

\noindent\textbf{Final:} "Recent causal analysis suggests a stabilizing feedback loop: while a strong AMOC pulls the ITCZ northward, a weakening AMOC can shift rainfall patterns in a way that increases dry-season precipitation over the Southern Amazon, potentially offsetting drying trends in that critical biome [18]."

\noindent\textcolor{darkred}{\textbf{Explanation:}} The final version expresses uncertainty more clearly ("can", "potentially") and discusses causal mechanisms.

\vspace{1em}

\noindent\textbf{Draft:} "While the IPCC assessment remains the standard, the recent proliferation of EWS studies and bias-corrected model analyses suggests the risk is non-negligible and potentially underestimated [16]."

\noindent\textbf{Final:} "While the IPCC assessment remains the standard, the recent emergence of EWS studies based on statistical physics [48] and bias-corrected model analyses [49] suggests the risk is non-negligible and potentially underestimated [45]."

\noindent\textcolor{darkred}{\textbf{Explanation:}} Increases detail by discussing methods and adding references.

\vspace{1em}

\noindent\textbf{Draft:} "There is Medium Confidence that the AMOC has weakened relative to the pre-industrial era."

\noindent\textbf{Final:} "While direct observations confirm a weakening trend in the early 21st century [25], there is Low Confidence in attributing a long-term decline solely to anthropogenic forcing relative to the pre-industrial era."

\noindent\textcolor{darkred}{\textbf{Explanation:}} The final version distinguished between trends and attribution and revises the confidence on anthropogenic attribution from "Medium" to "Low".

\section{Interaction Patterns and Revision Attribution}
\label{sec:appendix-interactions}

This section provides a detailed, non-systematic analysis of interaction logs and revision attribution. The categorizations presented here are based on Gemini and have not been independently validated; they should be interpreted as exploratory and indicative rather than definitive.

\subsection{Types of Interaction with AI}

During a session, authors do not just revise the text. They search for evidence sources, chat with the \assistant to research topics, send messages to co-authors, ask the \assistant to revise the report and more.
Table~\ref{tab:action-by-section} reports statistics about the interaction types observed in the \assistant. There were 1,151 interactions in total (one every $\sim$2.5 minutes on average), which we divide into three main categories: 

\noindent\textbf{Revisions:} By revision we mean any change to the document content, either via human edit (Manual) or via instructions to the Assistant (AI). Typically a session is characterized by multiple revisions, one of which is checked in as a new submitted version.

\noindent\textbf{Feedback:} Document-level (Doc) or text span-level (Span) feedback, either directed to AI or to the human co-authors.

\noindent\textbf{Tools:} Using the assistant tools to directly add articles (Add Evidence), search the corpus and the Web (Search) or chat with the assistant for various purposes (Chat).

\begin{table*}[ht]
    \centering
    \caption{\small Types of interactions, and their counts, in the assistant in Phase 1 (Outline), Phase 2 (Sections) and Phase 3 (Full Report). 'Doc' means feedback relevant to the full report, while 'Span' means related to a specific span of text. }
    \small
    \setlength{\tabcolsep}{3pt}
    \begin{tabular}{llrrrrrrrrr}
    \toprule
    & & & \multicolumn{6}{c}{\bf Phase 2} & & \\
    \cmidrule(lr){4-9} 
    \multicolumn{2}{c}{{\bf Interaction}} & {\bf Phase 1} & {\bf \S1} & {\bf \S2} & {\bf \S3} & {\bf \S4} & {\bf \S5} & {\bf \S6} & {\bf Phase 3} & {\bf Total}\\ \midrule
\multirow{2}{*}{{\bf Revisions}} 
    & Manual & 11 & 0 & 1  & 31 & 5  & 14 & 0 & 0  & 62 \\
    & AI     & 32 & 3 & 20 & 78 & 29 & 50 & 8 & 25 & 245 \\ 
    \addlinespace[0.3em]
\multirow{2}{*}{\hspace{1em}\textit{of which submitted versions}}
     & Manual & 4  & 0 & 1 & 12 & 3 & 6  & 0 & 0  & 26 \\
     & AI     & 12 & 2 & 8 & 18 & 9 & 14 & 5 & 10 & 78 \\ \midrule
        \multirow{4}{*}{{\bf Feedback (to)}} 
            & AI (Doc)    & 2  & 0 & 4  & 2  & 8  & 5  & 0  & 4   & 25 \\
            & AI (Span)   & 45 & 8 & 20 & 80 & 38 & 68 & 10 & 116 & 385 \\
            & Human (Doc) & 1  & 0 & 0  & 4  & 1  & 0  & 0  & 0   &  6   \\
            & Human (Span)& 27 & 0 & 4  & 40 & 6  & 25 & 0  & 3   & 105 \\ \midrule
        \multirow{3}{*}{{\bf Tools}}     
            & Add Evidence& 0  & 1 & 0  & 0  & 0  & 2  & 0 & 0  & 3 \\
            & Search      & 9  & 0 & 10 & 24 & 2  & 9  & 4 & 2  & 60 \\
            & Chat        & 29 & 3 & 24 & 99 & 31 & 48 & 7 & 19 & 260 \\ \bottomrule
    \end{tabular}
    \label{tab:action-by-section}
\end{table*}

About a third of the interactions involved feedback to the assistant on specific text spans---as instructions to generate revisions. The second most frequent interaction ($22.5\%$) are requests to the model to generate a new revision; model-generated revisions ($80\%$ of revisions, $75\%$ of submitted versions) are by far preferred over manual ones.
Authors engage frequently with the \assistant in chats ($21.7\%$ of interactions) and send messages to other authors ($9\%$ of interactions). 
Authors used the \assistant functionalities to search the document corpus (60 times) and to add sources manually by uploading individual papers (3 times).

\begin{table}[ht]
    \centering
    \caption{\small Distribution of Chat Topics: Chats concern scientific topics and literature questions, cover the writing process, tool functionalities and tracing the history of specific passages.}
    \small
    \label{tab:requests}
    \begin{tabular}{lr}
        \toprule
        \textbf{Chat Topic} & \textbf{\%} \\
        \midrule
        Review and Critique & 28.4 \\
        Request Content Revision & 24.7 \\
        Manage Sources & 16.0 \\
        Get Help with the Tool & 11.5 \\
        Ask for Information &  11.5 \\
        Retrieve Change History & 7.8 \\        
        \bottomrule
    \end{tabular}
\end{table}

Table~\ref{tab:requests} summarizes the topics of the chats, as categorized by Gemini, showing that chats with the assistant cover a wide set of issues. Table~\ref{tab:chat-examples} provides examples of the different types of messages in the \assistant's chatbox.

\begin{table}[ht]
    \centering
    \small
    \setlength{\tabcolsep}{3pt}
    \caption{\small Comparison of human-directed vs AI-directed feedback.}
    \label{tab:feedback-targets}
    \begin{tabular}{lrr}
        \toprule
        \textbf{Feedback type} & \textbf{AI-directed} & \textbf{Human-directed} \\
        \midrule
        Epistemological & 56\%  & 42\% \\
        Presentation    & 22\%  & 48\% \\
        Sources         & 17\%  & 8\% \\
        Tone            & 5\%   & 2\% \\
        \bottomrule
    \end{tabular}
\end{table}

Table~\ref{tab:feedback-targets} categorizes feedback distinguishing between \emph{epistemological} matters---dimensions such as accuracy, specificity, completeness and uncertainty, \emph{evidence} (sources), \emph{presentation} and \emph{tone}~\citep{bulian2024}. Feedback directed to AI differs somewhat from the human-directed feedback. The majority of the feedback to AI involves knowledge quality (epistemological) issues, while the human-directed feedback concerns presentation and knowledge. Table~\ref{tab:feedback-examples} provides examples of feedback to AI for the different categories, Table~\ref{tab:messages-examples} provides examples of messages to co-authors.

\subsection{AI's Contribution to Revisions Content}

Scientists edit the text directly, or provide critical \emph{feedback} and \emph{instructions} for the \assistant to revise the text. The AI-proposed revision, in turn, is accepted or rejected by the human, further worked on etc. 
Table~\ref{tab:revision-attribution} summarizes the results of an estimate (based on Gemini) of the fraction of revisions that can be attributed to AI vs Human. For example, an instruction such as ``Are there more bullet points to add on Mitigation?'' would lead to some credit to AI, if the AI made some suggestions and the revision was checked in. Similarly, sometimes the model `overgenerates', editing more than required. If the proposal is accepted this leads to some credit for the \assistant. The weighted average of the AI-credited revision content is $25.4\%$ (cf.\ \S\ref{sec:attribution-example} for an example).

\begin{table}[ht]
\centering
\footnotesize
\setlength{\tabcolsep}{1pt}
\caption{\small Revision attribution: Estimated percentage of the content of AI-generated versions which is credited to AI. The Revision row reports counts.}
\setlength{\tabcolsep}{4pt}
\begin{tabular}{@{}l rrrrrrrr @{}} 
\toprule
                   & Outline & \S1  & \S2  & \S3  & \S4  & \S5  & \S6  & Report \\ 
\midrule
\textbf{Revisions} & 16      & 2    & 9    & 30   & 12   & 20   & 5    & 10 \\
\textbf{AI (\%)}   & 38      & 0    & 29   & 31   & 16   & 25   & 4    & 13 \\ 
\bottomrule
\end{tabular}
\label{tab:revision-attribution}
\end{table}

At the same time, the \assistant can struggle to follow instructions. We observed a few errors in synthesizing number ranges from different sources (e.g., with respect to temperatures cooling in response to AMOC collapse); in these cases AI may pick a single number, from a single source. When requested to use specific criteria to make robust comparisons the model output was not always convincing. If pushed, adversarially, to take into account a study providing conflicting results AI may give the study a disproportional amount of space compared to the evidence from numerous other studies. This is a case of obsequious behavior, a well-known challenge for LLMs~\citep{perez-etal-2023-discovering, sharmatowards}, which expert-in-the-loop frameworks may help improve upon.
Lack of comprehensiveness was also noted, e.g., in assessing types of impacts and response options.

\section{Post-Study Survey}
\label{sec:survey-full}

We administered a post-study questionnaire to all 13 participating scientists after the completion of the experiment. The full instrument is reproduced in \S\ref{sec:survey-instrument}; \S\ref{sec:survey-responses} reports the aggregated response distributions for all closed-ended questions; and \S\ref{sec:survey-qualitative} synthesizes the free-text qualitative feedback.

\subsection{Survey Instrument}
\label{sec:survey-instrument}

The questionnaire comprised 15 questions, organized into five sections. Questions used a mixture of Likert-type scales, categorical multiple-choice, and open text. The full text of each question and its response options are listed below.

\subsection*{Section 1: Baseline Efficiency and Time Allocation}

\begin{enumerate}[label=\textbf{Q\arabic*.}, leftmargin=*]
    \item \textbf{Based on your previous experience, please estimate how many person-hours it would typically take you to produce a similar individual contribution without the aid of an AI assistant.} \\
    \emph{[Open text]}

    \item \textbf{How would you quantify the speedup provided by the AI assistant?}
    \begin{itemize}[label=$\circ$, leftmargin=1.5em, itemsep=2pt]
        \item 1x (No speedup)
        \item 2x--3x faster
        \item 4x--6x faster
        \item 7x--10x faster
        \item $>$10x faster
        \item Slower than traditional methods
    \end{itemize}

    \item \textbf{Which specific sub-task experienced the greatest efficiency gain? \emph{(Select all that apply)}}
    \begin{itemize}[label=$\square$, leftmargin=1.5em, itemsep=2pt]
        \item Initial literature retrieval and screening
        \item Structuring and outlining
        \item Drafting text and summarizing papers
        \item Harmonizing tone and presentation among co-authors
        \item Other
    \end{itemize}
\end{enumerate}

\vspace{1em}
\hrule
\vspace{1em}

\section*{Section 2: Usability, Workload, and Cognitive Load}

\begin{enumerate}[label=\textbf{Q\arabic*.}, start=4, leftmargin=*]
    \item \textbf{How satisfied were you with the overall usability of the AI Assistant environment?}
    \begin{itemize}[label=$\circ$, leftmargin=1.5em, itemsep=2pt]
        \item Very satisfied
        \item Somewhat satisfied
        \item Neutral
        \item Somewhat dissatisfied
        \item Very dissatisfied
    \end{itemize}
    \textbf{Optional: What specific UI features helped or hindered your workflow?} \\
    \emph{[Open text]}

    \item \textbf{How would you compare the cognitive burden of reviewing and editing the AI's drafts versus drafting the text from scratch?}
    \begin{itemize}[label=$\circ$, leftmargin=1.5em, itemsep=2pt]
        \item Significantly higher
        \item Slightly higher
        \item About the same
        \item Slightly lower
        \item Significantly lower
    \end{itemize}

    \item \textbf{How frequently did you find yourself needing to completely rewrite an AI-generated section?}
    \begin{itemize}[label=$\circ$, leftmargin=1.5em, itemsep=2pt]
        \item Very frequently
        \item Occasionally
        \item Rarely
        \item Never
    \end{itemize}
\end{enumerate}

\vspace{1em}
\hrule
\vspace{1em}

\section*{Section 3: Trust, Anchoring, and Task Delegation}

\begin{enumerate}[label=\textbf{Q\arabic*.}, start=7, leftmargin=*]
    \item \textbf{To what extent do you agree: ``The AI's initial draft anchored my thinking and narrowed the scope of evidence or perspectives I considered.''}
    \begin{itemize}[label=$\circ$, leftmargin=1.5em, itemsep=2pt]
        \item Strongly Agree
        \item Agree
        \item Neutral
        \item Disagree
        \item Strongly Disagree
    \end{itemize}

    \item \textbf{How much did you trust the AI to accurately represent nuanced uncertainties and conflicting evidence?}
    \begin{itemize}[label=$\circ$, leftmargin=1.5em, itemsep=2pt]
        \item Completely trusted it
        \item Mostly trusted it, but required minor verification
        \item Somewhat trusted it, required heavy verification
        \item Did not trust it at all
    \end{itemize}

    \item \textbf{Our data shows researchers frequently messaged AI for epistemological tasks while messaging human co-authors for presentation and structural issues. Does this align with your experience? Why?} \\
    \emph{[Open text]}
\end{enumerate}

\vspace{1em}
\hrule
\vspace{1em}

\section*{Section 4: AI Behavior and Model Differences}

\begin{enumerate}[label=\textbf{Q\arabic*.}, start=10, leftmargin=*]
    \item \textbf{Did you experience instances where the AI was overly agreeable (``sycophantic'')?}
    \begin{itemize}[label=$\circ$, leftmargin=1.5em, itemsep=2pt]
        \item Yes, frequently
        \item Yes, occasionally
        \item No, it maintained a balanced consensus view
        \item I did not test it in this manner
    \end{itemize}

    \item \textbf{We upgraded the model from Gemini 2.5 Pro to Gemini 3 Pro Preview. Did you notice a qualitative difference?} \\
    \emph{[Open text]}
\end{enumerate}

\vspace{1em}
\hrule
\vspace{1em}

\section*{Section 5: Collaboration Dynamics and Future Deployment}

\begin{enumerate}[label=\textbf{Q\arabic*.}, start=12, leftmargin=*]
    \item \textbf{For future high-stakes assessments, would you recommend keeping the ``flat'' collaborative structure, or would a different structure work better with the AI?} \\
    \emph{[Open text]}

    \item \textbf{How satisfied are you with the scientific rigor and overall quality of the final AMOC report?}
    \begin{itemize}[label=$\circ$, leftmargin=1.5em, itemsep=2pt]
        \item Very satisfied (Peer-review ready)
        \item Satisfied (Solid foundation, needs minor polish)
        \item Neutral
        \item Dissatisfied (Requires significant human overhaul)
        \item Very dissatisfied
    \end{itemize}

    \item \textbf{What specific safeguards are strictly necessary before deploying an AI assistant for an official, high-stakes assessment?} \\
    \emph{[Open text]}

    \item \textbf{Any additional comments on your experience?} \\
    \emph{[Open text]}
\end{enumerate}

\subsection{Aggregated Response Distributions}
\label{sec:survey-responses}

Table~\ref{tab:survey-full} reports the response distributions for all closed-ended questions. All 13 participants responded to every closed-ended question. Q3 allowed multiple selections; all other questions were single-response.

\renewcommand{\arraystretch}{1.3}
\small
\begin{longtable}{l p{8cm} r r}
\caption{\small Aggregated post-study survey results ($n{=}13$). Q3 accepts multiple selections; all other questions are single-response and sum to 13.}
\label{tab:survey-full} \\
\toprule
\textbf{Question} & \textbf{Response Options} & \textbf{Count} & \textbf{\%} \\
\midrule
\endfirsthead

\multicolumn{4}{c}%
{{\bfseries \tablename\ \thetable{} -- continued from previous page}} \\
\toprule
\textbf{Question} & \textbf{Response Options} & \textbf{Count} & \textbf{\%} \\
\midrule
\endhead

\midrule
\multicolumn{4}{r}{{Continued on next page}} \\
\endfoot

\bottomrule
\endlastfoot

\multirow{6}{*}{\shortstack[l]{\textbf{Q2:} AI Speedup}}
  & 1x (No speedup) & 0 & 0 \\
  & 2x--3x faster & 7 & 54 \\
  & 4x--6x faster & 1 & 8 \\
  & 7x--10x faster & 1 & 8 \\
  & $>$10x faster & 4 & 31 \\
  & Slower than traditional methods & 0 & 0 \\
\midrule
\multirow{4}{*}{\shortstack[l]{\textbf{Q3:} Greatest Efficiency\\ Gain \emph{(multi-select)}}}
  & Initial literature retrieval and screening & 7 & 54 \\
  & Structuring and outlining & 3 & 23 \\
  & Drafting text and summarizing papers & 8 & 62 \\
  & Harmonizing tone and presentation & 0 & 0 \\
\midrule
\multirow{5}{*}{\shortstack[l]{\textbf{Q4:} AI Assistant\\ Usability}}
  & Very satisfied & 1 & 8 \\*
  & Somewhat satisfied & 6 & 46 \\*
  & Neutral & 3 & 23 \\*
  & Somewhat dissatisfied & 2 & 15 \\*
  & Very dissatisfied & 1 & 8 \\*
\midrule
\multirow{5}{*}{\shortstack[l]{\textbf{Q5:} Cognitive Burden\\ vs.\ Scratch}}
  & Significantly higher & 2 & 15 \\*
  & Slightly higher & 2 & 15 \\*
  & About the same & 1 & 8 \\*
  & Slightly lower & 2 & 15 \\*
  & Significantly lower & 6 & 46 \\*
\midrule
\multirow{4}{*}{\shortstack[l]{\textbf{Q6:} Frequency of\\ Complete Rewrite}}
  & Very frequently & 2 & 15 \\*
  & Occasionally & 8 & 62 \\*
  & Rarely & 2 & 15 \\*
  & Never & 1 & 8 \\*
\midrule
\multirow{5}{*}{\shortstack[l]{\textbf{Q7:} AI Anchoring}}
  & Strongly Agree & 0 & 0 \\*
  & Agree & 4 & 31 \\*
  & Neutral & 6 & 46 \\*
  & Disagree & 2 & 15 \\*
  & Strongly Disagree & 1 & 8 \\*
\midrule
\multirow{4}{*}{\shortstack[l]{\textbf{Q8:} Trust in AI\\ Nuance Handling}}
  & Completely trusted it & 0 & 0 \\*
  & Mostly trusted, minor verification & 4 & 31 \\*
  & Somewhat trusted, heavy verification & 5 & 38 \\*
  & Did not trust it at all & 4 & 31 \\*
\midrule
\multirow{4}{*}{\shortstack[l]{\textbf{Q10:} AI Sycophancy}}
  & Yes, frequently & 1 & 8 \\*
  & Yes, occasionally & 7 & 54 \\*
  & No, maintained balanced view & 3 & 23 \\*
  & Did not test it & 2 & 15 \\*
\midrule
\multirow{5}{*}{\shortstack[l]{\textbf{Q13:} Satisfaction with\\ Final Report}}
  & Very satisfied (Peer-review ready) & 0 & 0 \\*
  & Satisfied (Solid foundation, minor polish) & 8 & 62 \\*
  & Neutral & 3 & 23 \\*
  & Dissatisfied (Requires significant overhaul) & 2 & 15 \\*
  & Very dissatisfied & 0 & 0 \\*
\end{longtable}

\subsection{Qualitative Feedback Synthesis}
\label{sec:survey-qualitative}

Seven of the fifteen questions elicited open-text responses. We synthesize the feedback thematically, with illustrative verbatim quotes.

\paragraph{Baseline Effort Estimates (Q1).}
Nine participants estimated the person-hours required to produce a comparable assessment without AI. Estimates ranged widely, from 16 hours (for an individual contribution) to over 300 person-hours (for the collective effort), with the majority clustering around 100--200 person-hours. Several participants emphasized that estimates depend strongly on prior familiarity with the topic and the scope of the task. For instance, one participant noted: ``A person who is very familiar with the respective literature might manage in 150 person hours, a climate scientist working in a different field may easily need 300 person-hours.'' Another estimated ``200 hours as a lower bound,'' while one participant who framed the question around their individual contribution (logged at 2 hours) estimated 16 hours. These estimates are broadly consistent with archival evidence and provide an independent baseline for the efficiency gains discussed in \S\ref{sect:results}.

\paragraph{UI and Workflow (Q4).}
Five participants provided feedback on the AI Assistant's interface. Two key themes emerged. First, the chatbox was valued for rapid cross-checking: ``The ability to rapidly cross-check aspects and queries with the chatbox was very valuable.'' Second, responsiveness and UI friction were recurring concerns: ``The slow response time of the AI Assistant and the clunkiness of the UI hindered workflow, as well as the tendency in earlier versions of the tool to make unrelated changes to the document in revisions (though this was substantially improved over the course of the project).'' One participant observed that the editing paradigm---prompting-based rather than direct editing---was ``powerful once used to prompting vs editing,'' but that ``collaboration is less intuitive than Google Docs.'' Another noted that the chat and draft panels ``felt disconnected and should be linked.''

\paragraph{Task Delegation Reasoning (Q9).}
Ten participants responded to whether AI was used for epistemological tasks while humans were consulted for presentation and structural issues. The responses broadly confirmed this pattern, but with important nuances. Participants consistently delegated source finding, summarization, and knowledge synthesis to AI: ``AI is strong in finding/synthesizing evidence (objective), but we trust human judgment.'' Conversely, statement verification, perspective balancing, and debating evidence quality were reserved for human co-authors: ``Checking statements is definitively more a work among scientists.'' One participant highlighted the temporal advantage of AI: ``when messaging the AI one gets a quasi-immediate answer or response, and with human co-authors it was unclear when a response would be received.'' However, the boundary was not rigid: some participants also used AI for tone adjustment, and several reported routing statement-checking to human co-authors when surprised by the AI's output.

\paragraph{Model Upgrade Perceptions (Q11).}
All 13 participants commented on the transition from Gemini 2.5 Pro (Phase~1) to Gemini 3 Pro Preview (Phases~2--3). Perceptions were split. A subset of participants noticed substantial improvements: ``Massively. Gemini 3 capabilities were far superior in quality of evidence, presentation, structure, and specific change requests.'' Another noted that ``Gemini 2.5 was very verbose and Gemini 3 was more on point, concise, and seemed to be more reliable.'' However, 9 of 13 participants either did not notice a meaningful change, could not separate model effects from concurrent tool improvements, or lacked sufficient cross-phase experience to assess differences. One participant captured this ambiguity: ``it is hard to separate out the effects of model version from other improvements made to the tool during the course of the experiment.''

\paragraph{Collaborative Structure (Q12).}
Eleven participants responded. A clear preference emerged for a \emph{layered} drafting model over the flat structure used in the experiment. Several proposed ``a small core team drafting the first version, while a separate review team critically reviews the generated draft.'' One participant elaborated: ``A more layered structure to the drafting would be valuable, which could include a set of core authors that lead and drive the text forward, supported by other authors that perform regular reviewing sweeps.'' Others endorsed the flat structure conditionally: ``The flat structure works reasonably well if experts are distributed across chapters and the performance (UI, response times, etc.) of the AI tooling is improved.'' One participant suggested that AI could help mitigate engagement imbalances: ``Ideally, an AI platform could improve and foster a more balanced engagement.''

\paragraph{Necessary Safeguards (Q14).}
Twelve participants identified safeguards needed before deploying AI for high-stakes assessments. The most frequently cited requirement was \textbf{evidence traceability}: ``An IPCC report would require full traceability of all statements to the original source.'' Closely related were calls for \textbf{grounding checks}: ``More grounding checks, particularly concerning numeracy (and all the usual pitfalls, hallucinated literature, wrong context, etc.).'' Other prominent themes included \textbf{bias mitigation and transparency}, \textbf{full literature access} (including behind paywalls and in other languages), and \textbf{uncertainty synthesis}: ``the assistant struggled to assess uncertainty ranges\ldots\ The current agent primarily quotes the individual ranges without synthesizing them.'' One participant called for ``a feature where each sentence has to be signed off for factual accuracy by a human expert.''

\paragraph{Additional Comments (Q15).}
Eight participants offered closing reflections. Despite the acknowledged limitations of the prototype, the overall sentiment was positive: ``Still with a lot of edges, but love the concept.'' Several expressed desire to continue using the tool: ``It made the review process quite fun, I discovered papers I otherwise would have\ldots\ I actually would love to use it more in my research.'' Another viewed the experiment as a stepping stone: ``Given the expansion of the scientific literature, a tool that allows human experts to come to grips with this knowledge is essential. This first experiment provides a glimpse of what might be possible.'' One participant noted that they ``deliberately included provocative literature to see how the AI, the functionality of the collaborative tool and co-authors would respond,'' underscoring the active, stress-testing stance adopted by the expert participants.

\renewcommand{\thesubsection}{\arabic{subsection}}

\small
\newpage
\section{AMOC Outline: Draft}
\label{sect:outline:draft}
\setcounter{subsection}{0}

\section*{An Assessment of Atlantic Meridional Overturning Circulation (AMOC) Stability: Tipping Points, Uncertainties, and Impacts}
Abstract

(Summary of the paper's key findings: the physical basis of AMOC, evidence for its slowdown, the debate surrounding tipping points, potential impacts of a collapse, and a concluding statement on the state of scientific understanding and future research needs.)
\subsection{Introduction}
\begin{itemize}
    \item Definition and Importance of the Atlantic Meridional Overturning Circulation (AMOC)
    \begin{itemize}
        \item Role as a key component of the global climate system.
        \item Contribution to heat, freshwater, and carbon transport.
    \end{itemize}
    \item The Emerging Concern: AMOC Instability and Abrupt Climate Change
    \begin{itemize}
        \item Historical context from paleoclimatic records.
        \item Modern concerns driven by anthropogenic climate change.
    \end{itemize}
    \item Scope and Objectives of the Review
    \begin{itemize}
        \item To synthesize the current scientific understanding of AMOC stability.
        \item To critically evaluate conflicting evidence regarding tipping points.
        \item To outline risks, impacts, and potential response strategies.
    \end{itemize}
\end{itemize}

\subsection{The Physical Science Basis of AMOC}
\begin{itemize}
    \item Driving Mechanisms
    \begin{itemize}
        \item Thermohaline Forcing: The role of temperature (thermo) and salinity (haline) gradients.
        \begin{itemize}
            \item Deep water formation in the North Atlantic (Labrador and Nordic Seas).
        \end{itemize}
        \item Wind-Driven Forcing: The contribution of wind stress to the upper-ocean circulation.
    \end{itemize}
    \item Structure and Pathways
    \begin{itemize}
        \item Northward flow of warm, salty surface water.
        \item Southward flow of cold, dense North Atlantic Deep Water (NADW).
    \end{itemize}
    \item AMOC's Role in the Earth System
    \begin{itemize}
        \item Regulating Northern Hemisphere temperatures.
        \item Influencing the position of the Intertropical Convergence Zone (ITCZ).
        \item Sequestration of heat and carbon in the deep ocean.
    \end{itemize}
\end{itemize}

\subsection{Evidence for AMOC Variability and Recent Slowdown}
\begin{itemize}
    \item Paleoclimatic Evidence of Past Abrupt Changes
    \begin{itemize}
        \item Dansgaard-Oeschger events and Heinrich stadials.
        \item The Younger Dryas cooling event as a potential analogue for AMOC collapse.
    \end{itemize}
    \item Modern Observational Evidence
    \begin{itemize}
        \item Direct measurements from mooring arrays (e.g., RAPID at 26.5$^\circ$N, OSNAP).
        \item Indirect proxies (e.g., sea surface temperature ``cold blob'' in the subpolar North Atlantic).
        \item Observed trends and multi-decadal variability.
    \end{itemize}
    \item Climate Model Projections
    \begin{itemize}
        \item Projections of a gradual 21st-century weakening in most CMIP6 models.
        \item Inter-model spread and differences in sensitivity to freshwater forcing.
    \end{itemize}
\end{itemize}

\subsection{The Tipping Point Debate: Conflicting Evidence and Scientific Uncertainties}
\begin{itemize}
    \item The Concept of Bistability and Hysteresis
    \begin{itemize}
        \item Stommel's box model: The foundation of AMOC's ``on'' and ``off'' states.
        \item The role of freshwater forcing as the primary trigger for a state transition.
    \end{itemize}
    \item Evidence Supporting an Approaching Tipping Point
    \begin{itemize}
        \item Studies indicating a loss of resilience and the emergence of early-warning signals (e.g., increased variance, autocorrelation).
        \item Analysis suggesting the AMOC is at its weakest state in over a millennium.
    \end{itemize}
    \item Evidence Against an Imminent Collapse
    \begin{itemize}
        \item IPCC AR6 assessment: Abrupt collapse before 2100 is considered very unlikely.
        \item Arguments that current models are overly stable or do not correctly represent key feedback mechanisms.
        \item The stabilizing effect of salinity advection feedback.
    \end{itemize}
    \item Key Scientific Uncertainties
    \begin{itemize}
        \item The precise level of freshwater forcing required to trigger a collapse.
        \item Relative contributions of Greenland ice melt, sea ice export, and precipitation changes.
        \item Fidelity of climate models in representing deep convection, ocean eddies, and boundary currents.
        \item Discrepancies between model-based and proxy-based reconstructions of past AMOC strength.
    \end{itemize}
\end{itemize}

\subsection{Risks, Impacts, and Vulnerabilities of AMOC Weakening or Collapse}
\begin{itemize}
    \item Regional and Global Climatic Impacts
    \begin{itemize}
        \item Widespread cooling across the North Atlantic and Northern Europe ($\sim$5--10$^\circ$C in some regions).
        \item Significant southward shift in tropical rainfall belts, affecting monsoons in Africa and South America.
        \item Dynamic sea-level rise along the Atlantic coast of North America.
        \item Weakening of the North Atlantic carbon sink.
    \end{itemize}
    \item Socioeconomic and Ecosystem Vulnerabilities
    \begin{itemize}
        \item Impacts on marine ecosystems, plankton productivity, and fisheries.
        \item Risks to agriculture and water security in Europe, the Sahel, and parts of the Americas.
        \item Potential for increased frequency or intensity of winter storms in Europe.
    \end{itemize}
\end{itemize}

\subsection{Mitigation and Adaptation Strategies}
\begin{itemize}
    \item Mitigation as the Primary Prevention
    \begin{itemize}
        \item The imperative of reducing greenhouse gas emissions to limit global warming and associated Greenland ice melt.
    \end{itemize}
    \item Adaptation to a Weaker AMOC
    \begin{itemize}
        \item Development of robust early-warning systems based on real-time observations and improved models.
        \item Building resilience in vulnerable sectors:
        \begin{itemize}
            \item Coastal planning for accelerated sea-level rise.
            \item Diversifying agricultural systems to cope with shifts in temperature and precipitation.
        \end{itemize}
        \item The challenge of planning for a high-impact, deeply uncertain event.
    \end{itemize}
\end{itemize}

\subsection{Conclusion and Future Directions}
\begin{itemize}
    \item Synthesis of the Current State of Knowledge
    \begin{itemize}
        \item Consensus on ongoing AMOC weakening, but deep uncertainty on the proximity to a tipping point.
        \item The risk profile: A low-probability (but potentially increasing) event with catastrophic consequences.
    \end{itemize}
    \item Key Research Priorities
    \begin{itemize}
        \item Sustaining and expanding ocean observation systems.
        \item Improving the representation of key physical processes in climate models.
        \item Integrating paleoclimatic data, observations, and models to better constrain AMOC stability thresholds.
        \item Enhancing interdisciplinary research on the cascading socioeconomic impacts of an abrupt change.
    \end{itemize}
\end{itemize}

References

\newpage
\section{AMOC Outline: Final}
\label{sect:outline:final}
\setcounter{subsection}{0}
\section*{An Assessment of Atlantic Meridional Overturning Circulation (AMOC) Stability: Tipping Points, Uncertainties, and Impacts}

\subsection*{Abstract}
\begin{itemize}
    \item Rationale: State the critical role of the AMOC in the global climate system and the emerging concern over its stability under anthropogenic forcing.
    \item Objectives: Outline the assessment's goal to synthesize current understanding, evaluate conflicting evidence on tipping points, and assess potential impacts, risks, and responses to reduce those risks.
    \item Key Findings: Summarize the assessment's key findings and associated confidence levels, including:
    \begin{itemize}
        \item High confidence in the fundamental physical drivers of the AMOC.
        \item Medium confidence that the AMOC has weakened since the mid-20th century, though the magnitude of this change is uncertain.
        \item There is deep uncertainty and conflicting evidence regarding the proximity to a tipping point, with recent studies suggesting a higher risk than most climate models project, leading to low confidence in predicting the timing of a potential abrupt collapse before 2100 and highlighting the conflict between the IPCC consensus (very unlikely before 2100) and recent studies indicating a more imminent risk.
        \item High confidence that an abrupt AMOC collapse would cause severe, widespread, and effectively irreversible impacts on climate and ecosystems, with the large-scale pattern of these changes being well-established, including Northern Hemisphere cooling and tropical rainfall shifts.
    \end{itemize}
    \item Conclusion: Conclude that an AMOC collapse is a high-impact event whose likelihood, while deeply uncertain, increases with continued greenhouse gas emissions. Emphasize that greenhouse gas emissions reductions (mitigation) are the primary strategy to reduce this risk and outline priorities for observation, modeling, and risk assessment.
\end{itemize}

\subsection{Introduction}
\begin{itemize}
    \item Definition and Importance of the Atlantic Meridional Overturning Circulation (AMOC)
    \begin{itemize}
        \item Role as a key component of the global climate system.
        \item Contribution to heat, freshwater, and carbon transport.
    \end{itemize}
    \item The Emerging Concern: AMOC Instability and Abrupt Climate Change
    \begin{itemize}
        \item Historical context from paleoclimatic records of past abrupt changes, including similarities and differences to the modern situation (e.g., mechanisms, background climate state).
        \item Modern concerns driven by anthropogenic climate change and freshwater input from melting ice sheets.
    \end{itemize}
    \item Scope and Objectives of the Assessment
    \begin{itemize}
        \item To synthesize the current scientific understanding of AMOC stability.
        \item To critically evaluate conflicting lines of evidence regarding tipping points.
        \item To assess the risks and impacts associated with AMOC decline.
    \end{itemize}
\end{itemize}

\subsection{The Physical Science Basis of AMOC}

\begin{itemize}
    \item Driving Mechanisms
    \begin{itemize}
        \item Thermohaline Forcing: The role of temperature (thermo) and salinity (haline) gradients.
        \begin{itemize}
            \item Deep water formation through intense cooling and brine rejection in the North Atlantic (Labrador and Nordic Seas).
        \end{itemize}
        \item Wind-Driven Forcing: The contribution of wind stress to the upper-ocean circulation and its interaction with thermohaline processes.
    \end{itemize}
    \item Structure and Pathways
    \begin{itemize}
        \item Northward flow of warm, salty surface water in the upper limb.
        \item Southward flow of cold, dense North Atlantic Deep Water (NADW) in the lower limb.
    \end{itemize}
    \item AMOC's Role in the Earth System
    \begin{itemize}
        \item Regulating Northern Hemisphere temperatures and sea-ice extent.
        \item Influencing the position of the Intertropical Convergence Zone (ITCZ) and associated rainfall patterns.
        \item Sequestration of anthropogenic heat and carbon in the deep ocean.
    \end{itemize}
    \item Assessment of Evidence and Confidence
    \begin{itemize}
        \item Strength and agreement of evidence on physical drivers from theory, observations, and models.
        \item Assessed confidence level: High confidence in the fundamental thermohaline and wind-driven mechanisms governing the AMOC.
    \end{itemize}
\end{itemize}

\subsection{Evidence for AMOC Variability and Recent Slowdown}
\begin{itemize}
    \item Paleoclimatic Evidence of Past Abrupt Changes
    \begin{itemize}
        \item Dansgaard-Oeschger events and Heinrich stadials as evidence of past instability during glacial periods.
        \item The Younger Dryas cooling event: A potential analogue for AMOC collapse, with an assessment of its limited applicability to present-day conditions due to differences in ice sheet configuration and background climate state.
    \end{itemize}
    \item Modern Observational Evidence
    \begin{itemize}
        \item Direct In-Situ Measurements: Trends and variability from continuous mooring arrays (e.g., RAPID at 26.5$^\circ$N, OSNAP in the subpolar North Atlantic).
        \item Indirect Proxies and Satellite Observation: Reconstructions based on Sea Surface Temperature (SST) ``cold blob'' fingerprints, Sea Surface Salinity (SSS), and ocean mass changes from gravity satellites.
        \item Observed trends showing multi-decadal variability and a potential slowdown since the mid-20th century.
        \item Large sub-seasonal and interannual variability observed in mooring records, which complicates the detection of a long-term anthropogenic trend from the natural ``noise.''
    \end{itemize}
    \item Climate Model Projections
    \begin{itemize}
        \item Introduction to the Hierarchy of Models: The hierarchy of models used in AMOC research ranges from simple conceptual models (e.g., Stommel's box model), and Earth System Models of Intermediate Complexity (EMICs), to fully-coupled General Circulation Models (GCMs) / Earth System Models (ESMs) like those in the CMIP6 ensemble.
        \item Projections of a gradual 21st-century weakening in nearly all CMIP6 models under all emission scenarios.
        \item Inter-model spread and differences in sensitivity to freshwater forcing.
        \item Assessment of model performance and biases (e.g., persistent ``cold bias'' in the North Atlantic) and their fitness for purpose in stability assessments.
        \item Conversely, some recent high-resolution models suggest the AMOC may be more stable than inferred from lower-resolution models, attributing this to a more realistic representation of ocean eddies and overflow processes. This creates a key tension in model-based assessments.
    \end{itemize}
    \item Assessment of Evidence and Confidence
    \begin{itemize}
        \item Agreement across paleoclimatic proxies, observational records, and model hindcasts.
        \item Assessed confidence level: Medium confidence that the AMOC has weakened over the 20th century, but low confidence in the magnitude of the change.
    \end{itemize}
\end{itemize}

\subsection{The Tipping Point Debate: Conflicting Evidence and Scientific Uncertainties}

\begin{itemize}
    \item The Concept of Bistability and Hysteresis
    \begin{itemize}
        \item Stommel's box model as the theoretical foundation for AMOC's potential ``on'' and ``off'' states; this bistability has since been demonstrated in a range of more complex models, including General Circulation Models (GCMs).
        \item The role of freshwater forcing as the primary trigger for a state transition, creating a hysteresis loop where recovery is difficult or irreversible on human timescales.
    \end{itemize}
    \item Evidence Suggesting an Approaching Tipping Point
    \begin{itemize}
        \item Studies identifying early-warning signals (EWS) like increased variance and autocorrelation in proxy and observational records (e.g., Boers 2021, Ditlevsen \& Ditlevsen 2023), though uncertainties remain regarding EWS derived from proxy data. Other analyses suggest CMIP models are overly stable and that a collapse could occur this century, based on physics-based indicators and EWS (e.g., van Westen et al. 2024).
        \item Assessment of the AMOC's role within a network of climate tipping points, where its collapse could trigger cascading effects on other systems (e.g., Amazon rainforest, West Antarctic Ice Sheet), amplifying the overall risk of abrupt global change.
    \end{itemize}
    \item Evidence Suggesting Greater Stability
    \begin{itemize}
        \item IPCC AR6 assessment concluding an abrupt collapse before 2100 is very unlikely.
        \item Studies highlighting the role of stabilizing feedbacks (e.g., salinity advection) that may counteract freshwater forcing.
        \item Critiques of the interpretation of EWS, noting that they may not be unambiguous indicators of an approaching bifurcation in the complex climate system.
        \item The observation that most CMIP6 models, while showing weakening, do not collapse before 2100 except under extreme, multi-century forcing scenarios.
        \item Findings from new high-resolution models that show a more stable AMOC, suggesting that improving model physics might reduce the risk of collapse seen in some other models.
    \end{itemize}
    \item Key Scientific Uncertainties
    \begin{itemize}
        \item The precise level, rate, and duration of freshwater forcing required to trigger a collapse.
        \item The relative contributions of Greenland ice melt, Arctic sea ice export, and precipitation/runoff changes.
        \item Lack of fidelity in climate models in representing critical processes like deep convection, overflow physics, ocean eddies, and boundary currents.
        \item The reliability of EWS in complex, stochastically forced systems like the real-world climate.
    \end{itemize}
    \item Assessment of Evidence and Confidence
    \begin{itemize}
        \item Evaluation of conflicting evidence from IPCC-assessed models versus recent EWS-based studies.
        \item Assessed confidence level: Low confidence in predicting a specific timing for an abrupt collapse, but Medium confidence that the risk of a collapse increases with cumulative emissions and continued warming.
    \end{itemize}
\end{itemize}

\subsection{Impacts, Risks, and Vulnerabilities of AMOC Decline}

\begin{itemize}
    \item Consequences of Gradual AMOC Weakening
    \begin{itemize}
        \item Regional cooling in the North Atlantic and adjacent landmasses, offset by background global warming.
        \item Dynamic sea-level rise along the North American Atlantic coast.
        \item Weakening of the North Atlantic carbon sink, creating a positive feedback to climate change.
    \end{itemize}
    \item Consequences of Abrupt AMOC Collapse
    \begin{itemize}
        \item Widespread and severe cooling across the North Atlantic and Northern Europe ($\sim$5--10$^\circ$C in winter), leading to dramatic changes in regional climate.
        \item Drastic southward shift in tropical rainfall belts, causing severe droughts in the Sahel and disrupting monsoons in Africa, South America, and Asia.
        \item Major disruption to marine ecosystems, including a collapse of plankton productivity and key fisheries.
        \item Long-term contribution to global sea-level rise from thermal expansion changes in the deep ocean.
        \item Compounding effects, where regional cooling interacts with global warming to create novel and extreme weather patterns (e.g., more intense European winter storms).
    \end{itemize}
    \item Socioeconomic and Ecosystem Vulnerabilities
    \begin{itemize}
        \item Risks to food security: Widespread disruption to agriculture and fisheries in the North Atlantic, Europe, the Sahel, and parts of Asia and the Americas.
        \item Risks to water security: Severe droughts and altered freshwater availability affecting populations and hydropower generation, with disproportionate impacts on vulnerable populations in the Global South.
        \item Risks to infrastructure and economy: Threats to coastal infrastructure from regional sea-level rise; disruption of global supply chains and energy systems due to extreme weather and altered climate patterns.
    \end{itemize}
    \item Assessment of Evidence and Confidence
    \begin{itemize}
        \item Strong agreement across models and paleoclimatic data on the spatial pattern of climatic changes.
        \item Assessed confidence level: High confidence in the general pattern of cooling and precipitation shifts following a collapse; Medium confidence in the precise magnitude and regional details of these impacts.
    \end{itemize}
    \item Mitigation, Adaptation, and Governance Strategies
    \begin{itemize}
        \item Mitigation of greenhouse gas emissions as the primary prevention strategy to reduce the likelihood of a tipping point and limit the magnitude of anthropogenic forcing.
        \item Adaptation strategies to prepare for consequences by building resilience under deep uncertainty, including strengthening food and water systems, developing operational early-warning systems, and planning for regional sea-level rise.
        \item Acknowledging the severe limits to adaptation in the event of an abrupt collapse.
        \item Governance and risk management for high-impact, low-likelihood events, utilizing frameworks like Decision Making under Deep Uncertainty (DMDU) to inform robust policy and communicate complex risks.
    \end{itemize}
\end{itemize}

\subsection{Conclusion and Priorities for Research and Assessment}
\begin{itemize}
    \item Synthesis of the Current State of Knowledge
    \begin{itemize}
        \item Consensus on ongoing AMOC weakening, but deep uncertainty and low confidence on the proximity to a tipping point.
        \item The risk profile is that of a low-probability, high-impact event whose likelihood is assessed to be increasing with continued anthropogenic forcing.
    \end{itemize}
    \item Priorities for Strengthening Scientific Understanding and Assessment
    \begin{itemize}
        \item Strengthening the detection and attribution of AMOC change by sustaining and expanding ocean observation systems (e.g., RAPID, OSNAP) and integrating them with satellite measurements of salinity and ocean mass to separate long-term trends from natural variability.
        \item Increasing confidence in AMOC stability projections through model development by improving the representation of deep convection, ocean-ice sheet interactions (including coupling to dynamic ice sheet models), and key overflow pathways in high-resolution climate models, and by reconciling the differences in stability between high- and low-resolution model ensembles. This would reduce uncertainty in model sensitivity to freshwater forcing and better capture freshwater drivers.
        \item Improving constraints on stability thresholds by developing formal methods to combine paleoclimatic proxy data with model simulations of past abrupt changes (e.g., the 8.2 ka event). This would provide a more robust test of model fidelity.
        \item Refining the theoretical basis and application of stability indicators by testing the reliability of early-warning signals (EWS) across a hierarchy of models and in paleoclimatic records to better understand their predictive power and limitations, and by monitoring physical indicators of stability such as freshwater transport.
        \item Enhancing the assessment of socioeconomic risk by coupling climate models simulating AMOC decline with integrated assessment models to quantitatively evaluate cascading impacts on agriculture, energy, and supply chains, moving beyond qualitative risk statements.
        \item Fostering interdisciplinary research to connect physical climate science with impact, adaptation, and vulnerability (IAV) studies, ensuring that risk assessments are not just physically accurate but also societally relevant and actionable.
    \end{itemize}
    \item Final Remarks
    \begin{itemize}
        \item The assessment indicates that reducing the magnitude of anthropogenic forcing (i.e., greenhouse gas emissions) is the most direct and effective strategy to limit the risk of AMOC instability. This underscores the need for a dual approach: aggressive mitigation to reduce the risk, combined with robust, adaptive observation and impact-assessment systems to prepare for the consequences.
    \end{itemize}
\end{itemize}

\subsection*{References}

\subsection*{Methodological Note}
\begin{itemize}
    \item The confidence levels (e.g., high confidence, low confidence) stated in this outline are placeholders representing the intended structure of the final assessment. These confidence judgments will be the formal outcome of the synthesis and evaluation of evidence conducted within the paper, following established methodologies (e.g., IPCC guidance), and are not pre-determined inputs to the assessment process.
\end{itemize}

\newpage
\section{AMOC Report: Draft}
\label{sect:report-draft}
\section*{An Assessment of Atlantic Meridional Overturning Circulation (AMOC) Stability: Tipping Points, Uncertainties, and Impacts}

\setcounter{subsection}{0}

\subsection{Introduction}

\subsubsection{Definition and Importance of the AMOC}

The Atlantic Meridional Overturning Circulation (AMOC) is a system of ocean currents in the North Atlantic Ocean characterized by a northward flow of warm, saline water in the upper layers of the Atlantic, and a southward flow of colder, deeper waters that are part of the thermohaline circulation [1].

This circulation system acts as a crucial "conveyor belt," transporting approximately 1.3 Petawatts (PW) of heat northward across the equator to the mid-latitudes [1]. Beyond heat transport, the AMOC plays a pivotal role in the global carbon cycle by sequestering anthropogenic carbon and heat in the deep ocean, thereby regulating the pace of global warming [2].

\subsubsection{The Emerging Concern: AMOC Instability and Abrupt Climate Change}

The stability of the AMOC has become a focal point of climate research due to its potential non-linear response to forcing. Paleoclimatic records demonstrate that the AMOC is bistable; it can exist in an active "on" state and a weak or collapsed "off" state [3]. Transitions between these states—such as those observed during Dansgaard-Oeschger (D-O) events or the Younger Dryas cooling—can occur with geological rapidity (decades to centuries) [4].

While past abrupt changes were often triggered by catastrophic glacial meltwater pulses, modern concerns arise from anthropogenic climate change. The accelerated melting of the Greenland Ice Sheet (GrIS) and increasing Arctic precipitation are introducing fresh water into the North Atlantic [5]. This freshening reduces the density of surface waters, inhibiting the deep convection required to sustain the overturning loop [5].

\subsubsection{Scope and Objectives}

This assessment aims to synthesize the state of knowledge regarding AMOC stability as of late 2023/early 2024. It specifically seeks to:

\begin{enumerate}
    \item Outline the physical mechanisms governing the circulation.
    \item Critically evaluate the tension between climate model projections (which generally show gradual weakening) and statistical Early Warning Signals (which suggest approaching tipping points). 
    \item Assess the risks and vulnerabilities associated with AMOC decline, particularly for food and water security.
\end{enumerate}

\subsection{The Physical Science Basis of AMOC}

\subsubsection{Driving Mechanisms}

The AMOC is sustained by a delicate interplay of buoyancy (thermohaline) and wind-driven forcing.

\textbf{Thermohaline Forcing:} The primary driver of the overturning cell is the formation of North Atlantic Deep Water (NADW). As warm, salty surface water reaches high latitudes, it loses heat to the atmosphere. This cooling, combined with brine rejection during sea-ice formation, increases water density, causing it to sink in the Labrador and Nordic Seas [6]. This sinking mass draws more surface water northward to replace it, sustaining the loop.

\textbf{Wind-Driven Forcing:} While buoyancy drives the sinking, wind stress significantly influences the upper limb of the AMOC, particularly via the Gulf Stream and the North Atlantic Current. Wind-driven upwelling in the Southern Ocean is also essential for returning deep water to the surface, closing the global overturning loop [6].

\subsubsection{Structure and Pathways}

The circulation is structurally defined by two main limbs:
\begin{enumerate}
    \item \textbf{The Upper Limb:} A northward transport of warm, saline water in the upper 1,000 meters [7].  
    \item \textbf{The Lower Limb:} A southward return flow of cold, dense NADW between 1,000 and 4,000 meters depth [7].
\end{enumerate}

\subsubsection{AMOC’s Role in the Earth System}

The heat transported by the AMOC is responsible for the relatively mild climate of Northwestern Europe compared to other regions at similar latitudes [6]. Furthermore, the inter-hemispheric energy imbalance created by the AMOC influences the position of the Intertropical Convergence Zone (ITCZ). A strong AMOC pulls the ITCZ northward; a weak AMOC pushes it southward, dictating rainfall patterns over the Sahel and the Amazon [8].

\subsubsection{Assessment of Evidence and Confidence}

There is \textbf{High Confidence} in the theoretical understanding of the physical drivers of the AMOC. The mechanisms of deep convection, geostrophic flow, and wind-driven transport are well-established through fluid dynamics theory, in-situ observations, and ocean modeling.

\subsection{Evidence for AMOC Variability and Recent Slowdown}

\subsubsection{Paleoclimatic Evidence of Past Abrupt Changes}

Paleoclimatic proxies, including ice cores and marine sediments, reveal that the AMOC has undergone repeated, abrupt state transitions [9]. The Dansgaard-Oeschger events of the last glacial period involved rapid warming in the Northern Hemisphere associated with AMOC resurgences. Conversely, Heinrich stadials represent periods of AMOC collapse driven by massive iceberg discharges [9]. The Younger Dryas event (approx. 12,900 years ago) serves as a potential analogue for collapse, although direct comparisons are limited by differences in boundary conditions (e.g., presence of the Laurentide Ice Sheet) compared to the modern Holocene baseline [10]. Recovery from such collapsed states may involve rapid sea-ice feedbacks, such as the opening of polynyas, to restart deep water formation [10].

\subsubsection{Modern Observational Evidence}

\textbf{Direct Measurements:} Since 2004, the RAPID-MOCHA-WBTS array at 26.5$^{\circ}$N has provided continuous monitoring of the AMOC. Data indicates a weakening trend from 2004 to 2012, followed by a period of relative stability, though with high interannual variability driven largely by wind forcing [11]. The OSNAP array (Overturning in the Subpolar North Atlantic Program), deployed later, has highlighted that the majority of deep water transformation occurs in the eastern subpolar gyre, challenging earlier assumptions focused solely on the Labrador Sea [12].

\textbf{Indirect Proxies and Fingerprints:} Due to the short duration of direct measurements, researchers rely on sea surface temperature (SST) and salinity (SSS) fingerprints. A persistent region of cooling in the subpolar North Atlantic (the "warming hole" or "cold blob") coupled with excessive warming along the US Northeast coast is consistent with a slowdown of the AMOC [13]. Reconstructions based on these proxies suggest the AMOC is currently in its weakest state in the last millennium [13].

\subsubsection{Climate Model Projections}

\textbf{CMIP6 Ensemble:} The Coupled Model Intercomparison Project Phase 6 (CMIP6) models consistently project a weakening of the AMOC over the 21st century under all emission scenarios [14]. However, the magnitude varies significantly, with decline estimates ranging from 24\% to 39\% under high-emission scenarios.

\textbf{Model Fidelity and Biases:} Standard resolution Global Circulation Models (GCMs) often suffer from biases, such as excessive stability due to poor representation of freshwater runoff spreading and mesoscale eddies. Conversely, recent high-resolution models that resolve ocean eddies suggest the AMOC may be more stable than lower-resolution models imply, attributing this resilience to eddy compensation mechanisms. However, recent studies using strongly eddying ocean-only models demonstrate that AMOC collapse is still possible under freshwater forcing, though eddies may maintain a weak (\~{}5 Sv) residual circulation in the collapsed state [15]. This creates a divergence in the evidence base regarding the precise stability threshold and collapsed state characteristics.

\textbf{Assessment of Evidence and Confidence}

There is \textbf{Medium Confidence} that the AMOC has weakened relative to the pre-industrial era. This assessment is limited by the "noise" of natural multi-decadal variability (e.g., the Atlantic Multidecadal Oscillation) which complicates the detection of an anthropogenic trend in short observational records.

\subsection{The Tipping Point Debate: Conflicting Evidence and Scientific Uncertainties}

The most contentious aspect of current AMOC science is the proximity to a "tipping point"—a critical threshold at which the system bifurcates from a strong mode to a weak/off mode.

\subsubsection{The Concept of Bistability and Hysteresis}

Theoretical understanding rooted in Stommel’s 1961 box model establishes that the AMOC exhibits hysteresis [16]. As freshwater forcing increases (via ice melt and precipitation), the circulation weakens gradually until it reaches a bifurcation point, triggering a rapid collapse. Once collapsed, restoring the AMOC requires reducing freshwater forcing to levels significantly lower than the initial tipping point [16].

\subsubsection{Evidence Suggesting an Approaching Tipping Point}

Recent studies utilizing statistical physics have raised alarms regarding an imminent collapse.

\begin{itemize}
    \item \textbf{Early Warning Signals (EWS):} As a complex system approaches a tipping point, it exhibits "critical slowing down," characterized by increased variance and increased lag-1 autocorrelation in time-series data [17]. Analyses of SST and salinity indices suggest these signals are present and statistically significant, with some studies estimating a potential collapse window within the 21st century (e.g., Ditlevsen \& Ditlevsen, 2023) [17].  
    \item \textbf{Overly Stable Models:} Physics-based indicators derived from observations suggest that many CMIP models are biased toward stability (the "salt-advection feedback" is modeled too weakly). Correcting for this bias suggests the real-world AMOC is closer to the tipping point than the model ensemble average indicates [16].
\end{itemize}

\subsubsection{Evidence Suggesting Greater Stability}

Contrasting the EWS studies, the consensus within the IPCC Sixth Assessment Report (AR6) is that while the AMOC will weaken, an abrupt collapse before 2100 is *very unlikely* [16], [17].

\begin{itemize}
    \item \textbf{Stabilizing Feedbacks:} Mechanisms such as the warming of the deep ocean and changes in the gyre circulation may counteract the freshening signal.
    \item \textbf{Critique of EWS:} Critics argue that EWS in climate data can be generated by changes in external forcing or "red noise" processes unrelated to bifurcation, leading to false positives [18]. Recent modeling shows that CSD indicators can robustly increase even in monostable regimes where no tipping point exists, suggesting these signals may not reliably distinguish between a slowing and a collapsing circulation [18]. 
    \item \textbf{High-Resolution Modeling:} As noted in Section 3, eddy-resolving models tend to show a more resilient circulation, suggesting that the "tipping point" found in intermediate complexity models might be an artifact of coarse resolution [16].
\end{itemize}

\subsubsection{Key Scientific Uncertainties}

The divergence in assessment stems from three deep uncertainties:

\begin{enumerate}
    \item \textbf{Freshwater Hosing:} The exact volume and rate of future meltwater from Greenland remain uncertain [17].  
    \item \textbf{Model Physics:} The inability of standard models to resolve overflow processes and mesoscale eddies accurately [16].  
    \item \textbf{Signal vs. Noise:} The difficulty in distinguishing a trend toward a tipping point from natural multidecadal variability or forced response [18].
\end{enumerate}

\subsubsection{Assessment of Evidence and Confidence}

There is \textbf{Low Confidence} in the timing of a potential collapse. While the IPCC assessment remains the standard, the recent proliferation of EWS studies and bias-corrected model analyses suggests the risk is non-negligible and potentially underestimated [16]. There is \textbf{Medium Confidence} that the risk of crossing a tipping point increases non-linearly with cumulative emissions.

\subsection{Impacts, Risks, and Vulnerabilities of AMOC Decline}

Regardless of whether the decline is gradual or abrupt, the consequences of AMOC weakening are profound.

\subsubsection{Consequences of Gradual Weakening}

Even a moderate slowdown (e.g., 20–30\%) will impact the North Atlantic region.

\begin{itemize}
    \item \textbf{Regional Cooling:} A reduction in northward heat transport leads to a relative cooling of the subpolar North Atlantic, often referred to as the "cold blob" or "warming hole" [19], [16]. While global warming dominates the long-term trend, this region may experience suppressed warming or cooling, affecting agricultural growing seasons in parts of Europe [16], [20].  
    \item \textbf{Dynamic Sea-Level Rise:} The geostrophic balance of the Gulf Stream means that as the current slows, water piles up along the North American coast. Models project that this could lead to an additional 15–20 cm of sea-level rise along the US East Coast by 2100 from this effect alone, exacerbating storm surge risks and coastal flooding [16], [21].
\end{itemize}

\subsubsection{Consequences of Abrupt Collapse}

A full collapse would result in catastrophic climate shifts:

\begin{itemize}
    \item \textbf{Severe Cooling:} Models project drastic cooling across the Northern Hemisphere. For example, simulations indicate that an AMOC shutdown could lead to significant temperature drops, potentially exceeding 10$^{\circ}$C in parts of the North Atlantic and affecting Northwestern Europe, radically altering the habitability and energy demand of the region [16].  
    \item \textbf{Tropical Rainfall Shifts:} The Intertropical Convergence Zone (ITCZ) would shift southward to compensate for the inter-hemispheric energy imbalance caused by reduced northward heat transport [16], [22]. This shift would likely induce severe drying in the West African Monsoon (WAM) region and the Indian Summer Monsoon (ISM) region, while potentially increasing rainfall in the southern Amazon, threatening the water and food security of billions dependent on these systems [22].  
    \item \textbf{Marine Ecosystems:} The cessation of deep convection would starve the deep ocean of oxygen and the surface ocean of nutrients. Earth system models coupled with marine ecosystem models project that a strong AMOC weakening could decrease Total System Biomass (TSB) by approximately 2\% to 4\% globally, with higher trophic levels suffering larger declines due to trophic amplification [14].
\end{itemize}

\subsubsection{Socioeconomic Vulnerabilities}

\begin{itemize}
    \item \textbf{Food Security:} The combined effects of European cooling and tropical drought would cause simultaneous breadbasket failures. In Great Britain, for instance, an AMOC collapse could reduce the area of arable land from 32\% to 7\% due to severe drying, costing the agricultural sector hundreds of millions of pounds annually [20].  
    \item \textbf{Energy and Infrastructure:} Extreme weather changes would render current infrastructure design standards obsolete. Furthermore, the disruption of the North Atlantic carbon sink—projected to reduce ocean carbon uptake by over 10 PgC by 2100 under certain scenarios—would accelerate global warming. This "AMOC carbon feedback" could increase the Social Cost of Carbon (SCC) by roughly 1\%, adding trillions of dollars in global economic damages [23].
\end{itemize}

\subsubsection{Response Strategies}

Given the *deep uncertainty* surrounding the timeline, response strategies must focus on resilience:

\begin{itemize}
    \item \textbf{Adaptation:} Strengthening food systems in the Sahel and Europe is critical. In Great Britain, widespread irrigation could theoretically ameliorate the loss of arable land caused by AMOC-induced drying, but the costs of such technological adaptation may outweigh the benefits unless food prices rise significantly [20].  
    \item \textbf{Risk Management:} Utilizing "Decision Making under Deep Uncertainty" (DMDU) frameworks to develop policies that are robust across a wide range of future AMOC states. The precautionary principle suggests that the only way to minimize the risk of passing the AMOC tipping point is to phase out fossil fuel use and stop deforestation rapidly [16].
\end{itemize}

\subsection{Conclusion and Priorities for Research and Assessment}

\subsubsection{Synthesis of Knowledge}

The scientific community has achieved consensus that the AMOC is weakening and that this weakening is driven significantly by anthropogenic forcing [16]. However, deep uncertainty remains regarding the proximity to a critical tipping point. The conflict between model-based stability assessments and observation-based early warning signals represents a critical knowledge gap [16], [24], [18]. The risk profile is that of a "low-probability, high-impact" event, but the probability is assessed to be increasing with every fraction of a degree of global warming [16].

\subsubsection{Priorities for Strengthening Future Assessments}

\begin{enumerate}
    \item \textbf{Observation:} Sustain and expand the RAPID and OSNAP arrays [7] and integrate them with Deep Argo floats and satellite gravimetry to better separate signal from noise [7], [25].  
    \item \textbf{Modeling:} Prioritize the development of high-resolution, eddy-resolving Earth System Models to resolve the discrepancy in stability sensitivity [26].  
    \item \textbf{Theory:} Refine the theoretical application of Early Warning Signals to stochastic climate data to reduce false positives and negatives [24], [18].  
    \item \textbf{Impact Assessment:} Couple climate models simulating AMOC collapse with Integrated Assessment Models (IAMs) to quantify economic and social risks [23].
\end{enumerate}

\subsubsection{Final Remarks}

The stability of the AMOC is not merely a theoretical oceanographic concern but a critical planetary boundary. This assessment concludes that while the timing of a potential collapse is uncertain, the mechanism is physically sound and the impacts would be globally destabilizing. Consequently, \textbf{mitigation}—the rapid reduction of greenhouse gas emissions—remains the only robust strategy to minimize the probability of triggering this tipping point [16].

\subsection*{References}

[1] M. W. Buckley and J. Marshall, "Observations, inferences, and mechanisms of the Atlantic Meridional Overturning Circulation: A review," *Rev. Geophys.*, vol. 54, no. 1, pp. 5–63, 2016. \\
Available: \href{http://doi.org/10.1002/2015RG000493}{http://doi.org/10.1002/2015RG000493}

[2] V. Caínzos, A. Velo, F. F. Pérez, and A. Hernández-Guerra, "Anthropogenic Carbon Transport Variability in the Atlantic Ocean Over Three Decades," *Global Biogeochem. Cycles*, vol. 36, no. 11, p. e2022GB007475, 2022. \\
Available: \href{http://doi.org/10.1029/2022GB007475}{http://doi.org/10.1029/2022GB007475}

[3] J. van den Berk, S. Drijfhout, and W. Hazeleger, "Characterisation of Atlantic meridional overturning hysteresis using Langevin dynamics," *Earth Syst. Dyn.*, vol. 12, no. 1, pp. 69–84, 2021. \\
Available: \href{http://doi.org/10.5194/esd-12-69-2021}{http://doi.org/10.5194/esd-12-69-2021}

[4] T. M. Lenton, V. N. Livina, V. Dakos, and M. Scheffer, "Climate bifurcation during the last deglaciation?," *Clim. Past*, vol. 8, no. 4, pp. 1127–1139, 2012. \\
Available: \href{http://doi.org/10.5194/cp-8-1127-2012}{http://doi.org/10.5194/cp-8-1127-2012}

[5] Q. Yang, T. H. Dixon, P. G. Myers, J. A. Bonin, D. P. Chambers, M. R. van den Broeke, M. H. Ribergaard, and J. Mortensen, "Recent increases in Arctic freshwater flux affects Labrador Sea convection and Atlantic overturning circulation," *Nat. Commun.*, vol. 7, p. 10525, 2016. \\
Available: \href{http://doi.org/10.1038/ncomms10525}{http://doi.org/10.1038/ncomms10525}

[6] T. Kuhlbrodt, A. Griesel, M. Montoya, A. Levermann, M. Hofmann, and S. Rahmstorf, "On the driving processes of the Atlantic meridional overturning circulation," *Reviews of Geophysics*, 2007. [Online]. \\
Available: \href{http://doi.org/10.1029/2004RG000166}{http://doi.org/10.1029/2004RG000166}

[7] E. Frajka‐Williams, N. P. Foukal, and G. Danabasoglu, "Should AMOC observations continue: how and why?" *Philosophical Transactions of the Royal Society A: Mathematical, Physical and Engineering Sciences*, 2023. [Online]. \\
Available: \href{http://doi.org/10.1098/rsta.2022.0195}{http://doi.org/10.1098/rsta.2022.0195}

[8] A. Högner, G. Di Capua, J. F. Donges, R. V. Donner, G. Feulner, and N. Wunderling, "Causal pathway from AMOC to Southern Amazon rainforest indicates stabilising interaction between two climate tipping elements," *Environmental Research Letters*, 2025. [Online]. \\
Available: \href{http://doi.org/10.1088/1748-9326/addb62}{http://doi.org/10.1088/1748-9326/addb62}

[9] A. H. L. Voelker and J. T. Andrews, "Millennial-Scale Ocean Climate Variability," in *Elsevier eBooks*, 2018. \\
Available: \href{http://doi.org/10.1016/B978-0-12-409548-9.11368-5}{http://doi.org/10.1016/B978-0-12-409548-9.11368-5}

[10] J. Velay-Vitow, D. Chandan, and W. R. Peltier, "Into the Holocene, anatomy of the Younger Dryas cold reversal and preboreal oscillation," *Scientific Reports*, 2024. \\
Available: \href{http://doi.org/10.1038/s41598-024-53591-2}{http://doi.org/10.1038/s41598-024-53591-2}

[11] B. Moat *et al.*, "Pending recovery in the strength of the meridional overturning circulation at 26$^{\circ}$ N," *Ocean Science*, 2020. \\
Available: \href{http://doi.org/10.5194/os-16-863-2020}{http://doi.org/10.5194/os-16-863-2020}

[12] F. Li *et al.*, "Subpolar North Atlantic western boundary density anomalies and the Meridional Overturning Circulation," *Nature Communications*, 2021. \\
Available: \href{http://doi.org/10.1038/s41467-021-23350-2}{http://doi.org/10.1038/s41467-021-23350-2}

[13] L. Caesar, S. Rahmstorf, A. Robinson, G. Feulner, and V. S. Saba, "Observed fingerprint of a weakening Atlantic Ocean overturning circulation," *Nature*, 2018. \\
Available: \href{http://doi.org/10.1038/s41586-018-0006-5}{http://doi.org/10.1038/s41586-018-0006-5}

[14] A. A. Boot, J. Steenbeek, M. Coll, A. S. von der Heydt, and H. A. Dijkstra, "Global Marine Ecosystem Response to a Strong AMOC Weakening Under Low and High Future Emission Scenarios," *Earth's Future*, 2025. \\
Available: \href{http://doi.org/10.1029/2024EF004741}{http://doi.org/10.1029/2024EF004741}

[15] R. M. van Westen, M. Kliphuis, and H. A. Dijkstra, "Collapse of the Atlantic Meridional Overturning Circulation in a Strongly Eddying Ocean-Only Model," *Geophysical Research Letters*, 2025. \\
Available: \href{http://doi.org/10.1029/2024GL114532}{http://doi.org/10.1029/2024GL114532}

[16] S. Rahmstorf, "Is the Atlantic Overturning Circulation Approaching a Tipping Point?," *Oceanography*, vol. 37, no. 3, pp. 16-29, 2024. [Online]. \\
Available: \href{http://doi.org/10.5670/oceanog.2024.501}{http://doi.org/10.5670/oceanog.2024.501}

[17] P. Ditlevsen and S. Ditlevsen, "Warning of a forthcoming collapse of the Atlantic meridional overturning circulation," *Nature Communications*, vol. 14, art. 4254, 2023. [Online]. \\
Available: \href{http://doi.org/10.1038/s41467-023-39810-w}{http://doi.org/10.1038/s41467-023-39810-w}

[18] C. C. Zimmerman, T. J. W. Wagner, E. A. Maroon, and D. E. McNamara, "Slowed Response of Atlantic Meridional Overturning Circulation Not a Robust Signal of Collapse," *Geophysical Research Letters*, vol. 52, e2024GL112415, 2025. [Online]. \\
Available: \href{http://doi.org/10.1029/2024GL112415}{http://doi.org/10.1029/2024GL112415}

[19] S. Rahmstorf, J. E. Box, G. Feulner, M. E. Mann, A. Robinson, S. Rutherford, and E. J. Schaffernicht, "Exceptional twentieth-century slowdown in Atlantic Ocean overturning circulation," *Nature Climate Change*, vol. 5, no. 5, pp. 475–480, 2015.\\
Available: \href{http://doi.org/10.1038/nclimate2554}{http://doi.org/10.1038/nclimate2554}

[20] G. S. Smith and P. D. L. Ritchie, "Modelled arable area for Great Britain under different climate and policy scenarios," *NERC Environmental Data Service*, 2019. 
Available: \href{http://doi.org/10.5285/e1c1dbcf-2f37-429b-af19-a730f98600f6}{http://doi.org/10.5285/e1c1dbcf-2f37-429b-af19-a730f98600f6}

[21] J. Yin, S. M. Griffies, M. Winton, M. Zhao, and L. Zanna, "Response of Storm-Related Extreme Sea Level along the U.S. Atlantic Coast to Combined Weather and Climate Forcing," *Journal of Climate*, 2020.\\
Available:
\href{http://doi.org/10.1175/JCLI-D-19-0551.1}{http://doi.org/10.1175/JCLI-D-19-0551.1}

[22] M. Ben-Yami et al., "Impacts of AMOC Collapse on Monsoon Rainfall: A Multi-Model Comparison," *Earth's Future*, 2024.\\
Available: \href{http://doi.org/10.1029/2023EF003959}{http://doi.org/10.1029/2023EF003959}

[23] F. Schaumann and E. Alastrué de Asenjo, "Weakening AMOC reduces ocean carbon uptake and increases the social cost of carbon," *Proceedings of the National Academy of Sciences*, 2025.\\
Available:
\href{http://doi.org/10.1073/pnas.2419543122}{http://doi.org/10.1073/pnas.2419543122}

[24] M. Ben-Yami, A. Morr, S. Bathiany, and N. Boers, "Uncertainties too large to predict tipping times of major Earth system components from historical data," *Science Advances*, vol. 10, no. 31, 2024. [Online]. \\
Available: \href{http://doi.org/10.1126/sciadv.adl4841}{http://doi.org/10.1126/sciadv.adl4841}

[25] H. Wei, K. Srinivasan, A. L. Stewart, A. Solodoch, G. E. Manucharyan, and A. M. Hogg, "Full-Depth Reconstruction of Long-Term Meridional Overturning Circulation Variability From Satellite-Measurable Quantities via Machine Learning," *Journal of Advances in Modeling Earth Systems*, vol. 17, 2025. [Online]. \\
Available: \href{http://doi.org/10.1029/2024MS004915}{http://doi.org/10.1029/2024MS004915}

[26] D. Swingedouw, M. Houssais, C. Herbaut, A. Blaizot, M. Devilliers, and J. Deshayes, "AMOC Recent and Future Trends: A Crucial Role for Oceanic Resolution and Greenland Melting?," *Frontiers in Climate*, vol. 4, 2022. \\Available: \href{http://doi.org/10.3389/fclim.2022.838310}{http://doi.org/10.3389/fclim.2022.838310}

%% file: sections-v2/amoc_final.tex
\newpage
\section{AMOC Report: Final}
\label{sect:report-final}
\section*{An Assessment of Atlantic Meridional Overturning Circulation (AMOC) Stability: Tipping Points, Uncertainties, and Impacts}

\setcounter{subsection}{0}
The current scientific understanding of Atlantic Meridional Overturning Circulation (AMOC) stability. This assessment will encompass its physical science basis, associated risks, impacts, and vulnerabilities, and relevant mitigation or adaptation strategies, with a particular emphasis on conflicting evidence and scientific uncertainties surrounding potential tipping points and abrupt changes.

\subsection{Introduction}

\subsubsection{Definition and Importance of the AMOC}

The Atlantic Meridional Overturning Circulation (AMOC) is a system of ocean currents in the Atlantic Ocean (extending into the South Atlantic) characterized by a northward flow of warm, saline water in the upper layers of the Atlantic, and a southward flow of colder, deeper waters that are part of the thermohaline circulation [1]. While the AMOC utilizes the Gulf Stream as a pathway for its upper limb, it is important to distinguish the two; the Gulf Stream is primarily a wind-driven western boundary current, and both the wind-driven gyre and the overturning circulation contribute to the northward transport of heat [1].

This circulation system contributes substantially to Atlantic meridional heat transport, with estimates reaching up to ~1.3 PW at low to mid-latitudes [1], [2]. Beyond heat transport, the AMOC plays an important role in the global carbon cycle. Observational data indicate it facilitates the sequestration of anthropogenic carbon and heat in the deep ocean, thereby influencing the vertical and regional distribution of heat within the climate system [1], [3].

\subsubsection{The Emerging Concern: AMOC Instability and Abrupt Climate Change}
The stability of the AMOC has become a focal point of climate research due to its potential non-linear response to forcing [1], [4], [5]. Paleoclimate evidence and theory suggest that the AMOC may admit multiple stable regimes (strong and weak), with transitions possible under sufficient forcing [6]. Transitions between these states—such as those observed during Dansgaard-Oeschger (D-O) events or the Younger Dryas cooling—can occur rapidly on climatic timescales (decades to centuries) [5].

While past abrupt changes were often triggered by glacial meltwater pulses, modern concerns arise from anthropogenic climate change [2]. Accelerated melting of the Greenland Ice Sheet (GrIS), increased precipitation over the North Atlantic and enhanced freshwater fluxes from the Arctic are introducing growing amounts of freshwater into the North Atlantic [7]. This freshening reduces the density of surface waters, inhibiting the deep convection required to sustain the overturning circulation [7].

\subsubsection{Scope and Objectives}

This assessment aims to synthesize the state of knowledge regarding AMOC stability as of the publication date of the latest evidence included in this assessment. It specifically seeks to:
\begin{enumerate}
    \item Assess the physical mechanisms governing the circulation.
    \item Critically evaluate the tension between climate model projections (which generally show gradual weakening) and statistical Early Warning Signals (which suggest approaching tipping points).
   \item Assess the risks and vulnerabilities associated with AMOC decline.
\end{enumerate}

\subsection{The Physical Science Basis of AMOC}
\subsubsection*{Driving Mechanisms}
The AMOC is sustained by a complex interplay of buoyancy (thermohaline) and mechanical (wind-driven) forcing. While often conceptually framed as a ``push'' from high northern latitudes and a ``pull'' from the Southern Ocean, the circulation is largely in geostrophic balance and linked to basin-scale density gradients [8], [1].

\textbf{Buoyancy Forcing and Deep Water Formation (The ``Push''):} The density-driven component involves the formation of North Atlantic Deep Water (NADW). As warm, saline surface water flows northward, heat loss to the atmosphere cools it. This cooling, together with salinity increases from brine rejection during periods of extensive sea-ice formation, enhances water mass densification and can contribute to deep and intermediate water formation, particularly under glacial or near-glacial conditions [8].

\textbf{The Salt-Advection Feedback and Bistability:} The system's potential for bistability is closely linked to the salt-advection feedback. The Atlantic's narrow basin geometry, in combination with atmospheric moisture transport and basin-scale circulation, facilitates a salinity contrast distinct from the Pacific [9], which supports the transport of salt necessary for deep water formation. This positive feedback creates the potential for an abrupt transition to a collapsed 'off' state [10], [46]. However, this bistability is theoretically constrained by mechanical wind-driven forcing. If Southern Ocean wind forcing and mechanically driven upwelling play a dominant role in setting overturning strength and stability—a subject of active debate [8]—they may substantially modify the stability properties inferred from salt-advection-dominated models, reducing the likelihood or parameter range of multiple equilibria.

\textbf{Wind-Driven Forcing and Global Closure (The ``Pull''):} Closing the overturning circulation requires returning dense deep water to the surface. This upwelling is not merely a passive suction but is mechanically driven by two main processes: strong westerly winds in the Southern Ocean induce northward Ekman transport and divergence, effectively lifting deep waters (the Drake Passage effect), while wind- and tide-generated internal waves drive diapycnal mixing in the ocean interior [8].

\textbf{Variability Timescales:} Observations indicate that on shorter timescales (intra-annual), AMOC variability is often strongly influenced by local wind forcing, whereas on interannual to decadal timescales buoyancy forcing becomes increasingly important, with contributions that vary by latitude and AMOC metric [1].

\subsubsection*{Structure and Pathways}
The circulation is organized into two primary limbs, which constitute the Atlantic component of the global thermohaline circulation that transports heat, salt, and biogeochemical tracers:
\begin{itemize}
    \item \textbf{The Upper Limb:} This northward flow, which originates in the South Atlantic and feeds into the Gulf Stream and North Atlantic Current systems, transports warm, saline water in the upper approximately 1,000 meters [12]. Recent modeling indicates this limb also plays a vital role in transporting alkalinity and anthropogenic carbon to subpolar regions, which modulates the ocean's capacity for regional carbon uptake [13].
\item \textbf{The Lower Limb:} The southward return flow consists of cold, dense North Atlantic Deep Water (NADW), primarily traversing the Deep Western Boundary Current (DWBC) between 1,000 and 4,000 meters depth [12]. New observational evidence highlights the DWBC as an intense stream of bioavailable dissolved organic carbon (DOC), exporting it from the subpolar gyre to the deep subtropics where it fuels deep-ocean ecosystems [14].
\end{itemize}

\textbf{Deep Water Formation:} Understanding of deep water formation sites has been refined by recent observational arrays. While deep convection is vigorously active in the Labrador Sea, recent evidence from the OSNAP array suggests that the net transformation of light-to-dense water responsible for overturning occurs primarily in the eastern subpolar gyre (Irminger and Iceland basins) and the Nordic Seas [15]. Estimates of the AMOC's strength from the RAPID array at 26.5°N generally average around 17 Sv (Sverdrup, $1Sv = 10^6m^3 s^{-1}$) [16], while recent hydrographic analyses over the last 90 years estimate the mean strength at approximately 18.4 ± 0.6 Sv [17].

\subsubsection*{AMOC’s Role in the Earth System}
The AMOC plays a fundamental role in regulating the Earth's climate through the redistribution of heat, carbon, and nutrients.

\begin{itemize}
    \item \textbf{Global Heat Redistribution:} The circulation transports significant heat northward. This meridional heat flux contributes substantially to maintaining the relatively mild climate of Northwestern Europe and regulating hemispheric heat balance [8], [1].
    \item \textbf{Carbon Sequestration and Ecosystem Support:} The AMOC acts as a critical pump for both inorganic and organic carbon, linking the surface and deep ocean:
    \begin{itemize}
    \item \textbf{Alkalinity and $CO_2$ Uptake:} The upper limb transports alkalinity northward, which enhances the North Atlantic's buffering capacity. This chemical transport is crucial for maintaining the region's ability to absorb anthropogenic $CO_2$ from the atmosphere [13].
    \item \textbf{Deep Ocean Fueling:} The sinking water masses export bioavailable dissolved organic carbon (DOC) via the Deep Western Boundary Current. This supply fuels deep-sea ecosystems, accounting for up to 38\% of the metabolic oxygen consumption in the deep North Atlantic [14]. 
    \end{itemize}
\item \textbf{ITCZ and Amazon Stability:} The inter-hemispheric energy imbalance created by the AMOC influences the position of the Intertropical Convergence Zone (ITCZ). Recent causal analysis suggests a stabilizing feedback loop: while a strong AMOC pulls the ITCZ northward, a weakening AMOC can shift rainfall patterns in a way that increases dry-season precipitation over the Southern Amazon, potentially offsetting drying trends in that critical biome [18].

\item \textbf{Decadal Variability:} A ``pacemaker region'' in the western transition zone between the subtropical and subpolar gyres has been identified as an important source region for decadal AMOC variability, which in turn influence sea surface temperature patterns that influence climate variability over surrounding landmasses [1].
\end{itemize}

\subsubsection*{Assessment of Evidence and Confidence}

There is \textbf{High Confidence} in the theoretical understanding of the component physical drivers of the AMOC (deep convection, geostrophic flow, and wind-driven transport) and its important role in global heat redistribution. While the general structural pathways are well-established, recent observations have refined the understanding of deep water formation sites, refining earlier views and highlighting the dominant role of the eastern subpolar gyre and Nordic Seas relative to the Labrador Sea [15]. There is increasing evidence supporting the AMOC's critical function in regulating regional carbon and alkalinity budgets [13], [14]. However, there remains \textbf{Medium Confidenc}e regarding the relative energetic contribution of buoyancy versus wind forcing in sustaining the overturning loop, as this remains an area of active research [8].

\subsection{Evidence for AMOC Variability: Past, Present, and Future}

Paleoclimatic Evidence of Past Abrupt Changes
The primary evidence that the AMOC is a bistable system—capable of existing in distinct ``on'' or ``off'' states—comes from the combination of paleoclimate evidence, theory, and modeling. Paleoclimatic records, including ice cores and marine sediments, are consistent with the AMOC crossing stability thresholds (tipping points) that trigger gradual or abrupt state transitions [19], [10]. These records extend our understanding beyond the short instrumental period and provide evidence of ocean-circulation changes during major climate reorganizations.

The Dansgaard–Oeschger events of the last glacial period illustrate the AMOC's capacity for rapid reorganization, featuring repeated, abrupt Northern Hemisphere warmings associated with circulation resurgences [20]. These events operated within a `climate see-saw' pattern, redistributing heat between hemispheres [20], and stood in sharp contrast to Heinrich stadials—periods of strongly reduced overturning driven by massive iceberg discharges and freshwater forcing [19]. This volatility continued through the last deglaciation, which unfolded as a sequence of abrupt transitions. Heinrich Stadial 1 (HS1; 18–14.7 ka) was a pronounced cold interval marked by high iceberg discharge and is widely viewed as a period of substantially reduced overturning [66]. This gave way to the Bølling–Allerød (14.7–12.6 ka), during which rapid North Atlantic warming coincided with a strong recovery of deep-water formation and a rapid reinvigoration of the AMOC, with broadly coherent climate responses across hemispheres as evidenced by multiple paleoceanographic proxies [21], [66].

The Younger Dryas event (approx. 12,900 years ago) serves as a potential analogue for a strongly weakened AMOC, although direct comparisons are limited by differences in boundary conditions (e.g., presence of the Laurentide Ice Sheet) [22]. Recovery from such collapsed states may involve rapid sea-ice feedbacks, such as the opening of polynyas, to restart deep water formation [22]. Modeling studies suggest that intermediate climate conditions may render the AMOC ``less stable'' than in peak glacial or interglacial conditions [66]. However, a long-standing ``meltwater paradox'' persists: most models require much larger freshwater fluxes to weaken the AMOC than those reconstructed from geological evidence of ice-sheet melting [23]. While most evidence for large AMOC reductions comes from glacial intervals, the 8.2 ka event in the early Holocene demonstrates that freshwater forcing can also weaken overturning during a warm interglacial climate, although the freshwater forcing during this event (catastrophic lake drainage) was likely larger than that expected in the near future [66][24].

\subsubsection*{Modern Observational Evidence}

\textbf{Opportunities and Challenges} 

Observing the AMOC involves balancing the accuracy of in-situ arrays with the spatial coverage of satellite data. In-situ arrays provide the gold standard for measuring volume transport but are spatially sparse, expensive, and offer records currently too short to robustly separate long-term anthropogenic trends from multi-decadal natural variability [25], [26]. Earth Observation (EO) technologies help bridge spatial gaps, though challenges remain in monitoring specific freshwater fluxes and transport variables [27], [28]. Moreover, current observational specifications generally do not satisfy the rigorous demands of tipping point early warning systems [28].

\textbf{Current State and Trends
}

Since 2004, the RAPID-MOCHA-WBTS array at 26.5°N has provided continuous, high-fidelity monitoring. This record reveals a weakening trend of approximately 1.0 Sv per decade over the 2004–2023 period [29]. Crucially, recent analysis of this record concludes that the observed weakening is consistent with climate model projections but ``not consistent with a collapse in the mid-21st century,'' providing robust observational evidence against a near-term collapse [29]. This decline, however, was not linear; it was most pronounced between 2004 and 2012, followed by a period of relative stability, though with high interannual variability strongly influenced by wind forcing [25], [30]. Additionally, the OSNAP array has highlighted that the majority of deep water transformation occurs in the eastern subpolar gyre, challenging earlier assumptions focused on the Labrador Sea [31]. Bridging the gap between sparse historical data and modern arrays, methods combining satellite altimetry with cable measurements have reconstructed variability back to 1993, aligning with the variability observed by arrays [32], [33].

\textbf{Indirect Proxies and Fingerprints
}

Due to the short duration of direct measurements, researchers rely on proxies. Reconstructions based on Sea Surface Temperature (SST), identifying the characteristic ``warming hole'' in the North Atlantic, suggest the AMOC has weakened by approximately 15\% since the mid-20th century and is in its weakest state in a millennium [34], [35]. However, surface signals can be contaminated by atmospheric ``noise'' [40]. Alternative reconstructions using subsurface density data, which capture deep circulation structure more directly, generally find no overall decline between 1981 and 2016, instead capturing a multi-decadal cycle [30]. Furthermore, hydrographic analyses extending back to the 1930s find no statistically significant decline in AMOC strength (estimating a decrease of only 0.4 ± 0.6 Sv with low confidence), which contradicts the significant weakening inferred from SST proxies [17].

New evidence challenges the robustness of SST-based proxies. Terhaar et al. (2025) show using 24 CMIP6 Earth System Models that subpolar SST anomalies ("warming holes") cannot robustly reconstruct the AMOC due to atmospheric noise and other factors. Instead, they propose that air-sea heat flux anomalies are a more physically robust proxy, tightly linked to AMOC strength through energy conservation on decadal timescales. Applying this method to ERA5 and JRA-55 reanalysis data, they reconstruct the AMOC at 26.5°N from 1963 to 2017 and find no decline, but rather substantial decadal variability [36]. This finding aligns with hydrographic evidence [17], [30] and clarifies that hydrographic data supports stability since the 1980s, but cannot address the mid-20th century decline directly inferred from SST.

To separate anthropogenic weakening from natural variability, recent analyses distinguish between a long-term ``Trend Mode,'' which exhibits the characteristic cooling pattern and a downward trajectory, and the Atlantic Multidecadal Oscillation (AMO) [37].

Beyond these spatial patterns, recent statistical analyses have identified signals consistent with critical slowing down in eight independent AMOC indices derived from observational SST and salinity data. These indices exhibit robust signs of critical slowing down (increasing variance and autocorrelation), suggesting a potential loss of stability over the last century [38].

\subsubsection*{Modeled Historical Evolution and Future Projections}

\textbf{Historical Evolution and Natural Variability:} Comparing observational proxies with climate models reveals discrepancies in the historical evolution of the AMOC. While SST-based fingerprints suggest a significant weakening since the mid-20th century [34], historical climate model simulations often depict a more stable circulation dominated by natural variability [39], [40]. For instance, analyses of CMIP6 historical simulations suggest that the multimodel mean AMOC strengthened from 1850 to 1985 due to anthropogenic aerosol forcing, contrasting with the weakening inferred from SST fingerprints [41]. Other studies argue that natural multidecadal variability has dominated since 1900, with models predicting only a minor slowing (approx. 1 Sv) after 1980 [39]. This divergence highlights uncertainties in separating forced signals from internal oscillations [42].

\textbf{Future Projections and Model Fidelity:} Despite historical discrepancies, CMIP6 models consistently project a weakening of the AMOC over the 21st century under all emission scenarios [54]. However, the magnitude varies significantly ($24\%$ to $39\%$ decline under high-emission scenarios). Standard resolution GCMs often suffer from biases, such as excessive stability due to poor representation of freshwater runoff spreading. Conversely, recent high-resolution models suggest resilience via eddy compensation mechanisms. However, recent studies using strongly eddying ocean-only models demonstrate that AMOC collapse is still possible under freshwater forcing, though eddies may maintain a weak ($\sim5 Sv$) residual circulation in the collapsed state [44].

\subsubsection*{Assessment of Evidence and Confidence}

While direct observations confirm a weakening trend in the early 21st century [25], there is \textbf{Low Confidence} in attributing a long-term decline solely to anthropogenic forcing relative to the pre-industrial era. This lower confidence reflects the growing divergence between SST-based proxies, which suggest a 15\% weakening [34], and a suite of alternative evidence including subsurface reconstructions [30], historical model simulations [39], hydrographic analyses [17], and physically-constrained heat flux inversions [36], all of which indicate stability or the dominance of natural variability since the mid-20th century. However, there is \textbf{Medium Confidence} that some indicators suggest a potential loss of stability (approaching a bifurcation point), evidenced by statistical early warning signals [38]. Furthermore, there is \textbf{High Confidence} that the AMOC has undergone rapid changes during the last glacial period and the deglaciation [6], [66].

\subsection{The Tipping Point Debate: Conflicting Evidence and Scientific Uncertainties}

The most contentious aspect of current AMOC science is the existence and proximity to a ``tipping point"—a critical threshold at which the system bifurcates from a strong mode (characterized by active overturning and heat transport) to a weak/off mode (where deep water formation is strongly reduced and circulation weakens substantially).

\subsubsection*{The Concept of Bistability and Hysteresis}

The potential for a tipping point arises from the non-linear dynamics of the salt-advection feedback. In the active ``on'' state, the AMOC transports salty water into the North Atlantic, maintaining the high surface density required for deep convection. A tipping point occurs when a perturbation—such as surface freshening or warming—weakens the circulation sufficiently that this salt transport is reduced, creating a self-amplifying feedback loop. If the salt transport drops below a critical threshold, the feedback reverses, preventing deep water formation and driving the system into a collapsed state [1], [46]. While this bistability implies hysteresis—meaning the pathway to recovery requires reducing forcing significantly below the collapse threshold—the core risk is the state shift triggered by the failure of the salt-advection pump [46].

\subsubsection*{Evidence Suggesting an Approaching Tipping Point}

Recent studies utilizing statistical physics have suggested an increased risk of a collapse within this century.

\begin{itemize}
    \item \textbf{Early Warning Signals (EWS):} As a complex system approaches a tipping point, it exhibits ``critical slowing down,'' characterized by increased variance and increased lag-1 autocorrelation in time-series data [45], [48]. Analyses of SST and salinity indices—including work by Boers (2021) and recent estimates by Ditlevsen \& Ditlevsen [38], [48]—suggest these signals may be present and statistically significant, with some studies estimating a potential collapse window within the 21st century [48].
\item \textbf{Dansgaard-Oeschger (DO) events:} Some general support, based on proxy data that a tipping of the AMOC might be detectable by prior EWS comes from DO events. DO events are characterized by millennial-scale abrupt climate changes during the glacial period, leading to abrupt temperature shifts in the North Atlantic region. Studies of climate bifurcation during the deglaciation suggest that such tipping points were approached in the past, providing qualitative support for the plausibility of EWS approaches [5].
\item \textbf{Overly Stable Models:} Physics-based indicators derived from observations suggest that many CMIP models are biased toward stability (the ``salt-advection feedback'' is modeled too weakly). Correcting for this bias suggests the real-world AMOC may be closer to a stability threshold than the model ensemble average indicates [45], [49].
\end{itemize}

\subsubsection*{Evidence Suggesting Greater Stability}
Contrasting the EWS studies, climate models generally project a more stable evolution.

\begin{itemize}

\item \textbf{Climate Model Projections:} The Coupled Model Intercomparison Project Phase 6 (CMIP6) models consistently project a weakening of the AMOC over the 21st century under all emission scenarios [45]. However, the magnitude varies significantly, with decline estimates ranging from 24\% to 39\% under high-emission scenarios. This consensus supports the assessment that while the AMOC will weaken, an abrupt collapse before 2100 is very unlikely (IPCC AR6 [50]).

\item \textbf{Observational Constraints:} Observations from the RAPID array suggest the recent weakening is consistent with climate models and not indicative of a collapse in the mid-21st century [29]. Furthermore, high-resolution models with realistic eddy fields often show low variability and a stable regime.

\item \textbf{High-Resolution Modeling:} Eddy-resolving models tend to show a more resilient circulation, suggesting that tipping behavior in some intermediate-complexity models may be sensitive to resolution [45].
\item \textbf{Stabilizing Feedback:} Mechanisms such as the warming of the deep ocean and changes in the gyre circulation may counteract the freshening signal.

\item \textbf{Critique of EWS:} Critics argue that EWS in climate data can be generated by changes in external forcing or ``red noise'' processes unrelated to bifurcation, leading to false positives [47]. Recent modeling by Zimmerman et al. [47] shows that Critical Slowing Down (CSD) indicators can robustly increase even in monostable regimes where no tipping point exists, suggesting these signals may not reliably distinguish between a slowing and a collapsing circulation.
\end{itemize}

\subsubsection*{Key Scientific Uncertainties}

The divergence in assessment stems from four deep uncertainties:

\textbf{Freshwater Hosing Plausibility and Model Bias}

A critical uncertainty is the magnitude of freshwater input required to trigger a collapse. There is a significant discrepancy between the stability of standard climate models and those corrected for known biases.

\begin{itemize}
    \item \textbf{Stability Bias: }Standard Global Circulation Models (GCMs) often exhibit a ``stability bias'' because they poorly represent the salt-advection feedback [45]. Consequently, these models often require extreme, potentially unrealistic freshwater ``hosing'' fluxes to induce a shutdown.
\item \textbf{Realistic Thresholds:} When corrected for this bias to match observed salinity budgets, the AMOC becomes significantly more sensitive. Research indicates that in bias-corrected models, the forcing generated by a doubling of $CO_2$ may be sufficient to trigger a collapse in some bias-corrected models, without the need for extreme external hosing [45], [49].
\end{itemize}

\textbf{Comparison of AMOC Stability Thresholds}

\begin{center}
    \captionof*{table}{Table: Comparison of Model Stability}
    \label{tab:simple_comparison}
    \small 
    \begin{tabularx}{\textwidth}{lXXX}
        \toprule
        \textbf{Model Type} & \textbf{Stability Characteristics} & \textbf{Trigger for Collapse} & \textbf{Plausibility of Trigger} \\
        \midrule
        
        \textbf{Standard GCMs} & 
        High Stability: Salt-advection feedback is modeled too weakly [45]. & 
        Extreme Hosing: Requires large, artificial freshwater fluxes often exceeding realistic melt rates. & 
        Low: The required fluxes may be physically implausible under standard climate change scenarios. \\
        \addlinespace 
        
        \textbf{Bias-Corrected} & 
        Realistic Sensitivity: Adjusted to match observed Atlantic salinity distribution [45, 49]. & 
        Climate Forcing: Can collapse under realistic warming scenarios (e.g., doubling of CO$_2$) due to internal hydrological changes. & 
        High: The trigger conditions fall within the range of expected future greenhouse gas forcing. \\
        
        \bottomrule
    \end{tabularx}
    \end{center}



\textbf{Model Fidelity and Biases
}

Standard resolution GCMs often suffer from biases, such as the inability to resolve overflow processes and mesoscale eddies accurately [45]. Conversely, recent high-resolution models that resolve ocean eddies suggest the AMOC may be more stable, attributing this resilience to eddy compensation mechanisms [45]. However, recent studies using strongly eddying ocean-only models demonstrate that AMOC collapse is still possible under freshwater forcing, though eddies may maintain a weak ($\sim5$ Sv) residual circulation in the collapsed state [44]. This creates a divergence in the evidence base regarding the precise stability threshold and collapsed state characteristics.

\textbf{Signal vs. Noise
}

Distinguishing a trend toward a tipping point from natural multidecadal variability or forced response remains difficult [47]. While statistical analyses suggest an increased risk of collapse within this century [48], critics argue that these signals can be generated by ``red noise'' processes or changes in external forcing unrelated to bifurcation, raising concerns that these methods may yield false positives [47].

\textbf{Timing and Delays
}

Even if the AMOC tipping threshold is crossed, the timing of the subsequent collapse remains highly uncertain due to the ``slow passage effect.'' Kim et al. [52] demonstrate that under time-varying freshwater forcing, the system can exhibit a significant lag—potentially up to 1,300 years—between passing the bifurcation point and the actual circulation collapse. This implies that observation-based EWS might detect the threshold crossing long before the full climatic impacts fully manifest [52].

\subsubsection*{Assessment of Evidence and Confidence}

There is \textbf{Low Confidence} in the timing of a potential collapse. While the IPCC assessment remains the standard, the recent emergence of EWS studies based on statistical physics [48] and bias-corrected model analyses [49] suggests the risk is non-negligible and potentially underestimated [45]. There is \textbf{Medium Confidence} that the risk of crossing a tipping point increases non-linearly with cumulative emissions of greenhouse gases.

\subsection{Impacts, Risks, and Vulnerabilities of AMOC Decline}

The potential decline of the Atlantic Meridional Overturning Circulation (AMOC) poses systemic and severe risks to the climate system, ecosystems, and human societies. Assessing these risks requires distinguishing between the impacts of a gradual slowdown, which has been observed over parts of the instrumental period and is projected to continue in most models, and the profound global climate shifts that would result from an abrupt collapse. Such a collapse represents a critical transition or ``tipping point,'' which could occur if self-amplifying feedbacks, such as the salt-advection mechanism, become dominant [45].

\subsubsection*{Consequences of Gradual Weakening}

A gradual weakening of the AMOC alters heat distribution, sea-level patterns, and ocean biogeochemistry, creating a baseline of risk even without a full collapse.

\begin{itemize}
\item \textbf{Ocean Circulation and Biogeochemistry:} A declining AMOC disrupts broader ocean circulation patterns, leading to nutrient depletion in surface waters and reducing the efficiency of the biological carbon pump. These changes affect marine productivity and carbon storage [45].
\item \textbf{Regional Cooling and Hydroclimate Extremes:} While the AMOC transports heat northward, its decline leads to a ``bipolar seesaw'' pattern with cooling in the Northern Hemisphere and slight warming in the Southern Hemisphere. Simulations and paleoclimatic data indicate that the AMOC contributes substantially to maintaining air temperatures over the North Atlantic and Europe 5°C to 8°C warmer than they would be in its absence [45]. Consequently, a weakening or collapse would induce a significant cooling tendency, estimated to involve seasonal temperature drops of 5°C to 8°C in winter [2], [49], [45], which in a future warming climate would compete with the background greenhouse gas-induced warming, potentially leading to reduced warming rates or localized cooling depending on the emission scenario [45]. Additionally, recent weakening trends and the associated ``cold blob'' of sea surface temperatures in the North Atlantic have been linked to summer heatwaves and drought conditions in Europe [1], [41]. This connection arises as AMOC-driven SST anomalies influence atmospheric circulation patterns, such as the jet stream, thereby increasing the frequency of extreme weather events over surrounding landmasses [1].
\item \textbf{Dynamic Sea-Level Rise:} The weakening circulation disrupts the geostrophic balance of the Gulf Stream, causing water to pile up along the North American coast. Models project this could contribute an additional 15–20 cm of sea-level rise to the U.S. East Coast by 2100, significantly increasing the frequency and severity of coastal flooding [53].
\item \textbf{Marine Ecosystems:} The reduction in overturning circulation alters nutrient supply to the surface ocean, leading to declines in Total System Biomass (TSB). Ecosystem modeling suggests that on top of anthropogenic climate change, TSB decreases by roughly 2\% to 4\% due to AMOC weakening, with regional declines reaching up to 30\%, disproportionately affecting high trophic levels [54].
\end{itemize}

\subsubsection*{Consequences of Abrupt Collapse}

Crossing a tipping point into a collapsed AMOC state would result in rapid and potentially long-lasting global climate shifts.

\begin{itemize}
    \item \textbf{Severe Cooling:} In a collapse scenario, the cessation of northward heat transport would lead to drastic cooling across the Northern Hemisphere. This would radically alter habitability and energy demand in regions such as Scandinavia and Great Britain [45].
\item \textbf{Tropical Rainfall Shifts:} The resulting inter-hemispheric energy imbalance would tend to shift the Intertropical Convergence Zone (ITCZ) southward. This shift is projected to cause severe drying in the West African Monsoon and Indian Summer Monsoon regions, with rainfall reductions of approximately 29\% and 19\% respectively [55].
\item \textbf{Amazon Stabilization vs. Destabilization:} While some models suggest a southward ITCZ shift could increase dry-season precipitation over the Southern Amazon, potentially stabilizing the rainforest [18], broader assessments warn that AMOC collapse could interact with other tipping elements, increasing the risk of cascading destabilization in the Amazon and ice sheets [55], [56].
\end{itemize}

\subsubsection*{Key Risk Assessment}

The interaction of physical hazards with exposed and vulnerable systems creates specific key risks. The following risk assessment adopts a multi-criteria framework consistent with IPCC approaches, evaluating hazards, exposure, vulnerability, magnitude of impacts, and limits to adaptation for key AMOC-related risks.

\textbf{
Key Risk 1: Global Food Security via Monsoon Disruption
}
\begin{itemize}
   
 \item \textbf{Hazard \& Exposure:} An abrupt AMOC collapse shifts the thermal equator and the ITCZ southward. This shift strongly affects the West African Monsoon (WAM) and Indian Summer Monsoon (ISM), regions that support billions of people [55].
 \item \textbf{Vulnerability:} Agricultural systems in West Africa and India are highly sensitive to AMOC-driven shifts in the ITCZ due to their heavy dependence on rain-fed agriculture.
\item \textbf{Magnitude:} Multi-model comparisons indicate substantial disruptions, with ensemble mean annual rainfall changes of -29\% for WAM and -19\% for ISM [55]. In Northern Europe, severe cooling could render current arable farming practices unviable; modeling for Great Britain suggests an AMOC collapse could reduce the area of arable land from 32\% to 7\% due to climatic unsuitability [57].
 \item \textbf{Adaptation Limits:} While technological adaptations such as widespread irrigation could theoretically ameliorate some effects, the costs and water requirements may be prohibitive, as demonstrated in assessments for Great Britain [57].
\end{itemize}

\textbf{Key Risk 2: Systemic Economic Impact via Carbon Feedback
}
\begin{itemize}
 \item \textbf{Hazard:} A weakening AMOC reduces the ocean's capacity to uptake carbon. Simulations reveal an approximately linear relationship between AMOC strength and carbon uptake reductions, constituting a positive carbon cycle feedback [58].
 \item \textbf{Vulnerability:} the global economy is vulnerable to the accelerated warming resulting from higher atmospheric $CO_2$ concentrations.
 \item \textbf{Magnitude:} This ``AMOC carbon feedback'' could increase the Social Cost of Carbon (SCC) by roughly 1\%. When incorporated into integrated climate-economy models, this feedback leads to additional global economic damages estimated in the trillions of dollars [58].
\end{itemize}

\textbf{Key Risk 3: Cascading Earth System Interactions
}

\begin{itemize}
    \item \textbf{Hazard:} The AMOC acts as a mediator in the global network of tipping elements. Its collapse alters global heat distribution, potentially triggering cascading interactions across multiple tipping elements [56].
    \item \textbf{Vulnerability:} Major tipping elements, including the Greenland Ice Sheet, West Antarctic Ice Sheet, and Amazon rainforest, are sensitive to temperature and precipitation changes driven by the AMOC.
    \item \textbf{Magnitude:} Interactions tend to destabilize the network. While specific feedbacks like increased rainfall in the Southern Amazon might offer local stabilization [18], the broader risk involves global domino effects that could accelerate ice sheet loss and ecosystem transitions [56].
\end{itemize}

\textbf{Key Risk 4: Coastal Infrastructure and Communities via Sea-Level Rise
}
\begin{itemize}
    \item \textbf{Hazard \& Exposure:} The weakening of the AMOC disrupts the geostrophic balance of the Gulf Stream, causing water to pile up against the North American coast. This dynamic sea-level rise (DSL) acts on top of global mean sea-level rise [53].
    \item \textbf{Vulnerability:} Major population centers and infrastructure along the U.S. East Coast and the Gulf of Mexico are particularly vulnerable to this ``hotspot'' of sea-level rise [53].
    \item \textbf{Magnitude:} Modeling indicates that the anthropogenic signal in extreme sea levels can emerge quickly, with dynamic sea level contributing significantly to flood risk. For example, along the Northeast Coast, dynamic sea level rise is projected to reach up to ~20 cm by 2100 under high emission scenarios, significantly reducing the return period of extreme flooding events [53].
    \item \textbf{Adaptation Limits:} Response strategies involve a trade-off between hard protection (e.g., sea walls), which requires substantial public investment and engineering, and managed retreat, which entails significant social and economic costs for affected communities [67].
\end{itemize}

\subsection{Response Strategies}

Given the deep uncertainty regarding the proximity of an AMOC tipping point, strategies must prioritize resilience, robustness, and risk avoidance.

\textbf{Strategies for Gradual Weakening
}

\begin{itemize}
    \item \textbf{Adaptation to Gradual Change:} Strategies for gradual weakening focus on robust adaptation planning that avoids lock-ins while managing uncertainty [59]. In the agricultural sector, farmers and agribusinesses can implement robust adaptation planning, such as adjusting crop varieties; modeling indicates that under standard climate change scenarios involving gradual AMOC weakening, arable farming in regions like Great Britain remains largely viable with conventional adaptations [57]. Similarly, fisheries managers can implement sectoral adjustments, such as shifting quotas and targeting different species [54]. Conventional options also include coastal management agencies reinforcing coastal defenses against dynamic sea-level rise [53].
\end{itemize}
\textbf{
Strategies for a Potential Abrupt Collapse
}
\begin{itemize}
    \item \textbf{Transformational Adaptation:} A tipping point scenario necessitates transformative options beyond incremental adjustments. This includes national governments planning for radical shifts in land suitability, such as the potential cessation of arable farming in parts of Northern Europe [57].
\item \textbf{Emergency Contingencies:} Strategies involve developing contingencies for rapid climatic shifts that could overwhelm standard disaster risk management systems, particularly regarding food security in monsoon-dependent regions [55].
\end{itemize}

\textbf{Robust Strategies
}

\begin{itemize}
    \item\textbf{Flexible Adaptation Pathways:} To address deep uncertainty, policymakers should develop adaptation pathways that remain viable across a range of AMOC futures, from gradual weakening to collapse [59].
    \item\textbf{Scenario Broadening:} Assessments suggest broadening planning scenarios to include high-impact, low-probability events like AMOC collapse, ensuring that long-term infrastructure and policy decisions are robust to extreme outcomes [59].
    \item\textbf{Monitoring and Early Warning: }Continuous monitoring is crucial for detecting early warning signals (EWS). Recent statistical analyses have suggested the possibility of a collapse as early as mid-century [48], although these estimates remain controversial due to uncertainties in historical data and model assumptions [60].
    \item\textbf{Mitigation:} The most effective strategy to minimize the probability of crossing a tipping point is the rapid reduction of greenhouse gas emissions. Phasing out fossil fuels limits the freshwater forcing from ice melt that destabilizes the AMOC [45].
\end{itemize}

\subsection{Conclusion and Priorities for Research and Assessment}

\subsubsection*{Synthesis of Knowledge}

Multiple lines of observational and proxy evidence suggest that the AMOC may have weakened relative to the mid-20th century [45], [61]. However, this interpretation is challenged by hydrographic reconstructions [17], [30] and air-sea heat flux inferences [36] which suggest the circulation has remained stable over the last 30 to 90 years. Assessing these conflicting perspectives, while longer-term proxy records [61] suggest a weakening, recent reconstructions based on air-sea heat fluxes [36] and hydrographic data [17], [30] find no decline since the mid-20th century. This suggests that the ``warming hole'' signal used in earlier proxies may be less tightly coupled to AMOC strength than previously thought [36]. While recent research highlights the significant role of natural multidecadal variability in observed signals [39], ``fingerprint'' analyses and aerosol-forcing studies indicate that human activity is acting as a force multiplier on the system [45], [62]. Furthermore, a critical discrepancy exists regarding stability: while some high-resolution models depict a stable regime [51], [45], observation-based early warning signals suggest the system may be losing resilience [38]. Assessment of these conflicting lines of evidence suggests that models may currently underestimate the risk of abrupt change due to potential structural biases [63], [64].

In summary, it is \textbf{very likely} that the AMOC will weaken over the 21st century [50]. Furthermore, it is considered \textbf{very unlikely} that an abrupt collapse (tipping point) will occur before 2100 based on the current climate model ensemble consensus [50]. However, there is \textbf{Low Confidence} in the precise timing and magnitude of future changes given potential biases in current models [45], [49] and recent statistical Early Warning Signals (EWS) suggesting the system may be closer to a stability threshold than some models indicate.

\subsubsection*{Priorities for Strengthening Future Assessments}

\begin{itemize}
    \item \textbf{Observation:} Sustain and expand the RAPID and OSNAP arrays [12] with a specific focus on distinguishing long-term anthropogenic trends from multi-decadal natural variability [39]. Integrate these with Deep Argo and satellite gravimetry to reconstruct broader circulation patterns [65].
    \item \textbf{Modeling:} Investigate the structural causes of AMOC stability in climate models. While some high-resolution, eddy-resolving models suggest the system is more stable than previously thought [51], others indicate it remains vulnerable to freshwater forcing [63]. Resolving this discrepancy is urgent.
    \item \textbf{Theory:} Move beyond purely statistical Early Warning Signals, which are prone to false positives [47], by developing and monitoring physics-based indicators, such as the freshwater transport at the southern boundary of the Atlantic [49].
    \item \textbf{Impact Assessment:} Couple climate models simulating AMOC collapse with Integrated Assessment Models (IAMs) to quantify economic and social risks, particularly regarding the social cost of carbon [58].
\end{itemize}

\subsection{Final Remarks}

Although the precise attribution of recent weakening is complex and debated [39], [61], [30], the theoretical robustness of AMOC bistability [45], [64] supports the physical plausibility of abrupt collapse. Given the catastrophic global impacts of such an event, the inability of current models to rule out this risk necessitates a precautionary approach. Consequently, mitigation—the rapid reduction of greenhouse gas emissions—remains the only robust strategy to minimize the probability of crossing a potential tipping point [45].

\textbf{References:}
[1] M. W. Buckley and J. Marshall, ``Observations, inferences, and mechanisms of the Atlantic Meridional Overturning Circulation: A review,'' Reviews of Geophysics, 2016.
[Online]. Available: \url{https://doi.org/10.1002/2015RG000493}


[2] W. Broeker, ``The great ocean conveyor,'' Oceanography, 1991. [Online]. Available: \url{https://tos.org/oceanography/assets/docs/4-2\_broecker.pdf}


[3] V. Caínzos, A. Velo, F. F. Pérez, and A. Hernández‐Guerra, ``Anthropogenic Carbon Transport Variability in the Atlantic Ocean Over Three Decades,'' Global Biogeochemical Cycles, 2022. [Online]. Available: \url{https://doi.org/10.1029/2022GB007475}


[4] J. van den Berk, S. Drijfhout, and W. Hazeleger, ``Characterisation of Atlantic meridional overturning hysteresis using Langevin dynamics,'' Earth System Dynamics, 2021. [Online]. Available: \url{https://doi.org/10.5194/esd-12-69-2021}


[5] T. M. Lenton, V. Livina, V. Dakos, and M. Scheffer, ``Climate bifurcation during the last deglaciation?,'' Climate of the past, 2012. [Online]. Available: \url{https://doi.org/10.5194/cp-8-1127-2012}


[6] S. Rahmstorf, ``Ocean circulation and climate during the past 120,000 years,'' Nature, 2002. [Online]. Available: \url{https://www.nature.com/articles/nature01090}


[7] Q. Yang et al., ``Recent increases in Arctic freshwater flux affects Labrador Sea convection and Atlantic overturning circulation,'' Nature Communications, 2016. [Online]. Available: \url{https://doi.org/10.1038/ncomms10525}


[8] T. Kuhlbrodt et al., ``On the driving processes of the Atlantic meridional overturning circulation,'' Reviews of Geophysics, 2007. [Online]. Available: \url{https://doi.org/10.1029/2004RG000166}


[9] B. Sinha et al., ``Mountain ranges favour vigorous Atlantic meridional overturning,'' Geophysical Research Letters, 2011. [Online]. Available: \url{https://doi.org/10.1029/2011gl050485}


[10] W. Weijer et al., ``Stability of the Atlantic Meridional Overturning Circulation: A Review and Synthesis,'' Journal of Geophysical Research Oceans, 2019. [Online]. Available: \url{https://doi.org/10.1029/2019jc015083}


[11] H. Stommel, ``Thermohaline Convection with Two Stable Regimes of Flow,'' Tellus A Dynamic Meteorology and Oceanography, 1961. [Online]. Available: \url{https://doi.org/10.1111/j.2153-3490.1961.tb00079.x}


[12] E. Frajka‐Williams, N. P. Foukal, and G. Danabasoglu, ``Should AMOC observations continue: how and why?,'' Philosophical Transactions of the Royal Society A Mathematical Physical and Engineering Sciences, 2023. [Online]. Available: \url{https://doi.org/10.1098/rsta.2022.0195}


[13] Q. Zhang, T. Ito, and A. Bracco, ``Modulation of regional carbon uptake by AMOC and alkalinity changes in the subpolar North Atlantic under a warming climate,'' Frontiers in Marine Science, 2024. [Online]. Available: \url{https://doi.org/10.3389/fmars.2024.1304193}


[14] M. Fontela, F. F. Pérez, H. Mercier, and P. Lherminier, ``North Atlantic Western Boundary Currents Are Intense Dissolved Organic Carbon Streams,'' Frontiers in Marine Science, 2020. [Online]. Available: \url{https://doi.org/10.3389/fmars.2020.593757}


[15] T. Petit, M. S. Lozier, S. A. Josey, and S. A. Cunningham, ``Atlantic Deep Water Formation Occurs Primarily in the Iceland Basin and Irminger Sea by Local Buoyancy Forcing,'' Geophysical Research Letters, 2020. [Online]. Available: \url{https://doi.org/10.1029/2020gl091028}


[16] W. E. Johns et al., ``Towards two decades of Atlantic Ocean mass and heat transports at 26.5° N,'' Philosophical Transactions of the Royal Society A Mathematical Physical and Engineering Sciences, 2023. [Online]. Available: \url{https://doi.org/10.1098/rsta.2022.0188}


[17] T. Rossby, J. B. Palter, and K. Donohue, ``What Can Hydrography Between the New England Slope, Bermuda and Africa Tell us About the Strength of the AMOC Over the Last 90 years?,'' Geophysical Research Letters, 2022. [Online]. Available: \url{https://doi.org/10.1029/2022gl099173}


[18] A. Högner et al., ``Causal pathway from AMOC to Southern Amazon rainforest indicates stabilising interaction between two climate tipping elements,'' Environmental Research Letters, 2025. [Online]. Available: \url{https://doi.org/10.1088/1748-9326/addb62}


[19] A. H. L. Voelker and J. T. Andrews, ``Millennial-Scale Ocean Climate Variability,'' Encyclopedia of Ocean Sciences (Third Edition), Academic Press, 2018. [Online]. Available: \url{https://doi.org/10.1016/B978-0-12-409548-9.11368-5}


[20] North Greenland Ice Core Project members, ``High-resolution record of Northern Hemisphere climate extending into the last interglacial period,'' Nature, 2004. [Online]. Available: \url{https://doi.org/10.1038/nature02805}


[21] T. Chen et al., ``Synchronous centennial abrupt events in the ocean and atmosphere during the last deglaciation,'' Science, 2015. [Online]. Available: \url{https://doi.org/10.1126/science.aac6159}


[22] J. Velay‐Vitow, D. Chandan, and W. R. Peltier, ``Into the Holocene, anatomy of the Younger Dryas cold reversal and preboreal oscillation,'' Scientific Reports, 2024. [Online]. Available: \url{https://doi.org/10.1038/s41598-024-53591-2}


[23] B. Snoll et al., ``A multi-model assessment of the early last deglaciation (PMIP4 LDv1): a meltwater perspective,'' Climate of the past, 2024. [Online]. Available: \url{https://doi.org/10.5194/cp-20-789-2024}


[24] G. Rush et al., ``The magnitude and source of meltwater forcing of the 8.2 ka climate event constrained by relative sea-level data from eastern Scotland,'' Quaternary Science Advances, 2023. [Online]. Available: \url{https://doi.org/10.1016/j.qsa.2023.100119}


[25] B. Moat et al., ``Pending recovery in the strength of the meridional overturning circulation at 26° N,'' Ocean science, 2020. [Online]. Available: \url{https://doi.org/10.5194/os-16-863-2020}


[26] D. Lobelle et al., ``Detectability of an AMOC Decline in Current and Projected Climate Changes,'' Geophysical Research Letters, 2020. [Online]. Available: \url{https://doi.org/10.1029/2020gl089974}


[27] R. Wood et al., ``Opportunities for Earth Observation to Inform Risk Management for Ocean Tipping Points,'' Surveys in Geophysics, 2024. [Online]. Available: \url{https://doi.org/10.1007/s10712-024-09859-3}


[28] S. Loriani et al., ``Monitoring the Multiple Stages of Climate Tipping Systems from Space: Do the GCOS Essential Climate Variables Meet the Needs?,'' Surveys in Geophysics, 2025. [Online]. Available: \url{https://doi.org/10.1007/s10712-024-09866-4}


[29] G. McCarthy et al., ``Signal and Noise in the Atlantic Meridional Overturning Circulation at 26°N,'' Geophysical Research Letters, 2025. [Online]. Available: \url{https://doi.org/10.1029/2025gl115055}


[30] E. Worthington et al., ``A 30-year reconstruction of the Atlantic meridional overturning circulation shows no decline,'' Ocean science, 2021. [Online]. Available: \url{https://doi.org/10.5194/os-17-285-2021}


[31] F. Li et al., ``Subpolar North Atlantic western boundary density anomalies and the Meridional Overturning Circulation,'' Nature Communications, 2021. [Online]. Available: \url{https://doi.org/10.1038/s41467-021-23350-2}


[32] E. Frajka‐Williams, ``Estimating the Atlantic overturning at 26°N using satellite altimetry and cable measurements,'' Geophysical Research Letters, 2015. [Online]. Available: \url{https://doi.org/10.1002/2015gl063220}


[33] A. Sanchez‐Franks, E. Frajka‐Williams, B. Moat, and D. Smeed, ``A dynamically based method for estimating the Atlantic meridional overturning circulation at 26° N from satellite altimetry,'' Ocean science, 2021. [Online]. Available: \url{https://doi.org/10.5194/os-17-1321-2021}


[34] L. Caesar, S. Rahmstorf, A. Robinson, G. Feulner, and V. S. Saba, ``Observed fingerprint of a weakening Atlantic Ocean overturning circulation,'' Nature, 2018. [Online]. Available: \url{https://doi.org/10.1038/s41586-018-0006-5}


[35] D. Thornalley et al., ``Anomalously weak Labrador Sea convection and Atlantic overturning during the past 150 years,'' Nature, 2018. [Online]. Available: \url{https://doi.org/10.1038/s41586-018-0007-4}


[36] J. Terhaar, L. Vogt, and N. P. Foukal, ``Atlantic overturning inferred from air-sea heat fluxes indicates no decline since the 1960s,'' Nature Communications, 2025. [Online]. Available: \url{https://doi.org/10.1038/s41467-024-55297-5}


[37] Dima, M., Lohmann, G., Ionita, M. et al., ``AMOC modes linked with distinct North Atlantic deep water formation sites,'' Clim Dyn, 2022. [Online]. Available: \url{https://doi.org/10.1007/s00382-022-06156-w}


[38] N. Boers, ``Observation-based early-warning signals for a collapse of the Atlantic Meridional Overturning Circulation,'' Nature Climate Change, 2021. [Online]. Available: \url{https://doi.org/10.1038/s41558-021-01097-4}


[39] M. Latif, J. Sun, M. Visbeck, and M. H. Bordbar, ``Natural variability has dominated Atlantic Meridional Overturning Circulation since 1900,'' Nature Climate Change, 2022. [Online]. Available: \url{https://doi.org/10.1038/s41558-022-01342-4}


[40] G. D. McCarthy and L. Caesar, ``Can we trust projections of AMOC weakening based on climate models that cannot reproduce the past?,'' Philosophical Transactions of the Royal Society A Mathematical Physical and Engineering Sciences, 2023. [Online]. Available: \url{https://doi.org/10.1098/rsta.2022.0193}


[41] M. Menary et al., ``Aerosol‐Forced AMOC Changes in CMIP6 Historical Simulations,'' Geophysical Research Letters, 2020. [Online]. Available: \url{https://doi.org/10.1029/2020gl088166}


[42] J. Robson, R. Sutton, M. Menary, and M. W. K. Lai, ``Contrasting internally and externally generated Atlantic Multidecadal Variability and the role for AMOC in CMIP6 historical simulations,'' Philosophical Transactions of the Royal Society A Mathematical Physical and Engineering Sciences, 2023. [Online]. Available: \url{https://doi.org/10.1098/rsta.2022.0194}


[43] Dima, M., Lohmann, G., Nichita, DR. et al., ``Structural stability changes of the Atlantic Meridional Overturning Circulation,'' npj Clim Atmos Sci, 2025. [Online]. Available: \url{https://doi.org/10.1038/s41612-025-00960-x}


[44] R. M. van Westen, M. Kliphuis, and H. A. Dijkstra, ``Collapse of the Atlantic Meridional Overturning Circulation in a Strongly Eddying Ocean‐Only Model,'' Geophysical Research Letters, 2025. [Online]. Available: \url{https://doi.org/10.1029/2024GL114532}


[45] S. Rahmstorf, ``Is the Atlantic Overturning Circulation Approaching a Tipping Point?,'' Oceanography, 2024. [Online]. Available: \url{https://doi.org/10.5670/oceanog.2024.501}


[46] M. Hofmann and S. Rahmstorf, ``On the stability of the Atlantic meridional overturning circulation,'' Proceedings of the National Academy of Sciences, 2009. [Online]. Available: \url{https://doi.org/10.1073/pnas.0909146106}


[47] C. Zimmerman, T. J. W. Wagner, E. Maroon, and D. E. McNamara, ``Slowed Response of Atlantic Meridional Overturning Circulation Not a Robust Signal of Collapse,'' Geophysical Research Letters, 2025. [Online]. Available: \url{https://doi.org/10.1029/2024GL112415}


[48] P. Ditlevsen and S. Ditlevsen, ``Warning of a forthcoming collapse of the Atlantic meridional overturning circulation,'' Nature Communications, 2023. [Online]. Available: \url{https://doi.org/10.1038/s41467-023-39810-w}


[49] R. M. van Westen, M. Kliphuis, and H. A. Dijkstra, ``Physics-based early warning signal shows that AMOC is on tipping course,'' Science Advances, 2024. [Online]. Available: \url{https://doi.org/10.1126/sciadv.adk1189}


[50] Intergovernmental Panel on Climate Change, ``Climate Change 2021: The Physical Science Basis,'' Cambridge University Press eBooks, 2023. [Online]. Available: \url{https://doi.org/10.1017/9781009157896}


[51] L. Gerber, J. Lippold, F. Süfke, O. Valk, P. Testorf, M. Ehnis, S. Tautenhahn, L. Max, C. M. Chiessi, M. Regelous, S. Szidat, O. Friedrich, F. Pöppelmeier, ``Low variability of the Atlantic Meridional Overturning Circulation throughout the Holocene.'' Nat Commun, 2025. [Online]. Available: \url{https://pmc.ncbi.nlm.nih.gov/articles/PMC12284099/}


[52] S. Kim, H. Kim, H. A. Dijkstra, and S. An, ``Slow and soft passage through tipping point of the Atlantic Meridional Overturning Circulation in a changing climate,'' npj Climate and Atmospheric Science, 2022. [Online]. Available: \url{https://doi.org/10.1038/s41612-022-00236-8}


[53] J. Yin, S. M. Griffies, M. Winton, M. Zhao, and L. Zanna, ``Response of Storm-Related Extreme Sea Level along the U.S. Atlantic Coast to Combined Weather and Climate Forcing,'' Journal of Climate, 2020. [Online]. Available: \url{https://doi.org/10.1175/JCLI-D-19-0551.1}


[54] A. Boot, J. Steenbeek, M. Coll, A. S. von der Heydt, and H. A. Dijkstra, ``Global Marine Ecosystem Response to a Strong AMOC Weakening Under Low and High Future Emission Scenarios,'' Earth's Future, 2025. [Online]. Available: \url{https://doi.org/10.1029/2024EF004741}


[55] M. Ben‐Yami et al., ``Impacts of AMOC Collapse on Monsoon Rainfall: A Multi‐Model Comparison,'' Earth's Future, 2024. [Online]. Available: \url{https://doi.org/10.1029/2023EF003959}


[56] N. Wunderling, J. F. Donges, J. Kurths, and R. Winkelmann, ``Interacting tipping elements increase risk of climate domino effects under global warming,'' Earth System Dynamics, 2021. [Online]. Available: \url{https://esd.copernicus.org/articles/12/601/2021/esd-12-601-2021.pdf}


[57] Smith, G. S., Ritchie, P.D.L., ``Modelled arable area for Great Britain under different climate and policy scenarios,'' NERC Environmental Data Service, 2019. [Online]. Available: \url{https://doi.org/10.5285/e1c1dbcf-2f37-429b-af19-a730f98600f6}


[58] F. Schaumann and E. Alastrué de Asenjo, ``Weakening AMOC reduces ocean carbon uptake and increases the social cost of carbon,'' Proceedings of the National Academy of Sciences, 2025. [Online]. Available: \url{https://doi.org/10.1073/pnas.2419543122}


[59] R. Biesbroek, M. Haasnoot, K. J. Mach, and A. C. Petersen, ``Adaptation planning in the context of a weakening and possibly collapsing Atlantic Meridional Overturning Circulation (AMOC),'' Regional Environmental Change, 2025. [Online]. Available: \url{https://doi.org/10.1007/s10113-025-02434-5}


[60] M. Ben-Yami, A. Morr, S. Bathiany, N Boers, ``Uncertainties too large to predict tipping times of major Earth system components from historical data,'' Sci. Adv., 2024. [Online]. Available: \url{https://doi.org/10.1126/sciadv.adl4841}


[61] L. Caesar, G. McCarthy, D. Thornalley, N. Cahill, and S. Rahmstorf, ``Current Atlantic Meridional Overturning Circulation weakest in last millennium,'' Nature Geoscience, 2021. [Online]. Available: \url{https://doi.org/10.1038/s41561-021-00699-z}


[62] T. Hassan, R. J. Allen, W. Liu, and C. A. Randles, ``Anthropogenic aerosol forcing of the Atlantic meridional overturning circulation and the associated mechanisms in CMIP6 models,'' Atmospheric chemistry and physics, 2021. [Online]. Available: \url{https://doi.org/10.5194/acp-21-5821-2021}


[63] D. Swingedouw et al., ``AMOC Recent and Future Trends: A Crucial Role for Oceanic Resolution and Greenland Melting?,'' Frontiers in Climate, 2022. [Online]. Available: \url{https://doi.org/10.3389/fclim.2022.838310}


[64] E. Hawkins et al., ``Bistability of the Atlantic overturning circulation in a global climate model and links to ocean freshwater transport,'' Geophysical Research Letters, 2011. [Online]. Available: \url{https://doi.org/10.1029/2011gl047208}


[65] H. Wei et al., ``Full‐Depth Reconstruction of Long‐Term Meridional Overturning Circulation Variability From Satellite‐Measurable Quantities via Machine Learning,'' Journal of Advances in Modeling Earth Systems, 2025. [Online]. Available: \url{https://doi.org/10.1029/2024MS004915}


[66] S.K.V. Hines, N.P. Foukal, K.M. Costa, D.W. Oppo, O. Marchal, L.D. Keigwin, and A. Condron. ``Is there robust evidence for freshwater-driven AMOC changes? A synthesis of data, models, and mechanisms.'' Oceanography 38(3), 2025.  [Online]. Available: \url{https://doi.org/10.5670/oceanog.2025.e301}


[67] Glavovic, B.C., R. Dawson, W. Chow, M. Garschagen, M. Haasnoot, C. Singh, and A. Thomas, ``Cross-Chapter Paper 2: Cities and Settlements by the Sea.'' Climate Change 2022: Impacts, Adaptation and Vulnerability. Contribution of Working Group II to the Sixth Assessment Report of the Intergovernmental Panel on Climate Change, 2022. [Online]. Available: \url{https://doi.org/10.1017/9781009325844.019}


%% file: main.bbl
\begin{thebibliography}{64}
\providecommand{\natexlab}[1]{#1}
\providecommand{\url}[1]{\texttt{#1}}
\expandafter\ifx\csname urlstyle\endcsname\relax
  \providecommand{\doi}[1]{doi: #1}\else
  \providecommand{\doi}{doi: \begingroup \urlstyle{rm}\Url}\fi

\bibitem[Adler and de~Alfaro(2007)]{Adler-Alfaro-2007}
B.~T. Adler and L.~de~Alfaro.
\newblock A content-driven reputation system for the wikipedia.
\newblock In \emph{Proceedings of the 16th International Conference on World Wide Web}, WWW '07. Association for Computing Machinery, 2007.
\newblock \doi{10.1145/1242572.1242608}.
\newblock URL \url{https://doi.org/10.1145/1242572.1242608}.

\bibitem[Allen and Olkin(1999)]{Allen1999}
I.~E. Allen and I.~Olkin.
\newblock Estimating time to conduct a meta-analysis from number of citations retrieved.
\newblock \emph{JAMA}, 282\penalty0 (7):\penalty0 634--635, 1999.

\bibitem[Anderson et~al.(2024)Anderson, Shah, and Kreminski]{barrett2024}
B.~R. Anderson, J.~H. Shah, and M.~Kreminski.
\newblock Homogenization effects of large language models on human creative ideation.
\newblock In \emph{Proceedings of the 16th Conference on Creativity \& Cognition}, C\&C '24, page 413–425, New York, NY, USA, 2024. Association for Computing Machinery.
\newblock ISBN 9798400704857.
\newblock \doi{10.1145/3635636.3656204}.
\newblock URL \url{https://doi.org/10.1145/3635636.3656204}.

\bibitem[Asai et~al.(2024)Asai, He, Shao, Shi, Singh, Chang, Lo, Soldaini, Feldman, D'arcy, et~al.]{asai2024openscholar}
A.~Asai, J.~He, R.~Shao, W.~Shi, A.~Singh, J.~C. Chang, K.~Lo, L.~Soldaini, S.~Feldman, M.~D'arcy, et~al.
\newblock Openscholar: Synthesizing scientific literature with retrieval-augmented lms.
\newblock \emph{arXiv preprint arXiv:2411.14199}, 2024.

\bibitem[Baumberger et~al.(2017)Baumberger, Knutti, and Hirsch~Hadorn]{baumberger2017}
C.~Baumberger, R.~Knutti, and G.~Hirsch~Hadorn.
\newblock Building confidence in climate model projections: an analysis of inferences from fit.
\newblock \emph{WIREs Climate Change}, 8\penalty0 (3):\penalty0 e454, 2017.
\newblock \doi{https://doi.org/10.1002/wcc.454}.
\newblock URL \url{https://wires.onlinelibrary.wiley.com/doi/abs/10.1002/wcc.454}.

\bibitem[Beel et~al.(2025)Beel, Kan, and Baumgart]{beel2025evaluating_arxiv}
J.~Beel, M.-Y. Kan, and M.~Baumgart.
\newblock Evaluating sakana's ai scientist: Bold claims, mixed results, and a promising future?, 2025.
\newblock URL \url{https://arxiv.org/abs/2502.14297}.

\bibitem[Bird et~al.(2009)Bird, Loper, and Klein]{nltk2009}
S.~Bird, E.~Loper, and E.~Klein.
\newblock \emph{Natural Language Processing with {Python}}.
\newblock O'Reilly Media Inc., 2009.

\bibitem[Boiko et~al.(2023)Boiko, MacKnight, Kline, and Gomes]{boiko2023autonomous}
D.~A. Boiko, R.~MacKnight, B.~Kline, and G.~Gomes.
\newblock Autonomous chemical research with large language models.
\newblock \emph{Nature}, 624\penalty0 (7992):\penalty0 570--578, 2023.

\bibitem[Borah et~al.(2017)Borah, Brown, Capers, and Kaiser]{Borah2017}
R.~Borah, A.~W. Brown, P.~L. Capers, and K.~A. Kaiser.
\newblock Analysis of the time and workers needed to conduct systematic reviews of medical interventions using data from the prospero registry.
\newblock \emph{BMJ Open}, 7\penalty0 (2):\penalty0 e012545, 2017.
\newblock \doi{10.1136/bmjopen-2016-012545}.

\bibitem[Borton(1970)]{Borton1970}
T.~Borton.
\newblock \emph{Reach, Touch, and Teach: Student Concerns and Process Education}.
\newblock McGraw-Hill, New York, 1970.

\bibitem[Bran et~al.(2024)Bran, Cox, White, and Schwaller]{bran2024chemcrow}
A.~M. Bran, S.~Cox, A.~D. White, and P.~Schwaller.
\newblock Augmenting large language models with chemistry tools.
\newblock \emph{Nature Machine Intelligence}, 6:\penalty0 525--535, 2024.

\bibitem[Brockriede and Ehninger(1960)]{brockriede1960toulmin}
W.~Brockriede and D.~Ehninger.
\newblock Toulmin on argument: An interpretation and application.
\newblock \emph{Quarterly Journal of Speech}, 46\penalty0 (1):\penalty0 44--53, 1960.

\bibitem[Bulian et~al.(2024)Bulian, Sch\"{a}fer, Amini, Lam, Ciaramita, Gaiarin, Chen~Huebscher, Buck, Mede, Leippold, and Strauss]{bulian2024}
J.~Bulian, M.~S. Sch\"{a}fer, A.~Amini, H.~Lam, M.~Ciaramita, B.~Gaiarin, M.~Chen~Huebscher, C.~Buck, N.~G. Mede, M.~Leippold, and N.~Strauss.
\newblock Assessing large language models on climate information.
\newblock In R.~Salakhutdinov, Z.~Kolter, K.~Heller, A.~Weller, N.~Oliver, J.~Scarlett, and F.~Berkenkamp, editors, \emph{Proceedings of the 41st International Conference on Machine Learning}, volume 235 of \emph{Proceedings of Machine Learning Research}, pages 4884--4935. PMLR, 21--27 Jul 2024.
\newblock URL \url{https://proceedings.mlr.press/v235/bulian24a.html}.

\bibitem[Cao et~al.(2025)Cao, Arora, Cento, Manta, Farahani, Cecere, Selemon, Sang, Gong, Kloosterman, et~al.]{cao2025automation}
C.~Cao, R.~Arora, P.~Cento, K.~Manta, E.~Farahani, M.~Cecere, A.~Selemon, J.~Sang, L.~X. Gong, R.~Kloosterman, et~al.
\newblock Automation of systematic reviews with large language models.
\newblock \emph{medRxiv}, pages 2025--06, 2025.

\bibitem[Cappello et~al.(2025)Cappello, Madireddy, Underwood, Getty, Chia, Ramachandra, Nguyen, Keceli, Mallick, Li, et~al.]{cappello2025eaira}
F.~Cappello, S.~Madireddy, R.~Underwood, N.~Getty, N.~L.-P. Chia, N.~Ramachandra, J.~Nguyen, M.~Keceli, T.~Mallick, Z.~Li, et~al.
\newblock Eaira: Establishing a methodology for evaluating {AI} models as scientific research assistants.
\newblock \emph{arXiv preprint arXiv:2502.20309}, 2025.

\bibitem[Chen(2025)]{202504.2088}
F.~Chen.
\newblock Beyond scaling laws: Towards scientific reasoning-driven llm architectures.
\newblock \emph{Preprints}, April 2025.
\newblock \doi{10.20944/preprints202504.2088.v1}.
\newblock URL \url{https://doi.org/10.20944/preprints202504.2088.v1}.

\bibitem[Cleland(2002)]{cleland2002methodological}
C.~E. Cleland.
\newblock Methodological and epistemic differences between historical science and experimental science.
\newblock \emph{Philosophy of Science}, 69\penalty0 (3):\penalty0 474--496, 2002.

\bibitem[Dell'Acqua et~al.(2023)Dell'Acqua, McFowland~III, Mollick, Lifshitz-Assaf, Kellogg, Rajendran, Krayer, Candelon, and Lakhani]{dell2023navigating}
F.~Dell'Acqua, E.~McFowland~III, E.~R. Mollick, H.~Lifshitz-Assaf, K.~Kellogg, S.~Rajendran, L.~Krayer, F.~Candelon, and K.~R. Lakhani.
\newblock Navigating the jagged technological frontier: Field experimental evidence of the effects of {AI} on knowledge worker productivity and quality.
\newblock \emph{Harvard Business School Technology \& Operations Mgt. Unit Working Paper}, 24-013, 2023.

\bibitem[Domfeh and Dancy(2025)]{Human_AI_DRR}
E.~A. Domfeh and C.~L. Dancy.
\newblock Human-{AI} use patterns for decision-making in disaster scenarios: A systematic review.
\newblock In \emph{2025 IEEE International Symposium on Technology and Society (ISTAS)}, pages 1--10, 2025.
\newblock \doi{10.1109/ISTAS65609.2025.11269624}.

\bibitem[Doshi and Hauser(2024)]{doshi2024generative}
A.~R. Doshi and O.~Hauser.
\newblock Generative {AI} enhances individual creativity but reduces the collective diversity of novel content.
\newblock \emph{Science Advances}, 10\penalty0 (28):\penalty0 eadn5290, 2024.

\bibitem[Du{\'e}{\~n}ez-Guzm{\'a}n et~al.(2025)Du{\'e}{\~n}ez-Guzm{\'a}n, Comanescu, Mao, McKee, Coppin, Sadedin, Chiappa, Vezhnevets, Bakker, Bachrach, et~al.]{duenez2025perceptual}
E.~A. Du{\'e}{\~n}ez-Guzm{\'a}n, R.~Comanescu, Y.~Mao, K.~R. McKee, B.~Coppin, S.~Sadedin, S.~Chiappa, A.~S. Vezhnevets, M.~A. Bakker, Y.~Bachrach, et~al.
\newblock Perceptual interventions ameliorate statistical discrimination in learning agents.
\newblock \emph{Proceedings of the National Academy of Sciences}, 122\penalty0 (25):\penalty0 e2319933121, 2025.

\bibitem[Edwards(2013)]{edwards2013vast}
P.~N. Edwards.
\newblock \emph{A vast machine: Computer models, climate data, and the politics of global warming}.
\newblock Mit press, 2013.

\bibitem[Faigley and Witte(1981)]{Faigley-Witte-1981}
L.~Faigley and S.~Witte.
\newblock Analyzing revision.
\newblock \emph{College Composition and Communication}, 32\penalty0 (4):\penalty0 400--414, 1981.

\bibitem[Gottweis et~al.(2025)Gottweis, Weng, Daryin, Tu, Palepu, Sirkovic, Myaskovsky, Weissenberger, Rong, Tanno, Saab, Popovici, Blum, Zhang, Chou, Hassidim, Gokturk, Vahdat, Kohli, Matias, Carroll, Kulkarni, Tomasev, Guan, Dhillon, Vaishnav, Lee, Costa, Penadés, Peltz, Xu, Pawlosky, Karthikesalingam, and Natarajan]{gottweis2025}
J.~Gottweis, W.-H. Weng, A.~Daryin, T.~Tu, A.~Palepu, P.~Sirkovic, A.~Myaskovsky, F.~Weissenberger, K.~Rong, R.~Tanno, K.~Saab, D.~Popovici, J.~Blum, F.~Zhang, K.~Chou, A.~Hassidim, B.~Gokturk, A.~Vahdat, P.~Kohli, Y.~Matias, A.~Carroll, K.~Kulkarni, N.~Tomasev, Y.~Guan, V.~Dhillon, E.~D. Vaishnav, B.~Lee, T.~R.~D. Costa, J.~R. Penadés, G.~Peltz, Y.~Xu, A.~Pawlosky, A.~Karthikesalingam, and V.~Natarajan.
\newblock Towards an {AI} co-scientist, 2025.
\newblock URL \url{https://arxiv.org/abs/2502.18864}.

\bibitem[Gridach et~al.(2025)Gridach, Nanavati, Mack, Abidine, and Mendes]{gridach2025agentic}
M.~Gridach, J.~Nanavati, C.~Mack, K.~Z.~E. Abidine, and L.~Mendes.
\newblock Agentic {AI} for scientific discovery: A survey of progress, challenges, and future directions.
\newblock In \emph{Towards Agentic {AI} for Science: Hypothesis Generation, Comprehension, Quantification, and Validation}, 2025.
\newblock URL \url{https://openreview.net/forum?id=TyCYakX9BD}.

\bibitem[Hao et~al.(2026)Hao, Xu, Li, and Evans]{Hao2026}
Q.~Hao, F.~Xu, Y.~Li, and J.~Evans.
\newblock Artificial intelligence tools expand scientists’ impact but contract science’s focus.
\newblock \emph{Nature}, jan 2026.
\newblock ISSN 1476-4687.
\newblock \doi{10.1038/s41586-025-09922-y}.
\newblock URL \url{https://doi.org/10.1038/s41586-025-09922-y}.

\bibitem[Hartung(2025)]{hartung2025ai}
T.~Hartung.
\newblock {AI}, agentic models and lab automation for scientific discovery--the beginning of scaince.
\newblock \emph{Frontiers in Artificial Intelligence}, 8:\penalty0 1649155, 2025.

\bibitem[Jin et~al.(2024)Jin, Zhao, Wang, et~al.]{jin2024agentreview}
Y.~Jin, Q.~Zhao, Y.~Wang, et~al.
\newblock Agentreview: Exploring peer review dynamics with llm agents.
\newblock In \emph{Proceedings of the 2024 Conference on Empirical Methods in Natural Language Processing (EMNLP)}, 2024.

\bibitem[Karpukhin et~al.(2020)Karpukhin, Oguz, Min, Lewis, Wu, Edunov, Chen, and Yih]{karpukhin-etal-2020-dense}
V.~Karpukhin, B.~Oguz, S.~Min, P.~Lewis, L.~Wu, S.~Edunov, D.~Chen, and W.-t. Yih.
\newblock Dense passage retrieval for open-domain question answering.
\newblock In B.~Webber, T.~Cohn, Y.~He, and Y.~Liu, editors, \emph{Proceedings of the 2020 Conference on Empirical Methods in Natural Language Processing (EMNLP)}, pages 6769--6781, Online, Nov. 2020. Association for Computational Linguistics.
\newblock \doi{10.18653/v1/2020.emnlp-main.550}.
\newblock URL \url{https://aclanthology.org/2020.emnlp-main.550/}.

\bibitem[Kittur and Kraut(2008)]{Kittur-Kraut-2008}
A.~Kittur and R.~E. Kraut.
\newblock Harnessing the wisdom of crowds in wikipedia: quality through coordination.
\newblock In \emph{Proceedings of the 2008 ACM Conference on Computer Supported Cooperative Work}, CSCW '08, New York, NY, USA, 2008. Association for Computing Machinery.
\newblock \doi{10.1145/1460563.1460572}.
\newblock URL \url{https://doi.org/10.1145/1460563.1460572}.

\bibitem[Knüsel et~al.(2019)Knüsel, Zumwald, Baumberger, Hirsch~Hadorn, Fischer, Bresch, and Knutti]{Knusel2019}
B.~Knüsel, M.~Zumwald, C.~Baumberger, G.~Hirsch~Hadorn, E.~M. Fischer, D.~N. Bresch, and R.~Knutti.
\newblock Applying big data beyond small problems in climate research.
\newblock \emph{Nature Climate Change}, 9\penalty0 (3):\penalty0 196--202, mar 2019.
\newblock ISSN 1758-6798.
\newblock \doi{10.1038/s41558-019-0404-1}.
\newblock URL \url{https://doi.org/10.1038/s41558-019-0404-1}.

\bibitem[Kochkov et~al.(2024)Kochkov, Yuval, Langmore, Norgaard, et~al.]{kochkov2024neural}
D.~Kochkov, J.~Yuval, I.~Langmore, P.~Norgaard, et~al.
\newblock Neural general circulation models for weather and climate.
\newblock \emph{Nature}, 632:\penalty0 1060--1066, 2024.

\bibitem[Lam et~al.(2023)Lam, Sanchez-Gonzalez, Willson, et~al.]{lam2023learning}
R.~Lam, A.~Sanchez-Gonzalez, M.~Willson, et~al.
\newblock Learning skillful medium-range global weather forecasting.
\newblock \emph{Science}, 382\penalty0 (6677):\penalty0 1416--1421, 2023.

\bibitem[Lee et~al.(2025)Lee, Chen, Dua, Cer, Shanbhogue, Naim, Ábrego, Li, Chen, Vera, Ren, Zhang, Salz, Boratko, Han, Chen, Huang, Rao, Suganthan, Han, Doumanoglou, Gupta, Moiseev, Yip, Jain, Baumgartner, Shahi, Gomez, Mariserla, Choi, Shah, Goenka, Chen, Xia, Chen, Duddu, Chen, Walker, Zhou, Ghiya, Gleicher, Gill, Dong, Seyedhosseini, Sung, Hoffmann, and Duerig]{geminiembedding2025}
J.~Lee, F.~Chen, S.~Dua, D.~Cer, M.~Shanbhogue, I.~Naim, G.~H. Ábrego, Z.~Li, K.~Chen, H.~S. Vera, X.~Ren, S.~Zhang, D.~Salz, M.~Boratko, J.~Han, B.~Chen, S.~Huang, V.~Rao, P.~Suganthan, F.~Han, A.~Doumanoglou, N.~Gupta, F.~Moiseev, C.~Yip, A.~Jain, S.~Baumgartner, S.~Shahi, F.~P. Gomez, S.~Mariserla, M.~Choi, P.~Shah, S.~Goenka, K.~Chen, Y.~Xia, K.~Chen, S.~M.~K. Duddu, Y.~Chen, T.~Walker, W.~Zhou, R.~Ghiya, Z.~Gleicher, K.~Gill, Z.~Dong, M.~Seyedhosseini, Y.~Sung, R.~Hoffmann, and T.~Duerig.
\newblock Gemini embedding: Generalizable embeddings from gemini, 2025.
\newblock URL \url{https://arxiv.org/abs/2503.07891}.

\bibitem[Lee et~al.(2022)Lee, Liang, and Yang]{coauthor2022}
M.~Lee, P.~Liang, and Q.~Yang.
\newblock Coauthor: Designing a human-ai collaborative writing dataset for exploring language model capabilities.
\newblock In \emph{Proceedings of the 2022 CHI Conference on Human Factors in Computing Systems}, CHI '22, New York, NY, USA, 2022. Association for Computing Machinery.
\newblock ISBN 9781450391573.
\newblock \doi{10.1145/3491102.3502030}.
\newblock URL \url{https://doi.org/10.1145/3491102.3502030}.

\bibitem[Lin(2004)]{lin-2004-rouge}
C.-Y. Lin.
\newblock {ROUGE}: A package for automatic evaluation of summaries.
\newblock In \emph{Text Summarization Branches Out}, pages 74--81, Barcelona, Spain, July 2004. Association for Computational Linguistics.
\newblock URL \url{https://aclanthology.org/W04-1013/}.

\bibitem[Lu et~al.(2024)Lu, Cong, et~al.]{lu2024aiscientist}
C.~Lu, Y.~Cong, et~al.
\newblock The {AI} scientist: Towards fully automated open-ended scientific discovery.
\newblock \emph{arXiv preprint arXiv:2408.06292}, 2024.

\bibitem[Lucas(1976)]{lucas1976econometric}
R.~E. Lucas.
\newblock Econometric policy evaluation: A critique.
\newblock In \emph{Carnegie-Rochester conference series on public policy}, volume~1, pages 19--46. North-Holland, 1976.

\bibitem[Mann and Thompson(1988)]{Mann1988RhetoricalST}
W.~C. Mann and S.~A. Thompson.
\newblock Rhetorical structure theory: Toward a functional theory of text organization.
\newblock \emph{Text \& Talk}, 8:\penalty0 243 -- 281, 1988.
\newblock URL \url{https://api.semanticscholar.org/CorpusID:60514661}.

\bibitem[Messeri and Crockett(2024)]{messeri2024artificial}
L.~Messeri and M.~J. Crockett.
\newblock Artificial intelligence and illusions of understanding in scientific research.
\newblock \emph{Nature}, 627\penalty0 (8002):\penalty0 49--58, 2024.

\bibitem[Moore et~al.(2025)Moore, Mann, and Chen]{moore2025automated}
S.~A. Moore, B.~P. Mann, and B.~Chen.
\newblock Automated global analysis of experimental dynamics through low-dimensional linear embeddings.
\newblock \emph{npj Complexity}, 2\penalty0 (1), 2025.

\bibitem[Oreskes(2004)]{oreskes2004scientific}
N.~Oreskes.
\newblock The scientific consensus on climate change.
\newblock \emph{Science}, 306\penalty0 (5702):\penalty0 1686--1686, 2004.

\bibitem[Papineni et~al.(2002)Papineni, Roukos, Ward, and Zhu]{papineni2002}
K.~Papineni, S.~Roukos, T.~Ward, and W.-J. Zhu.
\newblock Bleu: a method for automatic evaluation of machine translation.
\newblock In \emph{Proceedings of the 40th Annual Meeting on Association for Computational Linguistics}, ACL '02, page 311–318, USA, 2002. Association for Computational Linguistics.
\newblock \doi{10.3115/1073083.1073135}.
\newblock URL \url{https://doi.org/10.3115/1073083.1073135}.

\bibitem[Patel et~al.(2025)Patel, Arabzadeh, Gupta, Sundar, Stoica, Zaharia, and Guestrin]{patel2025deepscholar}
L.~Patel, N.~Arabzadeh, H.~Gupta, A.~Sundar, I.~Stoica, M.~Zaharia, and C.~Guestrin.
\newblock Deepscholar-bench: A live benchmark and automated evaluation for generative research synthesis.
\newblock \emph{arXiv preprint arXiv:2508.20033}, 2025.

\bibitem[Perez et~al.(2023)Perez, Ringer, Lukosiute, Nguyen, Chen, Heiner, Pettit, Olsson, Kundu, Kadavath, Jones, Chen, Mann, Israel, Seethor, McKinnon, Olah, Yan, Amodei, Amodei, Drain, Li, Tran-Johnson, Khundadze, Kernion, Landis, Kerr, Mueller, Hyun, Landau, Ndousse, Goldberg, Lovitt, Lucas, Sellitto, Zhang, Kingsland, Elhage, Joseph, Mercado, DasSarma, Rausch, Larson, McCandlish, Johnston, Kravec, El~Showk, Lanham, Telleen-Lawton, Brown, Henighan, Hume, Bai, Hatfield-Dodds, Clark, Bowman, Askell, Grosse, Hernandez, Ganguli, Hubinger, Schiefer, and Kaplan]{perez-etal-2023-discovering}
E.~Perez, S.~Ringer, K.~Lukosiute, K.~Nguyen, E.~Chen, S.~Heiner, C.~Pettit, C.~Olsson, S.~Kundu, S.~Kadavath, A.~Jones, A.~Chen, B.~Mann, B.~Israel, B.~Seethor, C.~McKinnon, C.~Olah, D.~Yan, D.~Amodei, D.~Amodei, D.~Drain, D.~Li, E.~Tran-Johnson, G.~Khundadze, J.~Kernion, J.~Landis, J.~Kerr, J.~Mueller, J.~Hyun, J.~Landau, K.~Ndousse, L.~Goldberg, L.~Lovitt, M.~Lucas, M.~Sellitto, M.~Zhang, N.~Kingsland, N.~Elhage, N.~Joseph, N.~Mercado, N.~DasSarma, O.~Rausch, R.~Larson, S.~McCandlish, S.~Johnston, S.~Kravec, S.~El~Showk, T.~Lanham, T.~Telleen-Lawton, T.~Brown, T.~Henighan, T.~Hume, Y.~Bai, Z.~Hatfield-Dodds, J.~Clark, S.~R. Bowman, A.~Askell, R.~Grosse, D.~Hernandez, D.~Ganguli, E.~Hubinger, N.~Schiefer, and J.~Kaplan.
\newblock Discovering language model behaviors with model-written evaluations.
\newblock In A.~Rogers, J.~Boyd-Graber, and N.~Okazaki, editors, \emph{Findings of the Association for Computational Linguistics: ACL 2023}, pages 13387--13434, Toronto, Canada, July 2023. Association for Computational Linguistics.
\newblock \doi{10.18653/v1/2023.findings-acl.847}.
\newblock URL \url{https://aclanthology.org/2023.findings-acl.847/}.

\bibitem[Rahmstorf(2024)]{Rahmstorf-Oceanography-2024}
S.~Rahmstorf.
\newblock Is the atlantic overturning circulation approaching a tipping point?
\newblock \emph{Oceanography}, 37\penalty0 (3), September 2024.
\newblock URL \url{https://doi.org/10.5670/oceanog.2024.501}.

\bibitem[Reimers and Gurevych(2019)]{reimers-2019-sentence-bert}
N.~Reimers and I.~Gurevych.
\newblock Sentence-bert: Sentence embeddings using siamese bert-networks.
\newblock In \emph{Proceedings of the 2019 Conference on Empirical Methods in Natural Language Processing}. Association for Computational Linguistics, 11 2019.
\newblock URL \url{https://arxiv.org/abs/1908.10084}.

\bibitem[Robertson and Zaragoza(2009)]{bm25}
S.~Robertson and H.~Zaragoza.
\newblock The probabilistic relevance framework: Bm25 and beyond.
\newblock \emph{Found. Trends Inf. Retr.}, 3\penalty0 (4):\penalty0 333–389, Apr. 2009.
\newblock ISSN 1554-0669.
\newblock \doi{10.1561/1500000019}.
\newblock URL \url{https://doi.org/10.1561/1500000019}.

\bibitem[Romera-Paredes et~al.(2024)Romera-Paredes, Barekatain, Novikov, et~al.]{romeraparedes2024mathematical}
B.~Romera-Paredes, M.~Barekatain, A.~Novikov, et~al.
\newblock Mathematical discoveries from program search with large language models.
\newblock \emph{Nature}, 625:\penalty0 468--477, 2024.

\bibitem[Shao et~al.(2025)Shao, Wang, Qian, Pan, Liu, and Wang]{shao2025sciscigpt}
E.~Shao, Y.~Wang, Y.~Qian, Z.~Pan, H.~Liu, and D.~Wang.
\newblock Sciscigpt: advancing human--{AI} collaboration in the science of science.
\newblock \emph{Nature Computational Science}, pages 1--15, 2025.

\bibitem[Sharma et~al.(2024)Sharma, Tong, Korbak, Duvenaud, Askell, Bowman, DURMUS, Hatfield-Dodds, Johnston, Kravec, Maxwell, McCandlish, Ndousse, Rausch, Schiefer, Yan, Zhang, and Perez]{sharmatowards}
M.~Sharma, M.~Tong, T.~Korbak, D.~Duvenaud, A.~Askell, S.~R. Bowman, E.~DURMUS, Z.~Hatfield-Dodds, S.~R. Johnston, S.~M. Kravec, T.~Maxwell, S.~McCandlish, K.~Ndousse, O.~Rausch, N.~Schiefer, D.~Yan, M.~Zhang, and E.~Perez.
\newblock Towards understanding sycophancy in language models.
\newblock In \emph{The Twelfth International Conference on Learning Representations}, 2024.
\newblock URL \url{https://openreview.net/forum?id=tvhaxkMKAn}.

\bibitem[Shaw and Nave(2026)]{shaw2026thinking}
S.~D. Shaw and G.~Nave.
\newblock Thinking---fast, slow, and artificial: How ai is reshaping human reasoning and the rise of cognitive surrender.
\newblock Research paper, The Wharton School, Jan 2026.
\newblock URL \url{https://ssrn.com/abstract=6097646}.
\newblock Available at SSRN and OSF (\url{https://doi.org/10.31234/osf.io/yk25n_v1}).

\bibitem[Summerfield et~al.(2025)Summerfield, Argyle, Bakker, Collins, Durmus, Eloundou, Gabriel, Ganguli, Hackenburg, Hadfield, et~al.]{summerfield2025impact}
C.~Summerfield, L.~P. Argyle, M.~Bakker, T.~Collins, E.~Durmus, T.~Eloundou, I.~Gabriel, D.~Ganguli, K.~Hackenburg, G.~K. Hadfield, et~al.
\newblock The impact of advanced {AI} systems on democracy.
\newblock \emph{Nature Human Behaviour}, pages 1--11, 2025.

\bibitem[Tessler et~al.(2024)Tessler, Bakker, Jarrett, Sheahan, Chadwick, Koster, Evans, Campbell-Gillingham, Collins, Parkes, et~al.]{tessler2024ai}
M.~H. Tessler, M.~A. Bakker, D.~Jarrett, H.~Sheahan, M.~J. Chadwick, R.~Koster, G.~Evans, L.~Campbell-Gillingham, T.~Collins, D.~C. Parkes, et~al.
\newblock {AI} can help humans find common ground in democratic deliberation.
\newblock \emph{Science}, 386\penalty0 (6719):\penalty0 eadq2852, 2024.

\bibitem[Teufel and Moens(2002)]{teufel-moens-2002-articles}
S.~Teufel and M.~Moens.
\newblock Summarizing scientific articles: Experiments with relevance and rhetorical status.
\newblock \emph{Computational Linguistics}, 28\penalty0 (4):\penalty0 409--445, 2002.
\newblock \doi{10.1162/089120102762671936}.
\newblock URL \url{https://aclanthology.org/J02-4002/}.

\bibitem[Toulmin(1958)]{toulmin1958}
S.~E. Toulmin.
\newblock \emph{The Uses of Argument}.
\newblock Cambridge University Press, Cambridge, 1958.

\bibitem[Vi\'{e}gas et~al.(2004)Vi\'{e}gas, Wattenberg, and Dave]{Viegas-et-al-2004}
F.~B. Vi\'{e}gas, M.~Wattenberg, and K.~Dave.
\newblock Studying cooperation and conflict between authors with history flow visualizations.
\newblock In \emph{Proceedings of the SIGCHI Conference on Human Factors in Computing Systems}, CHI '04, page 575–582. Association for Computing Machinery, 2004.
\newblock URL \url{https://doi.org/10.1145/985692.985765}.

\bibitem[Wang et~al.(2019)Wang, Weisz, Muller, Ram, Geyer, Dugan, Tausczik, Samulowitz, and Gray]{wang2019human}
D.~Wang, J.~D. Weisz, M.~Muller, P.~Ram, W.~Geyer, C.~Dugan, Y.~Tausczik, H.~Samulowitz, and A.~Gray.
\newblock Human-{AI} collaboration in data science: Exploring data scientists' perceptions of automated ai.
\newblock \emph{Proceedings of the ACM on human-computer interaction}, 3\penalty0 (CSCW):\penalty0 1--24, 2019.

\bibitem[Wang et~al.(2023)Wang, Fu, Du, Gao, Huang, et~al.]{wang2023scientific}
H.~Wang, T.~Fu, Y.~Du, W.~Gao, K.~Huang, et~al.
\newblock Scientific discovery in the age of artificial intelligence.
\newblock \emph{Nature}, 620\penalty0 (7972):\penalty0 47--60, 2023.

\bibitem[Wang et~al.(2020)Wang, Wei, Dong, Bao, Yang, and Zhou]{wang2020minilm}
W.~Wang, F.~Wei, L.~Dong, H.~Bao, N.~Yang, and M.~Zhou.
\newblock {MiniLM}: Deep self-attention distillation for task-agnostic compression of pre-trained transformers.
\newblock In \emph{Advances in Neural Information Processing Systems}, volume~33, pages 5776--5788, 2020.

\bibitem[Yang et~al.(2017)Yang, Halfaker, Kraut, and Hovy]{yang-etal-2017-identifying-semantic}
D.~Yang, A.~Halfaker, R.~Kraut, and E.~Hovy.
\newblock Identifying semantic edit intentions from revisions in {W}ikipedia.
\newblock In \emph{Proceedings of the 2017 Conference on Empirical Methods in Natural Language Processing}, pages 2000--2010. Association for Computational Linguistics, 2017.
\newblock \doi{10.18653/v1/D17-1213}.
\newblock URL \url{https://aclanthology.org/D17-1213/}.

\bibitem[Yasseri et~al.(2012)Yasseri, Sumi, Rung, Kornai, and Kert{\'e}sz]{yasseri2012dynamics}
T.~Yasseri, R.~Sumi, A.~Rung, A.~Kornai, and J.~Kert{\'e}sz.
\newblock Dynamics of conflicts in wikipedia.
\newblock \emph{PloS one}, 7\penalty0 (6):\penalty0 e38869, 2012.

\bibitem[Zhao et~al.(2025)Zhao, Liu, Wan, Tang, and Li]{Zhao_Liu_Wan_Tang_Li_2025}
S.~Zhao, Y.~Liu, J.~Wan, T.~Tang, and X.~Li.
\newblock \emph{Human–{AI} Collaboration for Scientific Discovery}, page 239–267.
\newblock Cambridge University Press, 2025.

\bibitem[Zheng et~al.(2023)Zheng, Koh, Ju, Nguyen, May, Webb, and Pan]{zheng2023large}
Y.~Zheng, H.~Y. Koh, J.~Ju, A.~T. Nguyen, L.~T. May, G.~I. Webb, and S.~Pan.
\newblock Large language models for scientific synthesis, inference and explanation.
\newblock \emph{arXiv preprint arXiv:2310.07984}, 2023.

\end{thebibliography}
